\documentclass{article}
\usepackage{amsmath}
\usepackage{graphicx}
\usepackage{multirow}
\usepackage[table,xcdraw]{xcolor}
\usepackage{subcaption}
\usepackage{wrapfig}
\usepackage{enumitem}
\definecolor{best}{RGB}{179, 179, 179}     
\definecolor{second}{RGB}{220, 220, 220}   
 \usepackage[preprint]{neurips_2026}

\usepackage[utf8]{inputenc} 
\usepackage[T1]{fontenc}    
\usepackage{hyperref}       
\usepackage{url}            
\usepackage{booktabs}       
\usepackage{amsfonts}       
\usepackage{nicefrac}       
\usepackage{microtype}      
\usepackage{float}
\usepackage{adjustbox}

\usepackage{amssymb}
\usepackage{color}
\usepackage{xspace}
\usepackage[normalem]{ulem}
\usepackage{bm}

\makeatletter
\DeclareRobustCommand\onedot{\futurelet\@let@token\@onedot}
\def\@onedot{\ifx\@let@token.\else.\null\fi\xspace}

\def\eg{e.g\onedot} 
\def\ie{i.e\onedot}

\makeatother

\def\clap#1{\hbox to 0pt{\hss #1\hss}}%
\def\initials#1{\protect\clap{\protect\smash{\protect\raisebox{1.4ex}{\protect\tiny{\protect\textsf{\protect\textit{#1}}}}}}}%
\makeatletter
\newcommand{\EDIT}[4][]{\protect\@ifundefined{hidecomments}{%
  \protect\strut{\color{#3}{\hspace{0pt}\initials{#2}\protect\sout{#1}{~#4}}}%
  }{#4}}
\newcommand{\NOTEboxed}[3]{\protect\@ifundefined{hidecomments}{%
  {\begin{center}\fbox{\parbox{0.97\linewidth}{\protect\EDIT{#1}{#2}{#3}}}\end{center}}
  }{}}
\newcommand{\COMM}[3]{\protect\@ifundefined{hidecomments}{%
  {\protect\EDIT{#1}{#2}{#3}}
  }{}}
\newcommand{\DefAuthor}[2] 
{%
  \expandafter\newcommand\csname #1edit\endcsname[2][]{\protect\EDIT[##1]{#1}{#2}{##2}}
  \expandafter\newcommand\csname #1\endcsname[1]{\protect\COMM{#1}{#2}{[##1]}}
  \expandafter\newcommand\csname #1boxed\endcsname[1]{\protect\NOTEboxed{#1}{#2}{##1}}
}
\definecolor{dfltgreen}       {rgb}{0.0,0.5,0.0}
\definecolor{dfltred}         {rgb}{0.7,0.0,0.0}
\newcommand{\REVadd}[1]{\protect\@ifundefined{hidecomments}{%
  \strut{\color{dfltgreen}{#1}}}{#1}}
\newcommand{\REVedit}[2][]{\protect\@ifundefined{hidecomments}{%
  \strut{\color{dfltred}{\protect\sout{#1}}\color{dfltgreen}{~#2}}}%
  {#2}}
\makeatother

\definecolor{darkred}  {rgb}{0.7 0.0 0.0}
\definecolor{darkgreen}{rgb}{0.0 0.7 0.0}
\definecolor{darkblue} {rgb}{0.0 0.0 0.8}
\definecolor{darkcyan} {rgb}{0.0 0.5 0.5}
\definecolor{darkmarg} {rgb}{0.5 0.0 0.5}
\DefAuthor{ME}{darkred}
\DefAuthor{PB}{darkgreen}
\DefAuthor{AK}{darkblue}

\title{NRF-GS: Neural Residual Fields for Expressive and Compact Gaussian Splatting}

\author{%
  Pratik Singh Bisht\\
  Computer Graphics Group\\
  University of Siegen\\
  \texttt{pratik.bisht@uni-siegen.de} \\
  \And
  Andreas Kolb\\
  Computer Graphics Group\\
  University of Siegen\\
  \texttt{andreas.kolb@uni-siegen.de} \\
  }

\begin{document}

\maketitle

\begin{abstract}

We revisit the role of appearance modeling in 3D Gaussian Splatting (3DGS) and show that limited expressiveness in view-dependent reflectance is a key driver of representation redundancy. In standard 3DGS, low-order spherical harmonics (SH) are used, restricting the splats' ability to model high-frequency directional effects, which is typically compensated by increasing the number of splats.
We propose \emph{NRF-GS: Neural Residual Fields for Gaussian Splatting}, a hybrid representation that replaces per-splat SH-bases with a shared neural residual field. Each Gaussian encodes a compact set of appearance features and a lambertian base color, while a lightweight \emph{global scene-level MLP} predicts view-dependent residuals conditioned on viewing direction, distance, and per-splat features. This formulation enhances directional reflectance modeling by combining diffuse per-splat reflectance representations with a shared global function for high-frequency details, enabling both higher expressiveness and parameter sharing across splats.
Our key insight is that by accurately capturing high-frequency directional reflectance, especially in specular regions, the GS-representation becomes more expressive, reducing the need for geometrically redundant splats. As a result, NRF-GS achieves comparable or better rendering quality while reducing the number of Gaussians by up to 50\%, and produces visibly improved specular and high-frequency details.

\end{abstract}

\begin{figure}[H]
    \centering
    \begin{minipage}[t]{0.245\textwidth}
        \centering
        \includegraphics[width=\linewidth]{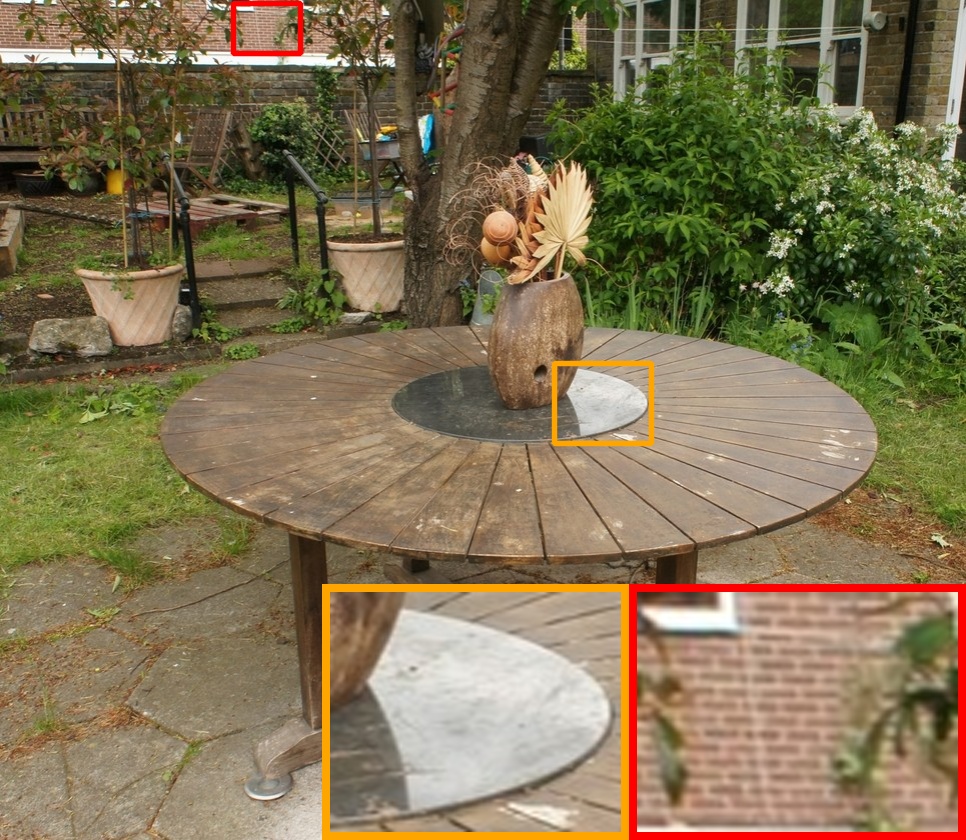}
        \subcaption{GT(PSNR/Points)}
    \end{minipage}
    \begin{minipage}[t]{0.245\textwidth}
        \centering
        \includegraphics[width=\linewidth]{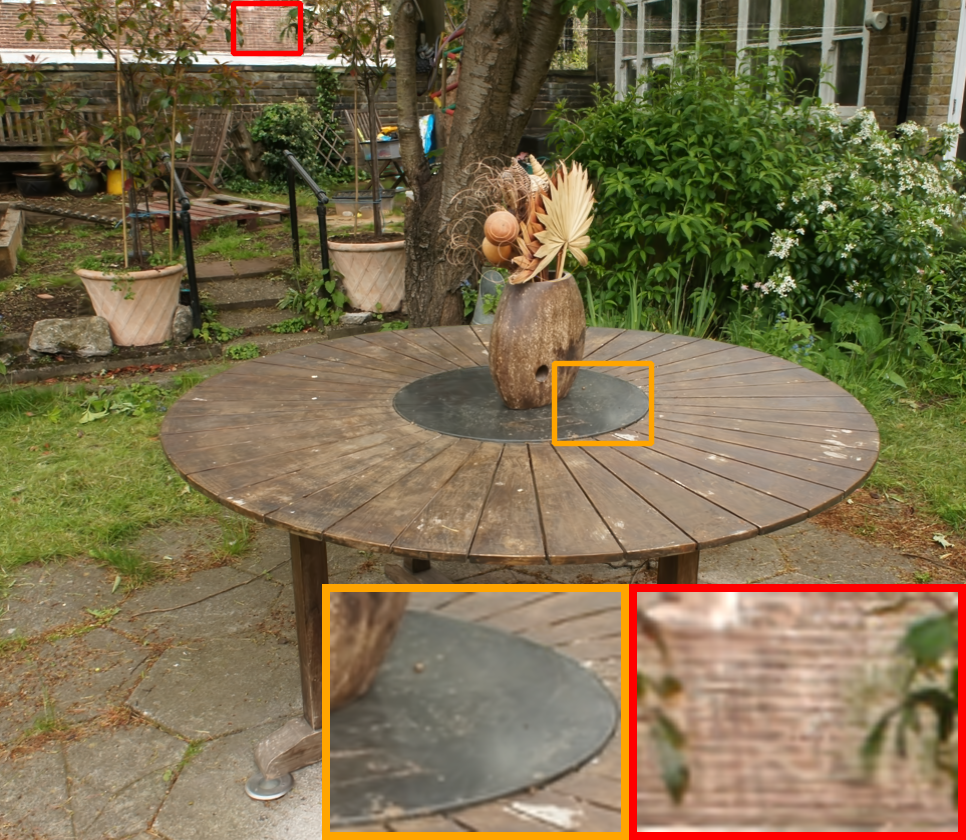}
        \subcaption{3DGS (28.55/5.8M)}
    \end{minipage}
    \begin{minipage}[t]{0.245\textwidth}
        \centering
        \includegraphics[width=\linewidth]{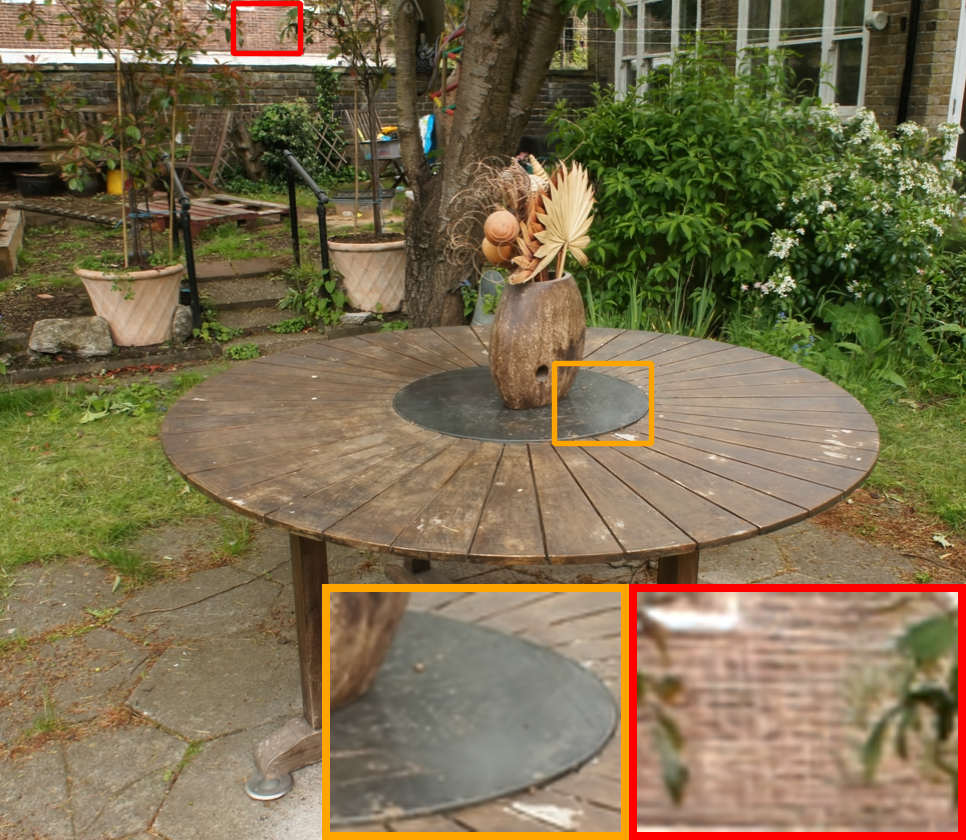}
        \subcaption{VDGS (28.76/5.7M)}
    \end{minipage}
    \begin{minipage}[t]{0.245\textwidth}
        \centering
        \includegraphics[width=\linewidth]{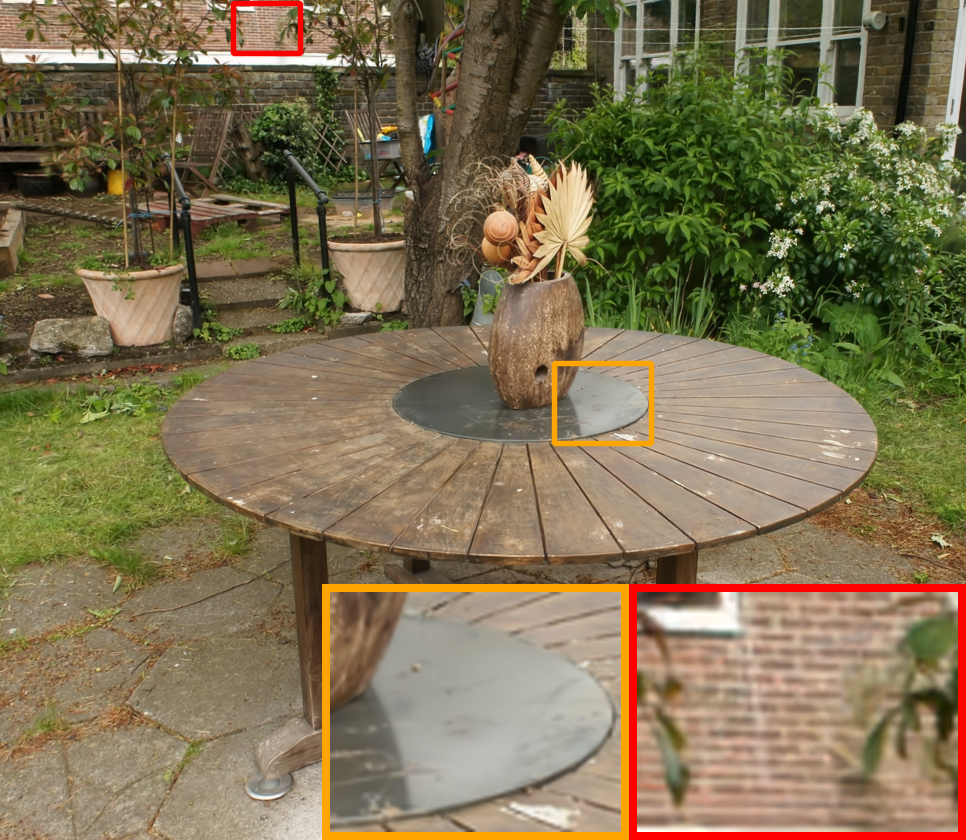}
        \subcaption{Ours (29.75/2.2M)}
    \end{minipage}
    \caption{Novel view from Mip-NeRF 360 dataset - Garden scene. Our method (d) achieves better specular representation on the foreground table as well as sharper details on the background wall with significantly lower number of points compared to other methods. 
    }
    \label{fig:teaser_images}
\end{figure}

\section{Introduction}

Novel view synthesis has seen rapid progress with the emergence of radiance-field-based and explicit primitive-based scene representations~\cite{aliev2020neural, kerbl3Dgaussians, mildenhall2021nerf}. Among these, 3D Gaussian Splatting (3DGS)~\cite{kerbl3Dgaussians} has recently gained significant attention due to its favorable balance between rendering quality, training efficiency, and real-time performance. By representing a scene as a set of anisotropic Gaussians and rendering them via differentiable splatting, 3DGS avoids expensive volumetric sampling while retaining high-fidelity reconstruction. This combination of explicit geometry and efficient rendering makes it particularly attractive for interactive view synthesis and large-scale scene reconstruction.

A central component of 3DGS is its appearance model, where each Gaussian independently represents the view-dependent reflectance using low-order spherical harmonic (SH) coefficients~\cite{muller2006spherical}. Although lightweight and efficient, SH coefficients are modeled per splat (commonly degree 3 in standard settings), which poses certain shortcomings: First, the expressiveness of low-order SH is limited for capturing complex view-dependent effects~\cite{wang2024sg, wang2024specgaussian}. Second, the limited expressiveness is partially compensated for by using more splats to better represent the high-frequency directional effects. Previous works~\cite{malarz2025gaussian, wang2024sg, wang2024specgaussian, zhou2024gaussian} partially address this by replacing the analytical basis with a neural one, but largely entangle geometry and appearance attributes in their neural formulation.

In this work, we propose \textbf{Neural Residual Fields for Gaussian Splatting (NRF-GS)}, in which we deliberately split the representation of geometry and view-dependent reflectance. We, therefore, create a hybrid formulation that retains explicit Gaussian geometry while shifting view-dependent appearance modeling to a shared neural field which enables us to capture complex directional effects while maintaining a compact per-splat geometric representation with significantly lower splat count and competitive quality. Each splat is represented by a diffuse base color and latent features, while a lightweight scene-level network conditioned on frequency-encoded viewing direction, splat distance, and per-splat features predicts a \emph{view-dependent residual reflectance}. This shared network introduces a scene-level appearance prior, enabling more expressive splats and a more efficient distribution. By effectively capturing high-frequency directional effects through the per-splat features and weighted frequency formulation of scene-level MLP, each Gaussian becomes more expressive, reducing the need for redundant splats. Based on this, our key insight is that appearance modeling plays a key role not only in rendering quality but also in controlling spatial complexity in Gaussian splatting.

In summary, this paper makes the following contributions:
\begin{enumerate}[leftmargin=*, itemsep=2pt, topsep=2pt]
    \item We propose \textbf{NRF-GS}, a hybrid representation that models view-dependent appearance using a shared neural residual field conditioned on viewing direction, camera-splat distance, and compact per-splat features.
    \item We introduce a frequency-decomposed network design, showing that low- and high-frequency components should be treated differently by selectively weighted high-frequency modeling.
    \item Through controlled simulations, we demonstrate that a feature + MLP formulation is more expressive than low-order, analytical spherical harmonics under a matched parameter budget.
    \item We show that improved shared appearance modeling leads to more compact scene representations, reducing Gaussian count by up to $50\%$ while maintaining rendering quality.
\end{enumerate}

Our results suggest that residual appearance modeling is a critical and under-explored direction for improving expressiveness in Gaussian splatting, and that combining explicit splat-based rendering with shared neural fields provides a simple and effective path toward more compact and improved scene representations.

\section{Background and Related Works}

\paragraph{Radiance Fields and Explicit Gaussian Representations.}
Neural Radiance Fields (NeRF)~\cite{mildenhall2021nerf} introduced a fully neural representation for novel view synthesis followed by extensive work focused on improving the efficiency and scalability of NeRF. Mip-NeRF~\cite{barron2021mip} addresses aliasing through multiscale sampling, while Plenoxels~\cite{fridovich2022plenoxels} replace neural networks with sparse voxel grids storing density and SH coefficients~\cite{muller2006spherical}. Instant-NGP~\cite{muller2022instant} accelerates training via multi-resolution hash encoding, and related approaches like~\cite{chen2022tensorf} explore structured representation such as tensor decompositions. Neural Reflectance Fields~\cite{bi2020neural} explicitly model view-dependent appearance, while atlas-based representations~\cite{kasten2021layered, schneider2026neural} learn spatially consistent mappings to better capture appearance variations.\newline
To overcome the cost of volumetric rendering, 3D Gaussian Splatting (3DGS)~\cite{kerbl3Dgaussians} represents scenes as collections of anisotropic Gaussians rendered via differentiable splatting~\cite{zwicker2001ewa}, enabling real-time performance while maintaining high reconstruction quality. Subsequent work~\cite{Hasssan_2026_WACV, held20253d, huang20242d, orgun2025dual, zhu20253d} extends this paradigm to improve geometric fidelity, representation, and stability. Recent hybrid approaches~\cite{rong2025nexels, tang2026neuralgs} further incorporate neural fields into explicit representations, but primarily couple geometry and appearance or focus on neural texture generation.

\textbf{Alternative Analytic Directional Bases.}
Other works attempt to improve the directional appearance model of GS by replacing SH with more expressive analytic functions. SG-splatting~\cite{wang2024sg} and ARS-GS~\cite{wu2026ars}, \eg, replaces SH~\cite{muller2006spherical} with Spherical Gaussian (SG) (or Anisotropic-SG)~\cite{xu2013anisotropic} bases for representing directional color variation, which provide a low parameter analytic representation that can capture angular reflectance patterns more flexibly, but remains limited for high-frequency directional reflectance since they use fixed 3 orthogonal SG lobes (equivalent to degree 3 SH).

\textbf{Neural Appearance and Feature-Based Gaussian Representations.}
Some works augment GS with neural networks or learned feature descriptors to improve appearance modeling. Viewing Direction GS (VDGS)~\cite{malarz2025gaussian} combines explicit Gaussian splats with a NeRF-style MLP and hash-encoded Gaussian means and camera position to predict view-dependent reflectance, while neural prediction remains tightly coupled to raw Gaussian parameters and local splat attributes.\newline
\emph{Feature-3DGS}~\cite{zhou2024feature} augments each splat with semantic feature embeddings aligned with 2D foundation models ( DINO~\cite{caron2021emerging}, CLIP~\cite{radford2021learning} and LSeg~\cite{li2022language}), enabling downstream tasks, but still keeping the SH based appearance. \emph{SpecGaussian with Latent Features} (Latent-SpecGS)~\cite{wang2024specgaussian} introduces a universal latent neural descriptor that are decoded by CNNs into diffuse and specular images separately. It is explicitly appearance-oriented, but its diffuse/specular decomposition and feature-map decoding are performed in image-space rather than at splat level.\newline
\emph{Gaussian Splatting with Neural Basis (GSNB)}~\cite{zhou2024gaussian} proposes to model view-dependent appearance using a learned set of basis functions, where per-Gaussian coefficients are combined with shared neural basis representations to produce directional color. This formulation increases the expressiveness of the appearance model by extending the degree 3 SH with additional neural information, but it is memory intensive due to the additional basis functions for higher frequency details.\newline
Another recent method, \emph{Neural Shell Texture Splatting (NeST)}~\cite{zhang2025neural} decouples geometry and appearance by representing scene surfaces with 2D Gaussians and modeling their appearance through a shared hash-encoded neural texture field which focuses on spatially varying surface textures, enabling detailed appearance reconstruction with fewer geometric primitives. \emph{Neural Texture Splatting (NTS)}~\cite{wang2025neural} introduces local RGBA texture fields predicted through a shared global tri-plane neural representation, improving per-splat expressiveness for appearance and geometry reconstruction.

NRF-GS differs from the approaches described above in where and how appearance capacity is introduced. Unlike analytic bases(SH/SG), it replaces per-splat directional bases with a shared neural residual field. Unlike VDGS/GSNB, which condition on raw Gaussian parameters or add neural basis extensions, and NeST/NTS, which model spatially varying neural textures, NRF-GS uses learned appearance features and a lightweight scene-level MLP to predict residual color and opacity directly at the splat level. Unlike Latent-SpecGS, which decodes diffuse and specular components in image space, NRF-GS remains integrated into the splatting pipeline. Specifically, our design decouples view-dependent appearance from the full Gaussian parameterization, improves high-frequency reflectance modeling, and reduces the number of splats.

\section{Method}
\label{s:method}

Our \textbf{Neural Residual Fields for Gaussian Splatting (NRF-GS)} combines the standard spatial representation using Gaussians (see Sec.~\ref{s:method.GS}) and replaces the Spherical Harmonics (SH) part in the appearance model with a learned, shared Neural Residual Field (NRF); see Sec.~\ref{s:method.NRF}. One key component of NRF is the decomposition of directional encoding into low- and high-frequency components, which is depicted in Sec.~\ref{s:method.frequ-decomp}.

\subsection{3D Gaussian Splatting in a Nutshell}
\label{s:method.GS}
In order to establish a common notational basis, we will briefly recap the standard 3D Gaussian Splatting (3DGS) pipeline; for further details, refer to ~\cite{kerbl3Dgaussians}.

3DGS represents a scene as a set of $N$ anisotropic Gaussian splats $\mathcal{G}_i = \{ \mathbf{x}_i, \mathbf{\Sigma}_i, \alpha_i^{\text{base}}, \mathbf{c}_i^{\text{base}} \},$
where Gaussian splat index $i \in \{1,\dots,N\}$, Gaussian position (\ie, mean) $\mathbf{x}_i \in \mathbb{R}^3$, covariance $\mathbf{\Sigma}_i \in \mathbb{R}^{3\times3}$, base opacity $\alpha_i^{\text{base}} \in [0,1]$, and view-independent base color $\mathbf{c}_i^{\text{base}}$. The directional appearance, \ie, view-dependent reflectance, is modeled using spherical harmonics (SH)~\cite{muller2006spherical}:
\begin{equation}
\label{eq:sh_color}
\mathbf{c}_i(\mathbf{d}) = \sum_{l=1}^{L} \sum_{m=-l}^{l} \mathbf{w}_{i,lm} \, Y_{lm}(\mathbf{d}),
\end{equation}
where unit sphere viewing direction $\mathbf{d} \in \mathbb{S}^2$, SH basis function $Y_{lm}(\cdot)$ of degree $l$ and order $m$, and per-Gaussian coefficients $\mathbf{w}_{i,lm} \in \mathbb{R}^3$ (one set per color channel) which are learned. In practice, a low-order expansion (e.g., max SH degree, $L=3$) is used for efficiency, resulting in $(L+1)^2$ basis functions per channel.
In total, there are 48 parameters used to encode view-dependent reflectance information and a front-to-back accumulation of the individual splat contributions per pixel. For a given splat $i$, the estimated color is computed as $\mathbf{c}_i=\mathbf{c}_i^{\text{base}}+\mathbf{c}_i(\mathbf{d})$, where $\mathbf{c}_i^{\text{base}}$ is the $l=0$ dc component.

3DGS utilizes a differentiable rasterization applied to the splat representation, which incorporates an efficient, tile-based binning and sorting on GPU as well as front-to-back compositing using the splat's estimated color $\mathbf{c}_i$ and (base-)opacity $\alpha_i^{\text{base}}$ as follows:
\begin{equation}
C = \sum_{i=1}^{N} T_i \alpha_i^{\text{base}} \mathbf{c}_i, \quad
T_i = \prod_{j<i} (1 - \alpha_j^{\text{base}}),
\end{equation}
where $T$ is the remaining transparency for the splats behind the current one. Note that we deliberately use $\alpha_i^{\text{base}}$ here for 3DGS, as our NRF-GS approach does not use only the base opacity; see Sec.~\ref{s:method.frequ-decomp}.

Further components of the 3DGS pipeline consist of densification and pruning heuristics to adapt the local splat density and to remove outliers during the learning process~\cite{kerbl3Dgaussians}. Our NRF-GS approach only modifies the color and density representation and estimation in the rendering state and leaves all other stages unchanged. This ensures that any observed improvements are attributable solely to the proposed appearance modeling, while being fully compatible with existing 3DGS implementations.

\begin{figure*}[t]
\centering
\includegraphics[width=\textwidth]{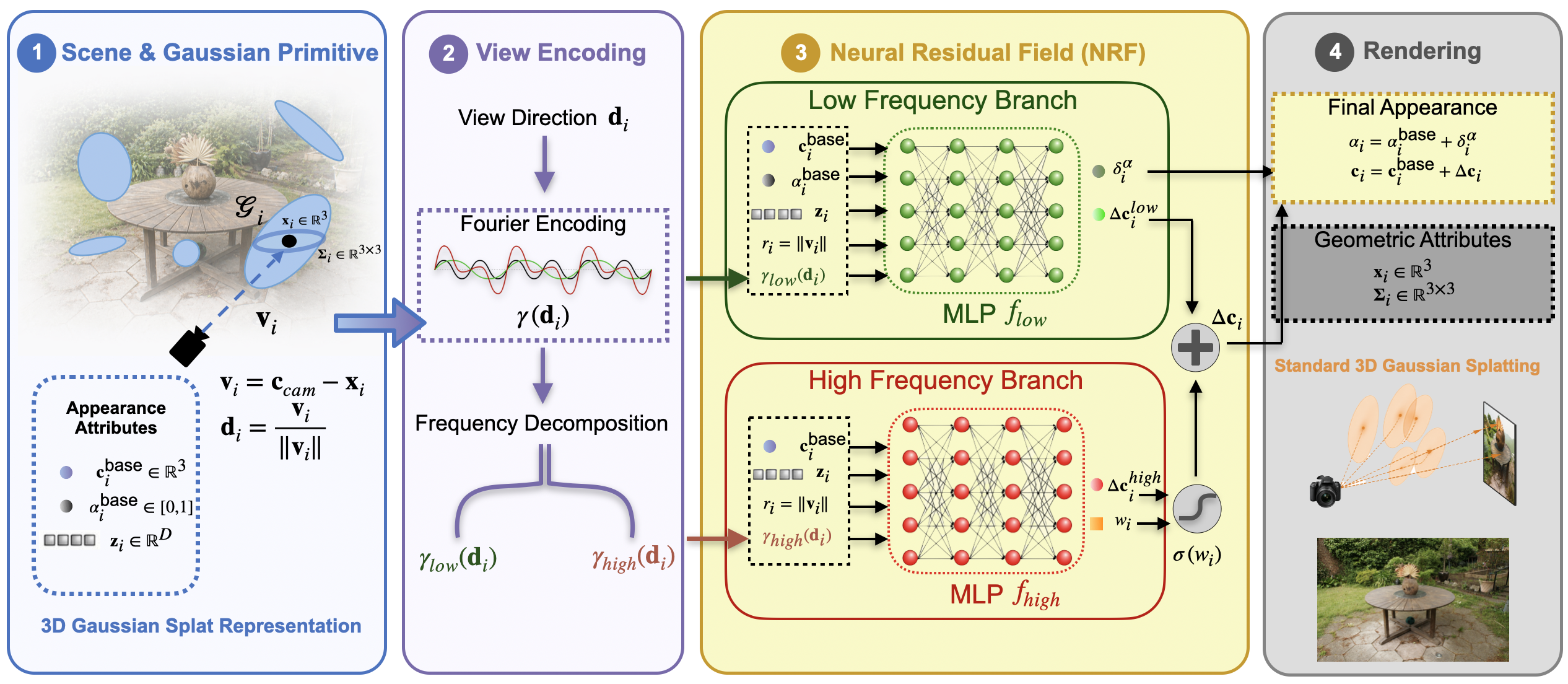}
\caption{
\textbf{NRF-GS architecture.}
Per Gaussian geometric attributes $(\mathbf{x}_i, \mathbf{\Sigma}_i)$ and appearance attributes ($\mathbf{c}_i^{\text{base}}$, $\alpha_i^{\text{base}}$, and a latent feature $\mathbf{z}_i$). The viewing direction $\mathbf{d}_i$ is frequency-encoded and decomposed into low ($\gamma_{\text{low}}(\mathbf{d}_i)$) and high-frequency ($\gamma_{\text{high}}(\mathbf{d}_i)$) components. Two MLP branches ($f_{\text{low}}$, $f_{\text{high}}$) predict a low-frequency residual color and opacity residual $(\Delta \mathbf{c}_i^{\text{low}}, \delta_i^\alpha)$ and a high-frequency color residual with a learned modulation weight $(\Delta \mathbf{c}_i^{\text{high}}, w_i)$. 
The final color and opacity are then rendered using the unchanged standard Gaussian splatting rasterizer.}
\vspace{-1.5em}
\label{fig:nrfgs_architecture}
\end{figure*}

\subsection{Shared Neural Residual Fields}
\label{s:method.NRF}
To improve the limited ability of existing GS-methods in efficiently representing high-frequency direction reflectance, \eg, of a low-order analytical basis function, NRF-GS replaces the per-Gaussian SH representation (Eq.~\ref{eq:sh_color}) with a shared neural residual formulation (Fig.~\ref{fig:nrfgs_architecture}) while retaining explicit geometry representation via splats.

In NRF-GS, each Gaussian is represented as $\mathcal{G}_i = \{ \mathbf{x}_i, \mathbf{\Sigma}_i, \alpha_i^{\text{base}}, \mathbf{c}_i^{\text{base}}, \mathbf{z}_i \},$
where $\mathbf{c}_i^{\text{base}} \in \mathbb{R}^3$ and $\alpha_i^{\text{base}} \in [0,1]$ represent base color and base opacity, and $\mathbf{z}_i \in \mathbb{R}^D$ is a learned latent feature that captures the appearance-related properties of the splat.
The final color is modeled as:
$\mathbf{c}_i = \mathbf{c}_i^{\text{base}} + \Delta \mathbf{c}_i,$
where $\Delta \mathbf{c}_i$ is a residual predicted from the Neural Residual Field (NRF) model. The view-independent base color $\mathbf{c}_i^{\text{base}}$ captures the dominant, diffuse component of the appearance and provides a stable anchor that is optimized directly by the strong supervision provided by all views (as done in the standard 3DGS~\cite{kerbl3Dgaussians} pipeline). The role of shared NRF is to model only the directional variations, essentially replacing the analytical, smooth, low frequency Spherical Harmonic basis. For each Gaussian, the viewing direction is computed as:
\begin{equation}
\mathbf{v}_i = \mathbf{c}_{cam} - \mathbf{x}_i, \quad
r_i = \|\mathbf{v}_i\|, \quad
\mathbf{d}_i = \frac{\mathbf{v}_i}{r_i},
\end{equation}
where $\mathbf{c}_{cam}$ denotes the camera center, $r_i$ the camera-splat distance, and $\mathbf{x}_i$ the splat position. In addition, a scalar distance feature is defined as $\tilde r_i = \log(\max(r_i, \epsilon))$, where $\epsilon$ is a small constant for numerical stability. This provides a compact representation of scale and helps the model adapt appearance behavior across varying distances based on the observation that the apparent angular frequency of view-dependent effects varies with distance (Tab.~\ref{tab:ablation}). For distant splats, directional variation appears smoother due to reduced parallax, whereas nearby splats exhibit stronger high-frequency variation. Conditioning on distance as well as frequency decomposition  approach (Sec.~\ref{s:method.frequ-decomp}) allows the model to adapt its response accordingly (refer Tab.~\ref{tab:ablation}). To enable modeling of directional variation at multiple scales, the viewing direction is encoded using standard Fourier encoding similar to NeRF baselines~\cite{mildenhall2021nerf, muller2022instant, tancik2020fourier}:
\begin{equation}
\label{eq:freq_enc}
\gamma(\mathbf{d}_i) =
\mathbf{d}_i,\left[\sin(2^k \pi \mathbf{d}_i),\; \cos(2^k \pi \mathbf{d}_i)\right]_{k=0}^{K-1}.
\end{equation}

\subsection{Frequency-Decomposed Neural Appearance}
\label{s:method.frequ-decomp}
A key design choice in NRF-GS is the decomposition of directional encoding into low- and high-frequency components (Fig.~\ref{fig:nrfgs_architecture}). This is motivated by empirical observations that different scenes exhibit varying sensitivity to directional frequency: in particular, bounded (indoor) scenes often benefit from higher-frequency angular modeling, while unbounded (outdoor) scenes are more sensitive to overfitting from high-frequency encoding (refer Sec.~\ref{subsec:results_abl} and supplementary Tab.~\ref{tab:single_vs_ours}). This suggests that high-frequency directional effects should be modeled selectively rather than uniformly. Accordingly, we split the encoding as follows:
\begin{equation}
\gamma(\mathbf{d}_i) = [\gamma_{\text{low}}(\mathbf{d}_i), \gamma_{\text{high}}(\mathbf{d}_i)],
\end{equation}
where $\gamma_{\text{low}}(\mathbf{d}_i)$ captures coarse angular variation (including the lowest frequency components), and $\gamma_{\text{high}}(\mathbf{d}_i)$ captures higher-frequency directional details. By separating low- and high-frequency components, the model ensures that smooth appearance variations are always represented, while high-frequency effects are selectively incorporated. This prevents overfitting to noisy directional signals, particularly in unbounded scenes where view sampling is sparse.

Two separate neural branches are used to model the low- and high-frequency components.

\paragraph{Low-frequency branch ($f_{\text{low}}$).}
The low-frequency branch models the smooth appearance variations that gradually change with the viewing direction. It takes as input the latent feature $\mathbf{z}_i$, base color $\mathbf{c}_i^{\text{base}}$, base opacity $\alpha_i^{\text{base}}$, scalar distance $\tilde r_i$ and $\gamma_{\text{low}}(\mathbf{d}_i)$ given to the light-weight $f_{\text{low}}$ MLP:
\begin{equation}
(\Delta \mathbf{c}_i^{\text{low}}, \delta_i^\alpha)
=
f_{\text{low}}(\mathbf{z}_i, \mathbf{c}_i^{\text{base}}, \alpha_i^{\text{base}}, \tilde r_i, \gamma_{\text{low}}(\mathbf{d}_i)).
\end{equation}

Here, $\Delta \mathbf{c}_i^{\text{low}}$ denotes the low-frequency color residual and $\delta_i^\alpha$ is an intermediate opacity residual. Similarly to the findings in VDGS~\cite{malarz2025gaussian}, we observed that updating the base opacity $\alpha_i^{\text{base}}$ with the residual based on per-splat learned features improves the representation further. Restricting opacity prediction to this low-frequency branch encourages stable and smooth transmittance, and avoids introducing high-frequency instability and flickering behavior during rendering~\cite{malarz2025gaussian}. 

\paragraph{High-frequency branch ($f_{\text{high}}$).}
The high-frequency branch takes the decomposed higher frequencies $\gamma_{\text{high}}(\mathbf{d}_i)$ from the full frequency encoding $\gamma_(\mathbf{d}_i)$ along with other appearance attributes (except opacity) and distance to model sharper directional effects.
\begin{equation}
(\Delta \mathbf{c}_i^{\text{high}}, w_i)
=
f_{\text{high}}(\mathbf{z}_i, \mathbf{c}_i^{\text{base}}, \tilde r_i, \gamma_{\text{high}}(\mathbf{d}_i)).
\end{equation}

The scalar $w_i$ acts as a learned modulation signal controlling the contribution of high-frequency components. The design of $f_{\text{high}}$ MLP is similar to the $f_{\text{low}}$ branch but does not get the base density as input and results in a separate set of learned parameters.

\paragraph{Residual composition.}
The final color and opacity are obtained through bounded residual updates. The view-dependent color residual is computed as
\begin{equation}
\label{eq:res_comp}
\Delta \mathbf{c}_i =
s_c \cdot \tanh\left(
\Delta \mathbf{c}_i^{\text{low}} + \sigma(w_i)\Delta \mathbf{c}_i^{\text{high}}
\right),
\end{equation}
where $\sigma(\cdot)$ denotes the sigmoid function and $s_c$ is a scaling factor hyper-parameter that bounds the magnitude of the color residual. $\sigma(w_i)$ enables adaptive incorporation of high-frequency effects while bounded $\tanh$ non-linearity maintains stable predictions. The learned modulation is particularly important because not all regions exhibit strong view-dependent effects; enforcing uniform high-frequency modeling can lead to artifacts and overfitting (see Sec.~\ref{subsec:results_abl} and the supplementary Tab.~\ref{tab:single_vs_ours}).

\begin{table}[t]
\centering
\begin{minipage}[t]{0.48\linewidth}
\centering
\caption{Ablation study of NRF-GS components on MipNeRF360 dataset.}
\label{tab:ablation}
\resizebox{\linewidth}{!}{
\begin{tabular}{lcccc}
\toprule
Method & PSNR$\uparrow$ & SSIM$\uparrow$ & LPIPS$\downarrow$ & L1$\downarrow$ \\
\midrule
NRF-GS (Ours) & 27.86 & 0.814 & 0.191 & 0.029 \\
w/o split & 27.38 & 0.796 & 0.211 & 0.032 \\
w/o opacity & 27.51 & 0.801 & 0.195 & 0.031 \\
w/o distance & 27.65 & 0.809 & 0.201 & 0.031 \\
\bottomrule
\end{tabular}
}
\end{minipage}
\hfill
\begin{minipage}[t]{0.48\linewidth}
\centering
\caption{Average parameter and quality comparison across simulation setups.}
\label{tab:sim_params}
\resizebox{\linewidth}{!}{
\begin{tabular}{lccc}
\toprule
Setup & \#Points & Params (SH/NRF) & PSNR (SH / NRF)$\uparrow$ \\
\midrule
Synthetic & 328 & 15,744 / 15,707 & 21.1 / 29.0 \\
Real (pixel) & 328 & 15,744 / 15,707 & 27.8 / 28.8 \\
Real (image) & 160k & 7.68M / 5.60M & 31.3 / 31.9 \\
\bottomrule
\end{tabular}
}
\end{minipage}
\vspace{-1.5em}
\end{table}

Finally, the opacity is updated using a similar residual formulation:
\begin{equation}
\label{eq:final_opacity}
\alpha_i = \alpha_i^{\text{base}} + s_\alpha \cdot \tanh(\delta_i^\alpha),
\end{equation}
where $\delta_i^\alpha$ is the predicted opacity residual from only low-frequency branch and $s_\alpha$ controls its magnitude. Bounding the opacity update ensures stable transmittance accumulation during rendering.

\subsection{Implementation Details}
\label{subsec:impl_details}

We use a shared neural appearance model implemented with tiny-cuda~\cite{tiny-cuda-nn} FullyFusedMLP layers, similar to other neural baselines~\cite{malarz2025gaussian,zhou2024gaussian}. Each Gaussian stores a latent feature $\mathbf{z}_i \in \mathbb{R}^{32}$ along with $\alpha_i^{\text{base}}$ and $ \mathbf{c}_i^{\text{base}}$. The viewing direction is encoded using $K=3$ bands (Eq.~\ref{eq:freq_enc}), split into low-frequency $\gamma_{\text{low}}(\mathbf{d}_i)$ (raw direction and $k=0$) and high-frequency $\gamma_{\text{high}}(\mathbf{d}_i)$ (remaining components). Both branches use a hidden layer of 64 neurons and ReLU activation, with hyper-parameters $s_c = 0.2$ and $s_\alpha=0.2$. This constrains updates to act as refinements over the base appearance and improves early training stability. We additionally perform a short \emph{warmup}, once per view, where the residual field is trained to predict zero color residual and neutral opacity across views. This initializes the model to a view-independent state and avoids early overfitting to directional noise.

\subsection{Compatibility}
\label{subsec:compatibility}
NRF-GS modifies only the appearance representation of standard 3DGS, replacing per-Gaussian SH coefficients with a base color, opacity, latent appearance feature, and a lightweight shared decoder. Gaussian geometry, projection, rasterization, compositing, and densification/pruning remain unchanged. This modular design makes NRF-GS architecturally compatible with feed-forward Gaussian prediction and compact or compressed 3DGS approaches, although integration would require adapting their appearance prediction or encoding components

\section{Experiments and Results}
\label{sec:experiments}

We perform extensive experiments with multiple datasets and multiple runs, mainly to compare our method with standard 3DGS~\cite{kerbl3Dgaussians} and other neural methods for view-dependent reflectance prediction (\cite{malarz2025gaussian} and ~\cite{zhou2024gaussian}). Additionally, we propose a study to validate our claim of more expressiveness of our feature based NRF method compared to fixed analytical approach under \emph{equal-memory conditions}.

\subsection{Ablation Study}
\label{subsec:results_abl}
We perform the ablation results and justification for three main design choices of NRF-GS:

\textbf{Impact of frequency decomposition.}
We compare our split low-/high-frequency design with a single-branch MLP processing all frequencies jointly. For fairness, the single branch uses a larger hidden size (128 vs 64), resulting in a slightly higher parameter count ($\sim$8.0k vs $\sim$6.7k). As shown in Tab.~\ref{tab:ablation}, w/o split, the single-branch variant consistently underperforms. Per-scene results (supplementary Tab.~\ref{tab:single_vs_ours}) show larger degradation in challenging unbounded scenes (\emph{Bicycle}, \emph{Garden}, \emph{Stump}, \emph{Flowers}). This indicates that frequency separation improves robustness, especially when high-frequency signals are sparse or unreliable due to limited parallax.

\textbf{Effect of opacity residual learning.}
We also test a variant that predicts only color (no opacity residual). Consistent with~\cite{malarz2025gaussian}, this leads to slightly worse performance (Tab.~\ref{tab:ablation}, w/o opacity, middle row), showing that learning opacity residuals (or view-conditioned opacity correction) provides additional flexibility and improves overall reconstruction quality.

\textbf{Effect of distance conditioning.}
Removing distance input (Tab.~\ref{tab:ablation},  w/o distance, last row) results in a small but consistent drop in performance. This shows that distance implicitly encodes the strength of parallax and the scale at which directional variations are observed. Without this information, the model must infer such effects solely from the viewing direction, which becomes ambiguous, especially for far-away splats, where angular changes correspond to smaller apparent variations.

\begin{figure*}[t]
    \centering

    \begin{subfigure}[t]{0.19\textwidth}
        \centering
        \includegraphics[width=\linewidth]{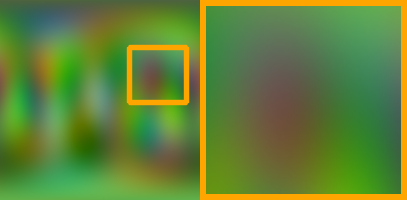}
        \caption{GT}
    \end{subfigure}
    \begin{subfigure}[t]{0.19\textwidth}
        \centering
        \includegraphics[width=\linewidth]{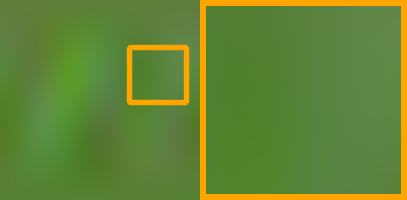}
        \caption{SH (20.98db)}
    \end{subfigure}
    \begin{subfigure}[t]{0.19\textwidth}
        \centering
        \includegraphics[width=\linewidth]{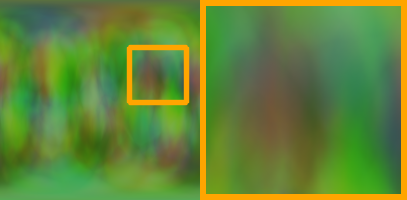}
        \caption{NRF (29.28db)}
    \end{subfigure}
    \begin{subfigure}[t]{0.19\textwidth}
        \centering
        \includegraphics[width=\linewidth]{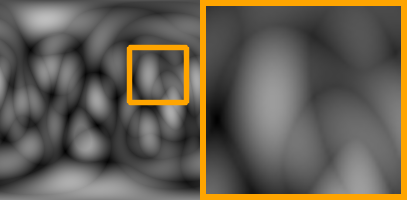}
        \caption{SH MSE (0.14)}
    \end{subfigure}
    \begin{subfigure}[t]{0.19\textwidth}
        \centering
        \includegraphics[width=\linewidth]{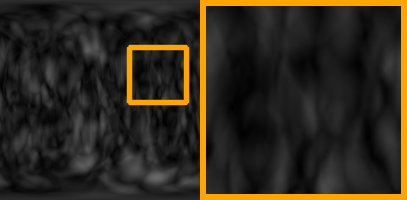}
        \caption{NRF MSE (0.02)}
    \end{subfigure}

    \vspace{0.5em}

    \begin{subfigure}[t]{0.19\textwidth}
        \centering
        \includegraphics[width=\linewidth]{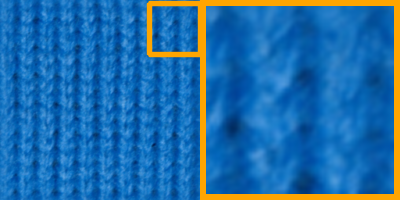}
        \caption{GT}
    \end{subfigure}
    \begin{subfigure}[t]{0.19\textwidth}
        \centering
        \includegraphics[width=\linewidth]{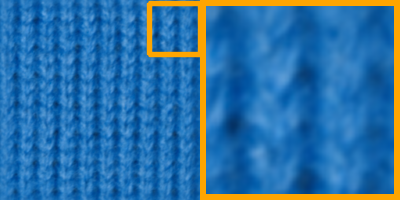}
        \caption{SH (28.24db)}
    \end{subfigure}
    \begin{subfigure}[t]{0.19\textwidth}
        \centering
        \includegraphics[width=\linewidth]{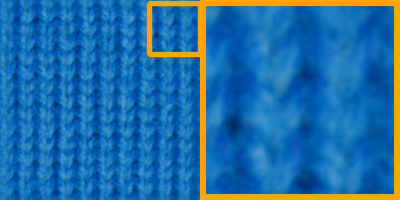}
        \caption{NRF (31.92db)}
    \end{subfigure}
    \begin{subfigure}[t]{0.19\textwidth}
        \centering
        \includegraphics[width=\linewidth]{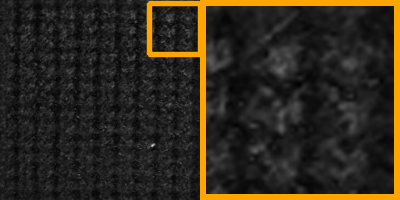}
        \caption{SH MSE (0.010)}
    \end{subfigure}
    \begin{subfigure}[t]{0.19\textwidth}
        \centering
        \includegraphics[width=\linewidth]{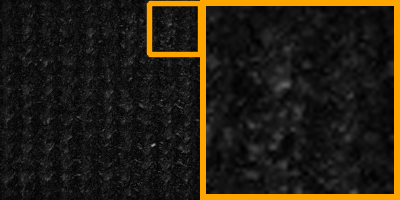}
        \caption{NRF MSE (0.006)}
    \end{subfigure}

    \caption{
    Simulation results for synthetic degree-5 SH (top row) and real BTF data~\cite{sattler-2003-efficient} (bottom row). 
    Each row shows ground truth, degree-3 SH and NRF reconstruction, and corresponding error maps.
    }
    \label{fig:simulation_combined}
    \vspace{-1.5em}
\end{figure*}

\subsection{Expressiveness Evaluation}
\label{subsec:simulation}

To evaluate view-dependent appearance modeling in terms of expressiveness independent of geometry, visibility, and rasterization, we design a controlled study comparing degree-3 SH modeling with our feature-based neural residual formulation. To independently assess the color estimation dependent on the viewing direction $\mathbf{d}$, we observe a set of sampled viewing directions with corresponding ground-truth color and train a model to reconstruct the directional reflectance using (a) degree-3 SH (\textbf{SH model}) and (b) our per-splat latent features with a shared MLP (\textbf{NRF-GS}). Tab.~\ref{tab:sim_params} summarizes the average parameter budgets and reconstruction quality across all simulation settings, where for two settings we match the learnable parameter budgets.

\textbf{Synthetic data.} In the \emph{synthetic} experiment, GT directional appearance is generated using degree-5 SH, producing high-frequency color variation. Tab.~\ref{tab:sim_params} shows that the shared MLP conditioned on latent features can represent high-frequency directional structure more effectively under matched budget. The results in Fig.~\ref{fig:simulation_combined} (top) show that NRF-GS recovers details beyond the low-order SH basis.

\textbf{Real-world data.} We also evaluate on real measured reflectance from the BTF dataset~\cite{sattler-2003-efficient}, which contains material samples captured under many view and illumination directions. In the \emph{pixel-based} equal-memory setting, NRF-GS again improves reconstruction quality despite sparse angular sampling (Tab.~\ref{tab:sim_params}). Since the equal memory setting uses only 328 sampled pixels, we additionally perform a dense $200 \times 200$ full \emph{image-based} experiment for quantitative evaluation, in which NRF-GS uses fewer parameters while achieving higher quality; see Fig.~\ref{fig:simulation_combined} (bottom).

\textbf{Higher order SH comparison.} To further investigate whether increasing the order of conventional spherical harmonics can achieve comparable quality with reduced primitives, we do the full novel view synthesis evaluation for degree-4 and degree-5 SH baselines. As shown in the supplementary Tab.~\ref{tab:sh_degree_comparison}, higher-order SH yields only marginal reconstruction improvements, but significantly increases memory and training time, while NRF-GS achieves higher quality with substantially fewer Gaussians.

Overall, Fig.~\ref{fig:simulation_combined}, Tab.~\ref{tab:sim_params} and Tab.~\ref{tab:sh_degree_comparison} show that feature-based neural representations are more expressive than fixed low-order SH under matched budgets and scale better under relaxed constraints.

\subsection{Benchmark Experiments}

We evaluate \textbf{NRF-GS} on standard benchmark datasets for novel view synthesis and compare them with various baselines.

\begin{wrapfigure}{r}{0.4\linewidth}
\vspace{-0.8em}
\centering
\includegraphics[width=\linewidth]{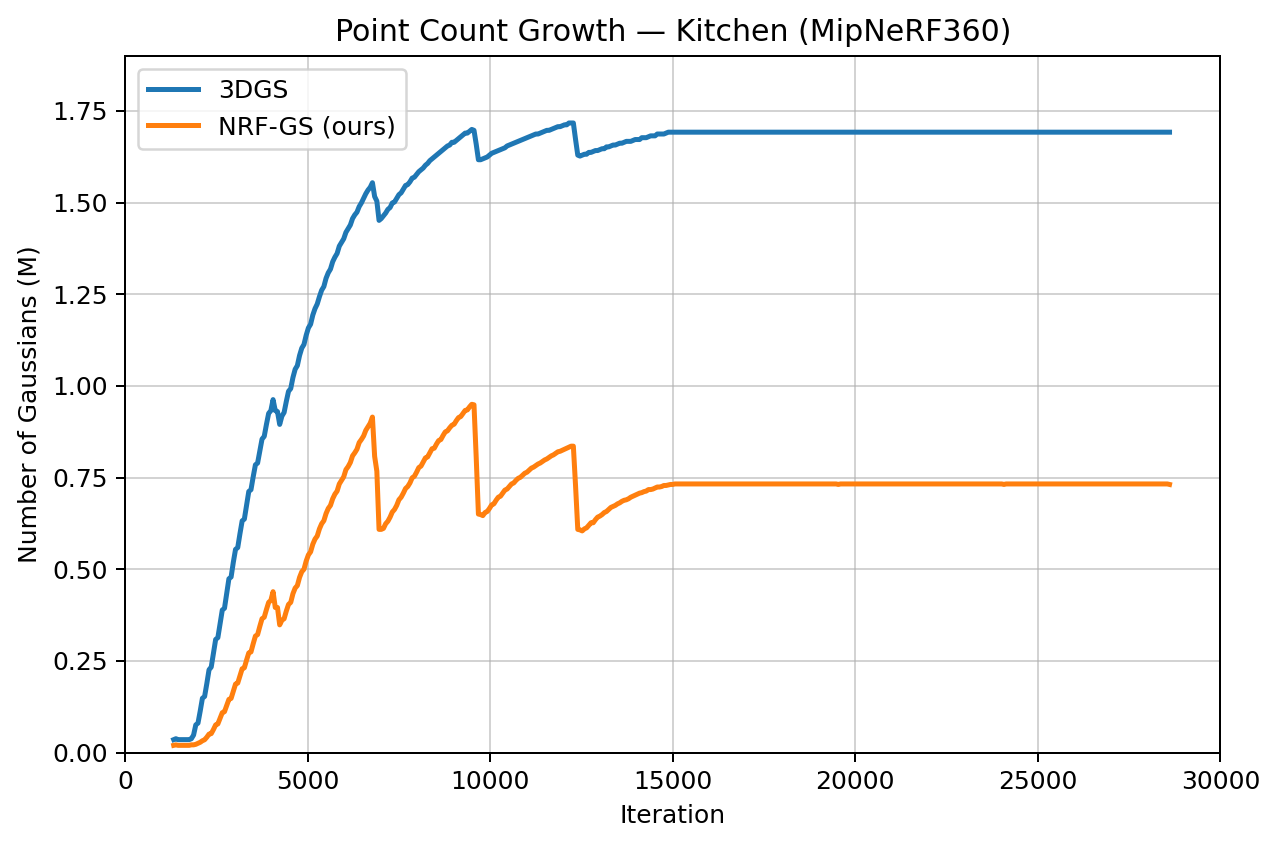}
\caption{
Point count growth for the \textbf{Kitchen} scene (MipNeRF360). NRF-GS converges to fewer Gaussians under identical densification/pruning.
}
\label{fig:point_growth}
\vspace{-1em}
\end{wrapfigure}

\subsubsection{Datasets and Baselines}

We evaluate \textbf{NRF-GS} on three standard datasets: \textbf{MipNeRF360}~\cite{barron2022mip}, \textbf{DL3DV}~\cite{ling2024dl3dv}, and \textbf{Tanks and Temples (TnT)}~\cite{knapitsch2017tanks}. \textbf{MipNeRF360} and \textbf{TnT} are taken from the NerfBaselines~\cite{kulhanek2024nerfbaselines} package which uses standard resolutions ($\sim$1400p and 960p respectively), while \textbf{DL3DV} is evaluated at its standard benchmark~\cite{ling2024dl3dv} resolution of 960p. We randomly chose 9 scenes from the DL3DV dataset (10k scenes) to match the count with MipNeRF360 (9 scenes), while for \textbf{TnT} dataset all 19 scenes are used. For baseline methods, we compare \textbf{NRF-GS} against standard \textbf{3DGS}~\cite{kerbl3Dgaussians} and two neural color prediction methods, \textbf{VDGS}~\cite{malarz2025gaussian} and \textbf{GSNB}~\cite{zhou2024gaussian}. We exclude ~\cite{wang2024specgaussian} due to its image-space formulation and high memory requirements, while ~\cite{wang2024sg} is not included due to the lack of a publicly available reference implementation, preventing a reliable and reproducible comparison.

\if 10
\begin{table*}[t]
\centering
\caption{
Average test-set results on MipNeRF360, DL3DV, and Tanks and Temples. 
Points are reported in millions. 
\emph{MipNeRF360*} excludes \emph{Bicycle} and \emph{Garden}, where GSNB goes out-of-memory. 
Best and second-best values in each dataset are highlighted.
}
\label{tab:main_results}
\resizebox{\textwidth}{!}{
\begin{tabular}{llcccccc}
\toprule
Dataset & Method & Points (M)$\downarrow$ & Time (min)$\downarrow$ & MS-SSIM$\uparrow$ & LPIPS$\downarrow$ & L1$\downarrow$ & PSNR$\uparrow$\\
\midrule
\multirow{3}{*}{MipNeRF360}
& 3DGS   & \cellcolor{second}3.359 & \cellcolor{best}\textbf{26.46} & \cellcolor{second}0.813 & \cellcolor{best}\textbf{0.184} & 0.030 & 27.40 \\
& VDGS   & 3.548 & 41.27 & 0.813 & \cellcolor{second}0.186 & \cellcolor{second}0.029 & \cellcolor{second}27.65 \\
& \textbf{Ours} & \cellcolor{best}\textbf{1.656} & \cellcolor{second}31.38 & \cellcolor{best}\textbf{0.814} & 0.191 & \cellcolor{best}\textbf{0.029} & \cellcolor{best}\textbf{27.86} \\
\midrule
\multirow{4}{*}{MipNeRF360*}
& 3DGS   & 2.605 & \cellcolor{best}\textbf{24.22} & 0.812 & \cellcolor{second}0.199 & 0.030 & 27.74 \\
& VDGS   & 2.828 & 36.70 & \cellcolor{second}0.813 & 0.200 & \cellcolor{second}0.029 & \cellcolor{second}28.03 \\
& GSNB   & 2.667 & 59.06 & 0.800 & 0.201 & 0.031 & 27.71 \\
& NeST   & \cellcolor{best}\textbf{0.855} & 327.9 & 0.780 & \cellcolor{best}\textbf{0.177} & 0.033 & 26.95 \\
& NTS\^   & 0.989 & 274.9 & 0.923 & 0.136 & 0.016 & 31.11 \\
& \textbf{Ours} & \cellcolor{second}1.361 & \cellcolor{second}28.83 & \cellcolor{best}\textbf{0.815} & 0.205 & \cellcolor{best}\textbf{0.028} & \cellcolor{best}\textbf{28.27} \\
\midrule
\multirow{4}{*}{DL3DV}
& 3DGS   & 1.158 & \cellcolor{best}\textbf{12.74} & 0.916 & 0.088 & 0.021 & 29.48 \\
& VDGS   & 1.404 & 21.15 & 0.918 & 0.089 & 0.021 & 29.92 \\
& GSNB   & 1.032 & 25.28 & \cellcolor{second}0.920 & \cellcolor{second}0.086 & \cellcolor{second}0.020 & \cellcolor{second}30.60 \\
& NeST   & \cellcolor{best}\textbf{0.334} & 196.4 & 0.917 & \cellcolor{best}\textbf{0.072} & 0.021 & 26.79 \\
& \textbf{Ours} & \cellcolor{second}0.595 & \cellcolor{second}14.52 & \cellcolor{best}\textbf{0.925} & 0.089 & \cellcolor{best}\textbf{0.018} & \cellcolor{best}\textbf{30.82} \\
\midrule
\multirow{4}{*}{TnT}
& 3DGS   & 1.868 & \cellcolor{best}\textbf{15.35} & 0.831 & 0.165 & 0.046 & 23.85 \\
& VDGS   & 2.113 & 26.37 & \cellcolor{second}0.834 & \cellcolor{best}\textbf{0.163} & \cellcolor{second}0.045 & \cellcolor{second}24.12 \\
& GSNB   & \cellcolor{second}1.705 & 35.75 & 0.830 & \cellcolor{second}0.164 & 0.046 & 23.92 \\
& \textbf{Ours} & \cellcolor{best}\textbf{1.040} & \cellcolor{second}19.91 & \cellcolor{best}\textbf{0.845} & \cellcolor{second}0.164 & \cellcolor{best}\textbf{0.040} & \cellcolor{best}\textbf{24.68} \\
\bottomrule
\vspace{-2.0em}
\end{tabular}

}
\end{table*}
\fi

\if 11
\begin{table*}[t]
\centering
\caption{
Average test-set results on MipNeRF360, DL3DV, and Tanks and Temples. 
Points are reported in millions. 
\emph{MipNeRF360*} excludes \emph{Bicycle} and \emph{Garden}; \emph{MipNeRF360**} excludes \emph{Bicycle}, \emph{Flowers}, \emph{Garden}, \emph{Stump} and \emph{Treehill}; due to different methods, mainly GSNB, NeST and NTS, going out-of-memory. 
Best and second-best values in each dataset are highlighted.
}
\label{tab:main_results}
\resizebox{\textwidth}{!}{
\begin{tabular}{p{0.08\textwidth}lccccccccc}
\toprule
Dataset & Method & \#Pts (M)$\downarrow$ & Time (m)$\downarrow$ & SSIM$\uparrow$ & LPIPS$\downarrow$ & L1$\downarrow$ & PSNR$\uparrow$ & Model (GB)$\downarrow$ & VRAM (GB)$\downarrow$ & FPS$\uparrow$\\
\midrule
\multirow{3}{0.08\textwidth}{MipNeRF 360}
& 3DGS   & \cellcolor{second}3.359 & \cellcolor{best}\textbf{26.46} & \cellcolor{second}0.813 & \cellcolor{best}\textbf{0.184} & 0.030 & 27.40 & \cellcolor{second}0.769 & \cellcolor{second}8.20 & \cellcolor{best}\textbf{109.35}\\
& VDGS   & 3.548 & 41.27 & 0.813 & \cellcolor{second}0.186 & \cellcolor{second}0.029 & \cellcolor{second}27.65 & 0.844 & 9.13 & 60.13\\
& \textbf{Ours} & \cellcolor{best}\textbf{1.656} & \cellcolor{second}31.38 & \cellcolor{best}\textbf{0.814} & 0.191 & \cellcolor{best}\textbf{0.029} & \cellcolor{best}\textbf{27.86} & \cellcolor{best}\textbf{0.325} & \cellcolor{best}\textbf{7.59} & \cellcolor{second}98.40\\
\midrule
\multirow{5}{0.08\textwidth}{MipNeRF 360*}
& 3DGS   & 2.605 & \cellcolor{best}\textbf{24.22} & 0.812 & \cellcolor{second}0.199 & 0.030 & 27.74 & 0.596 & \cellcolor{second}8.14 & \cellcolor{best}\textbf{115.14\phantom{0}} \\
& VDGS   & 2.828 & 36.70 & \cellcolor{second}0.813 & 0.200 & \cellcolor{second}0.029 & \cellcolor{second}28.03 & 0.674 & 9.01 & 65.28\\
& GSNB   & 2.667 & 59.06 & 0.800 & 0.201 & 0.031 & 27.71 & 0.639 & 10.14\phantom{0} & 91.86\\
& NeST   & \cellcolor{best}\textbf{0.855} & 327.90\phantom{0} & 0.780 & \cellcolor{best}\textbf{0.177} & 0.033 & 26.95 & \cellcolor{second}0.410 & 9.49 & 13.90\\
& \textbf{Ours} & \cellcolor{second}1.361 & \cellcolor{second}28.83 & \cellcolor{best}\textbf{0.815} & 0.205 & \cellcolor{best}\textbf{0.028} & \cellcolor{best}\textbf{28.27} & \cellcolor{best}\textbf{0.267} & \cellcolor{best}\textbf{7.57} & \cellcolor{second}105.28\phantom{0}\\
\midrule
\multirow{6}{0.08\textwidth}{MipNeRF 360**}
& 3DGS   & 1.454 & \cellcolor{best}\textbf{23.50} & 0.923 & 0.138 & 0.016 & 31.01 & 0.347 & \cellcolor{second}8.02 & \cellcolor{best}\textbf{124.00\phantom{0}} \\
& VDGS   & 1.760 & 33.98 & \cellcolor{best}\textbf{0.926} & 0.136 & 0.016 & 31.42 & 0.436 & 8.83 & 76.25 \\
& GSNB   & 1.662 & 43.79 & \cellcolor{second}0.925 & \cellcolor{second}0.132 & 0.016 & \cellcolor{second}31.62 & 0.414 & 9.30 & 101.25\phantom{0} \\
& NeST   & \cellcolor{best}\textbf{0.358} & 374.78\phantom{0} & 0.910 & \cellcolor{best}\textbf{0.121} & 0.018 & 30.48 & 0.289 & 10.51\phantom{0} & 14.82 \\
& NTS   & 0.989 & 274.90\phantom{0} & 0.923 & 0.136 & 0.016 & 31.11 & \cellcolor{second}0.263 & 15.48\phantom{0} & \phantom{0}5.27 \\
& \textbf{Ours} & \cellcolor{second}0.641 & \cellcolor{second}24.77 & \cellcolor{best}\textbf{0.926} & 0.139 & \cellcolor{best}\textbf{0.015} & \cellcolor{best}\textbf{31.82} & \cellcolor{best}\textbf{0.126} & \cellcolor{best}\textbf{7.51} & \cellcolor{second}115.25\phantom{0} \\
\midrule
\multirow{5}{*}{DL3DV}
& 3DGS   & 1.158 & \cellcolor{best}\textbf{12.74} & 0.916 & 0.088 & 0.021 & 29.48 & \cellcolor{second}0.265 & \cellcolor{second}7.85 & \cellcolor{best}\textbf{130.21\phantom{0}}\\
& VDGS   & 1.404 & 21.15 & 0.918 & 0.089 & 0.021 & 29.92 & 0.335 & 8.65 & 86.28\\
& GSNB   & 1.032 & 25.28 & \cellcolor{second}0.920 & 0.086 & \cellcolor{second}0.020 & \cellcolor{second}30.60 & 0.308 & 9.05 & 112.10\phantom{0}\\
& NeST   & \cellcolor{best}\textbf{0.334} & 196.40\phantom{0} & 0.917 & \cellcolor{best}\textbf{0.072} & 0.021 & 29.81 & 0.281 & 9.96 & 24.78\\
& NTS  & 1.371 & 215.60\phantom{0} & 0.918 & \cellcolor{second}0.081 & 0.021 & 29.71 & 0.338 & 12.01\phantom{0} & \phantom{0}6.51\\
& \textbf{Ours} & \cellcolor{second}0.595 & \cellcolor{second}14.52 & \cellcolor{best}\textbf{0.925} & 0.089 & \cellcolor{best}\textbf{0.018} & \cellcolor{best}\textbf{30.82} & \cellcolor{best}\textbf{0.117} & \cellcolor{best}\textbf{7.41} & \cellcolor{second}122.28\phantom{0} \\
\midrule
\multirow{4}{*}{TnT}
& 3DGS   & 1.868 & \cellcolor{best}\textbf{15.35} & 0.831 & 0.165 & 0.046 & 23.85 & 0.428 & \cellcolor{second}8.53 & \cellcolor{best}\textbf{122.73\phantom{0}}\\
& VDGS   & 2.113 & 26.37 & \cellcolor{second}0.834 & \cellcolor{best}\textbf{0.163} & \cellcolor{second}0.045 & \cellcolor{second}24.12 & 0.503 & 8.85 & 78.33\\
& GSNB   & 1.705 & 35.75 & 0.830 & \cellcolor{second}0.164 & 0.046 & 23.92 & 0.489 & 9.25 & 101.27\phantom{0}\\
& NeST   & \cellcolor{best}\textbf{0.616} & 182.39\phantom{0} & 0.804 & 0.169 & 0.049 & 22.79 & \cellcolor{second}0.347 & 10.37\phantom{0} & 24.98\\
& NTS   & 1.999 & 221.72\phantom{0} & 0.817 & 0.177 & 0.050 & 23.18 & 0.377 & 13.49\phantom{0} & \phantom{0}3.72\\
& \textbf{Ours} & \cellcolor{second}1.040 & \cellcolor{second}19.91 & \cellcolor{best}\textbf{0.845} & \cellcolor{second}0.164 & \cellcolor{best}\textbf{0.040} & \cellcolor{best}\textbf{24.68} & \cellcolor{best}\textbf{0.204} & \cellcolor{best}\textbf{7.48} & \cellcolor{second}115.94\phantom{0}\\
\bottomrule
\vspace{-2.0em}
\end{tabular}

}
\end{table*}
\fi

\subsubsection{Results}
\label{sec:results}

We report average splat count, training time, and standard metrics (1:8 split) in Tab.~\ref{tab:main_results}. Experiments are run on a single RTX 3090 (24GB) with a 24-core CPU and 64GB RAM. NRF-GS consistently yields the most compact representations while slightly improving quality over all baselines~\cite{kerbl3Dgaussians, malarz2025gaussian, zhou2024gaussian}.

\textbf{Parameter comparison.} 3DGS with degree-3 SH~\cite{muller2006spherical} stores 48 appearance coefficients per Gaussian. NRF-GS instead uses a 32 latent, base color, and opacity (36 parameters), plus two small shared decoders ($\sim$6.7k parameters total). For large scenes (e.g., 1M Gaussians), this reduces per-splat appearance parameters by $\sim$25\% ($48$M vs $36$M $+ 6.7$k).

\textbf{Splat count.} A key distinction of NRF-GS lies in the number of Gaussian splats required to represent the scene. Across all datasets, NRF-GS consistently converges to fewer splat count, reducing the number of splats by approximately \textbf{40--55\%}; see Fig.~\ref{fig:point_growth}, which compares point growth over training iterations for a representative scene. Despite using the \emph{same densification and pruning heuristics} as classical 3DGS, NRF-GS stabilizes at a lower point count, suggesting that each Gaussian captures richer appearance variation and reduces the need for redundant splats.

\begin{figure}[t]
\centering
\resizebox{\textwidth}{!}{
\setlength{\tabcolsep}{0.5pt}

\begin{tabular}{c c c c c c c}

& \textbf{Kitchen}
& \textbf{Room}
& \textbf{Palace}
& \textbf{Panther}
& \textbf{387e}
& \textbf{06da} \\

\raisebox{1.5\height}{\rotatebox[origin=c]{90}{\textbf{3DGS}}} &
\includegraphics[width=0.15\textwidth]{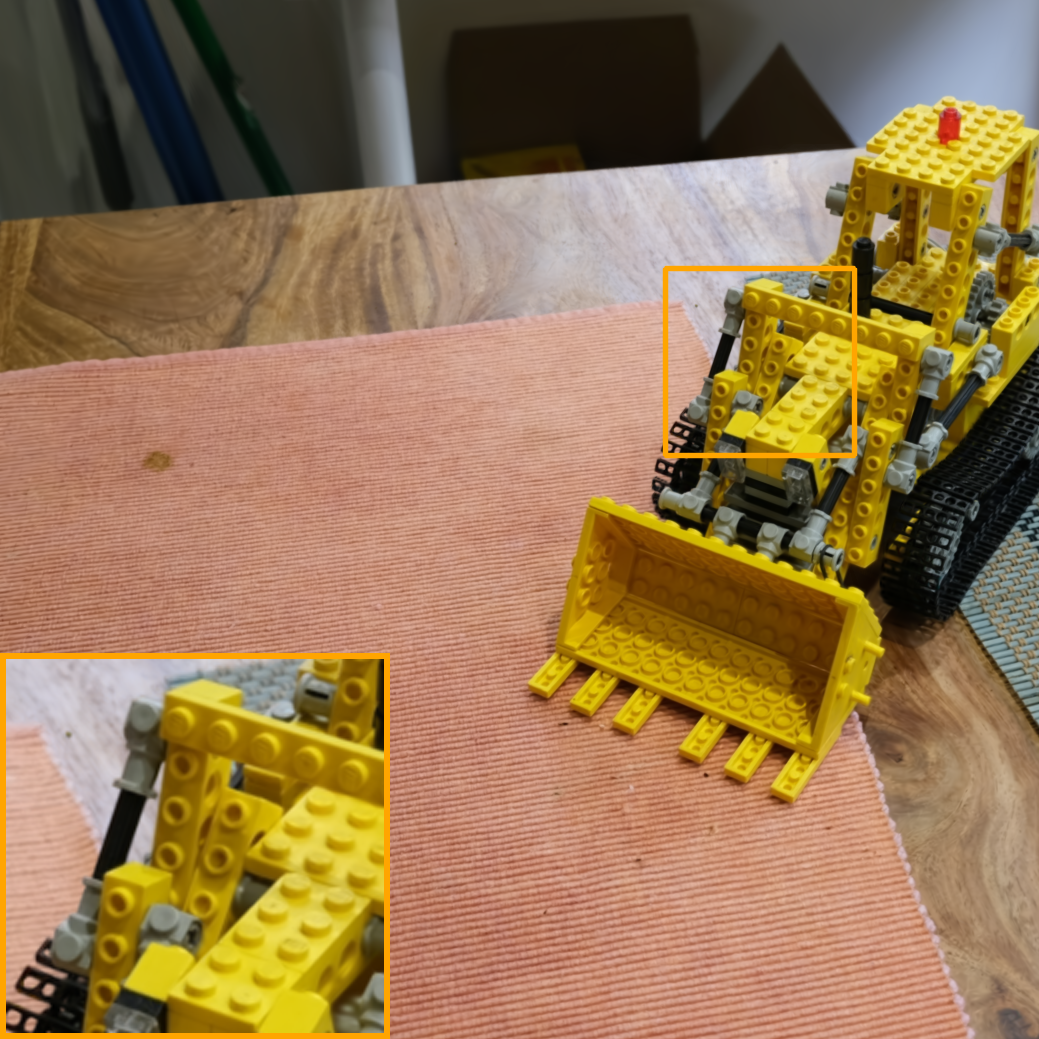} &
\includegraphics[width=0.15\textwidth]{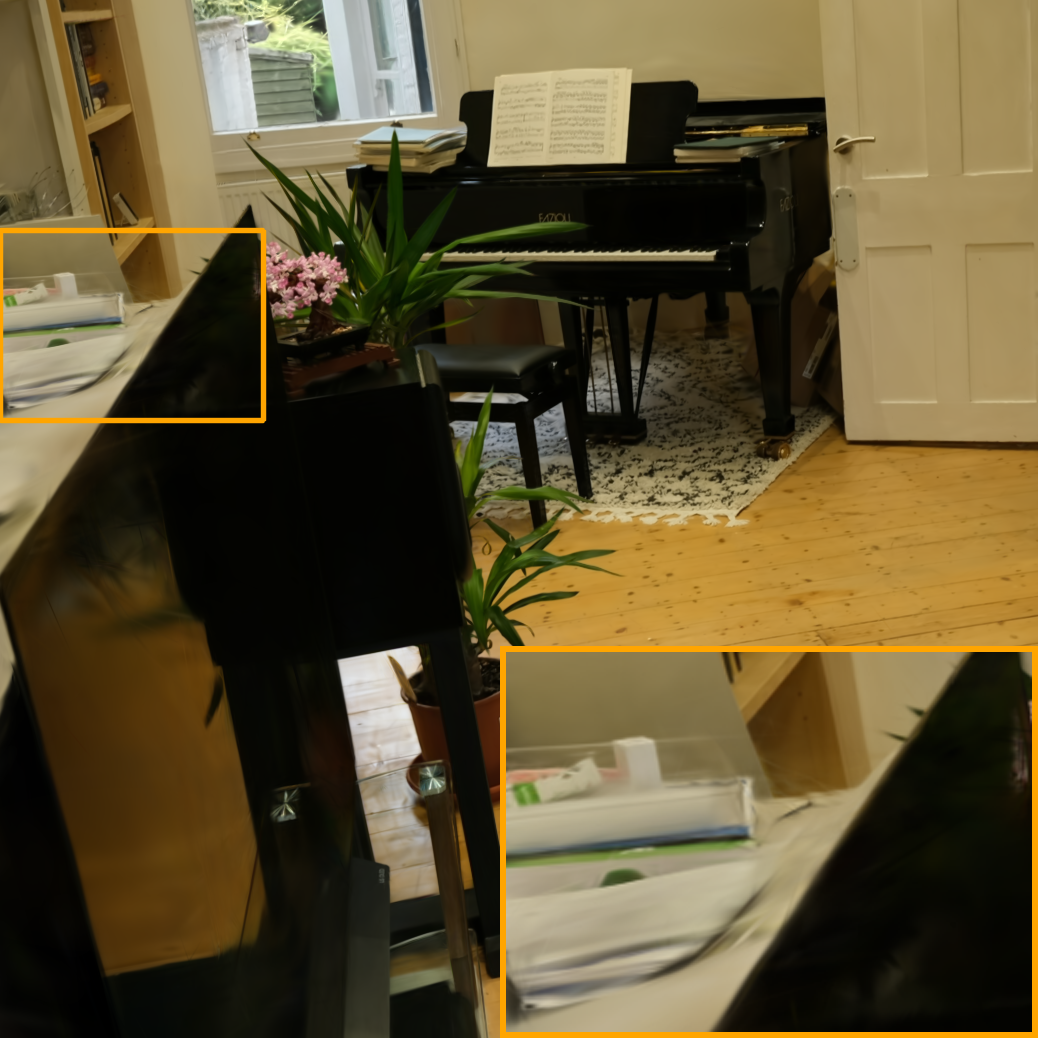} &
\includegraphics[width=0.15\textwidth]{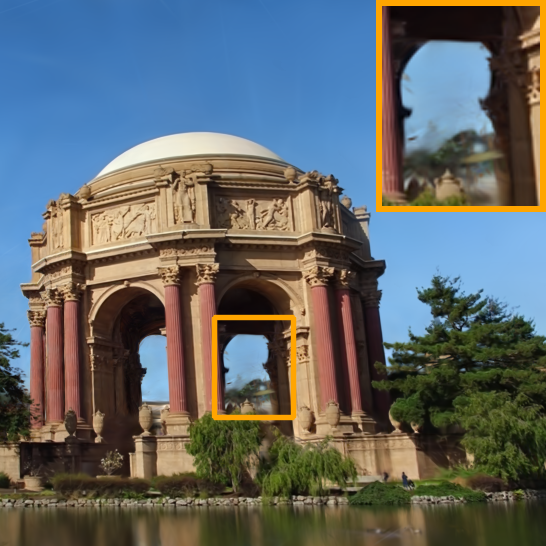} &
\includegraphics[width=0.15\textwidth]{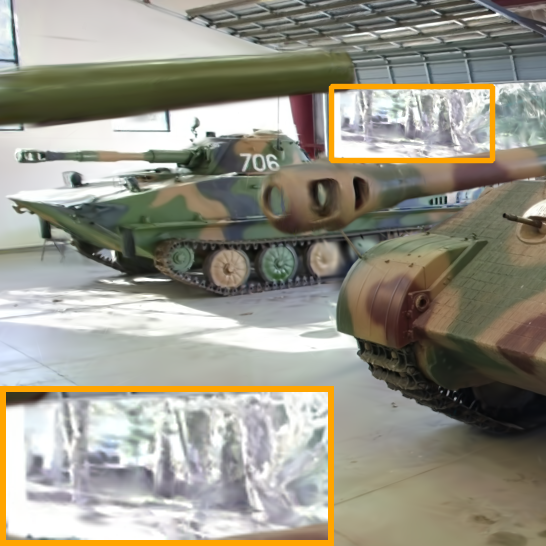} &
\includegraphics[width=0.15\textwidth]{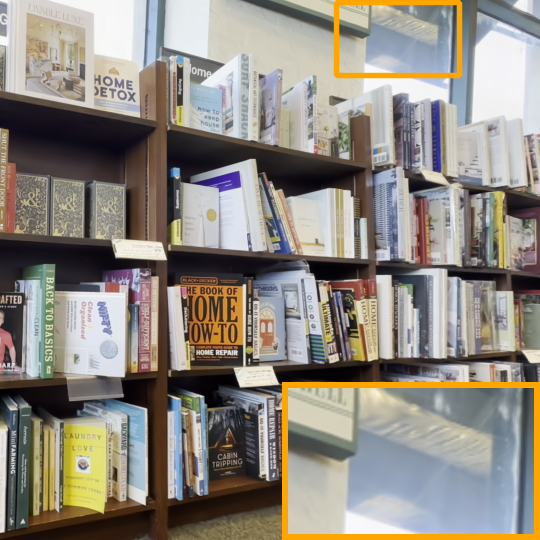} &
\includegraphics[width=0.15\textwidth]{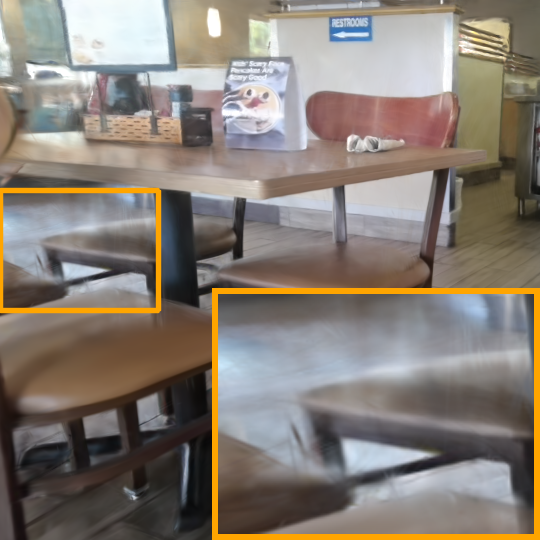} \\

\raisebox{1.5\height}{\rotatebox[origin=c]{90}{\textbf{VDGS}}} &
\includegraphics[width=0.15\textwidth]{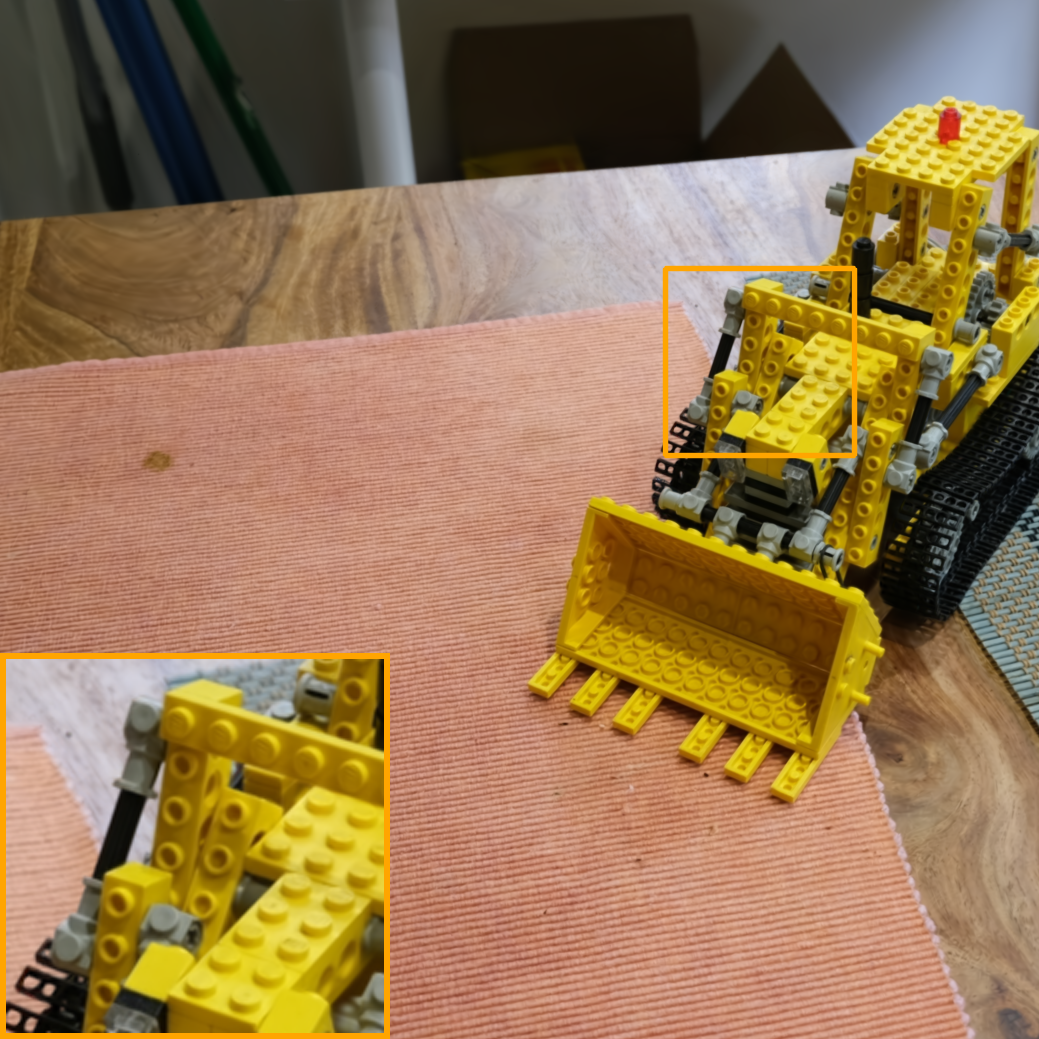} &
\includegraphics[width=0.15\textwidth]{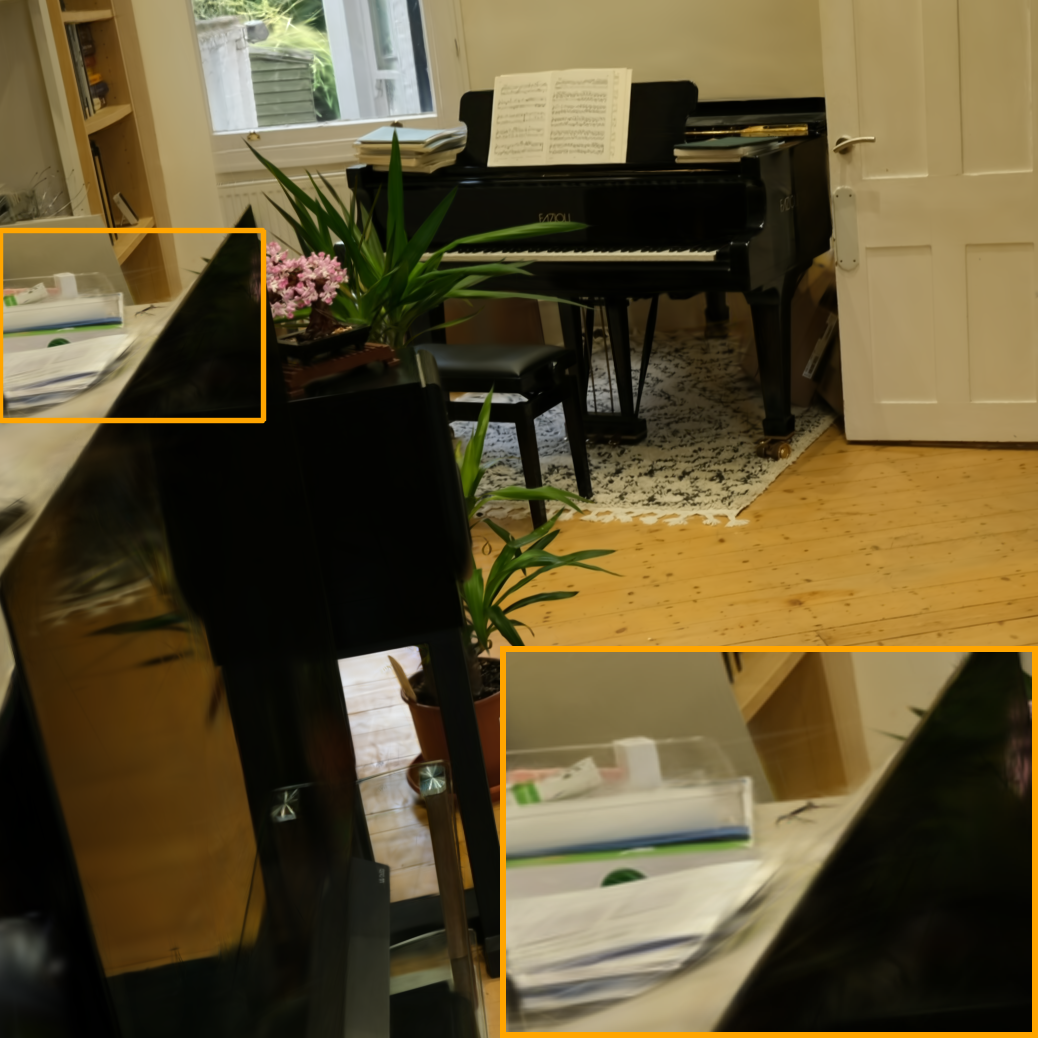} &
\includegraphics[width=0.15\textwidth]{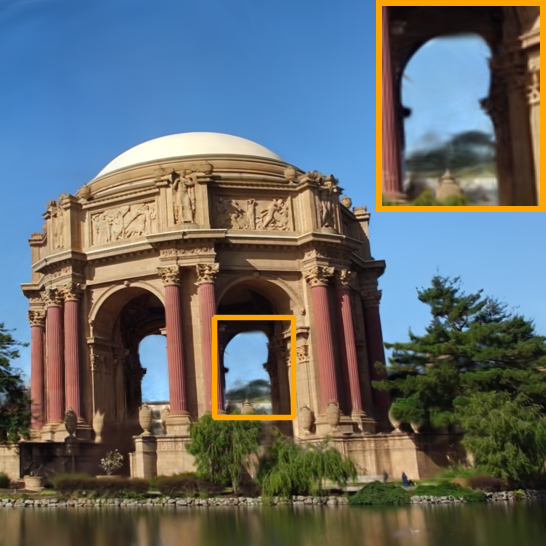} &
\includegraphics[width=0.15\textwidth]{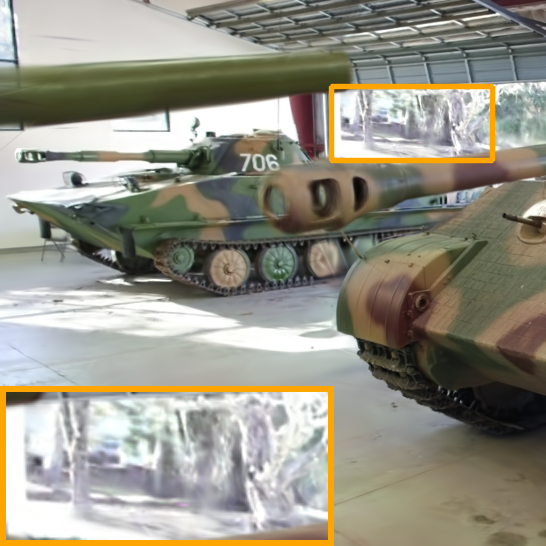} &
\includegraphics[width=0.15\textwidth]{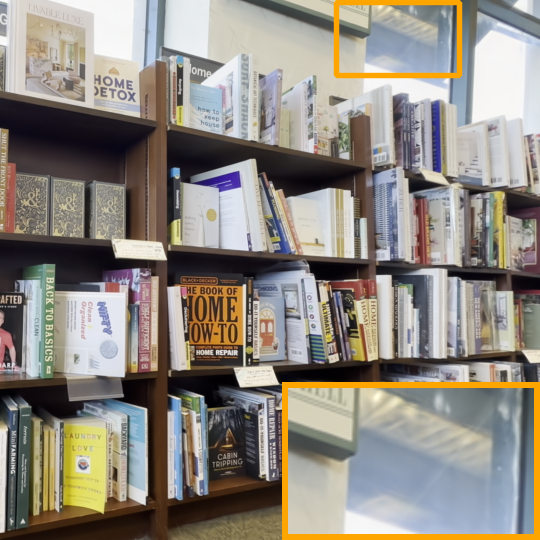} &
\includegraphics[width=0.15\textwidth]{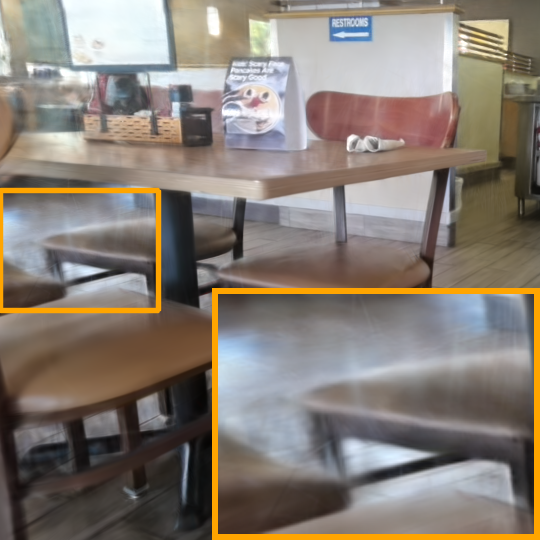} \\

\raisebox{1.5\height}{\rotatebox[origin=c]{90}{\textbf{GSNB}}} &
\includegraphics[width=0.15\textwidth]{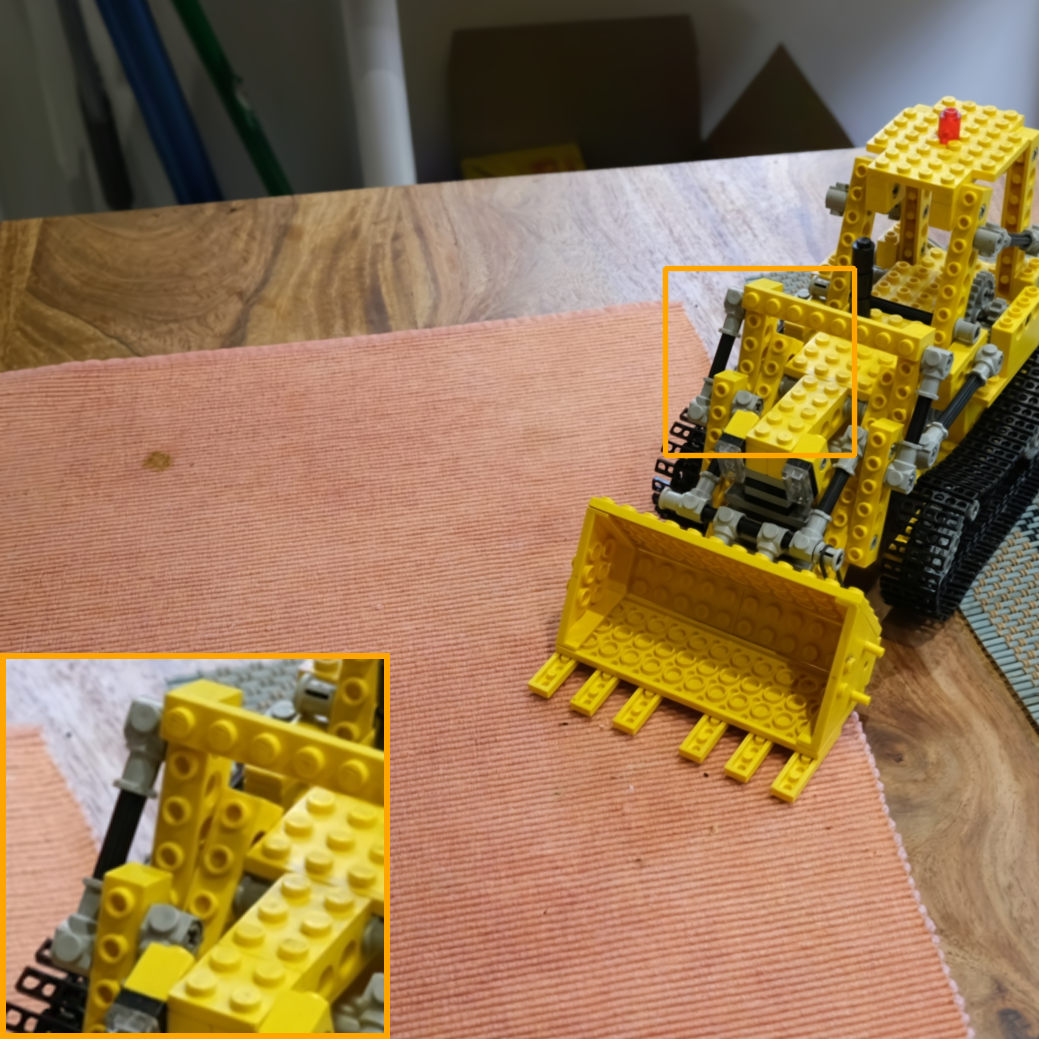} &
\includegraphics[width=0.15\textwidth]{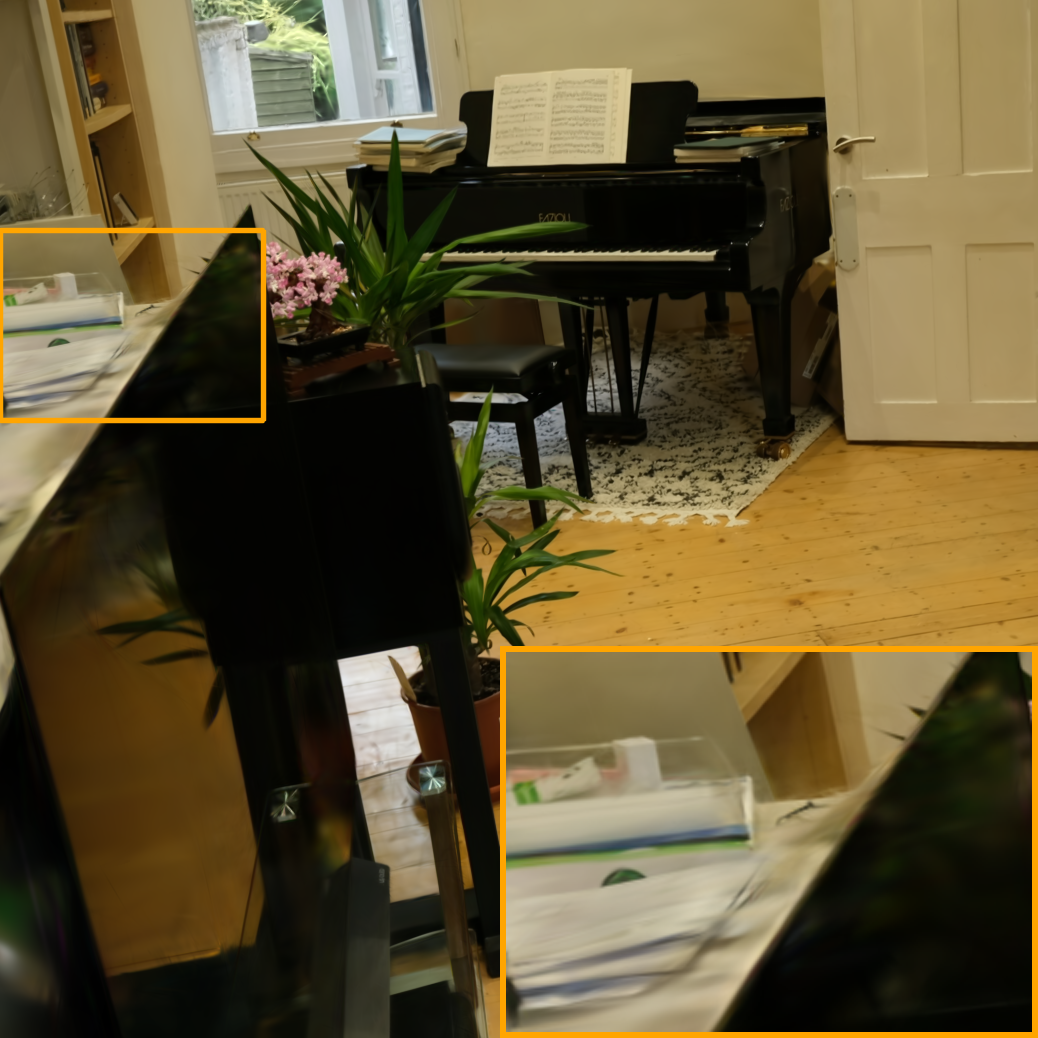} &
\includegraphics[width=0.15\textwidth]{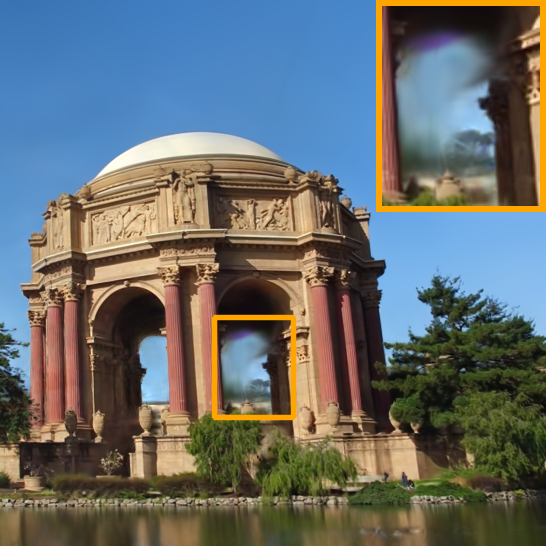} &
\includegraphics[width=0.15\textwidth]{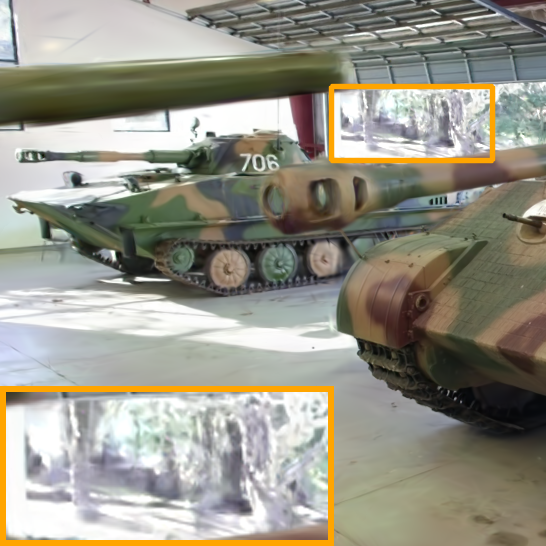} &
\includegraphics[width=0.15\textwidth]{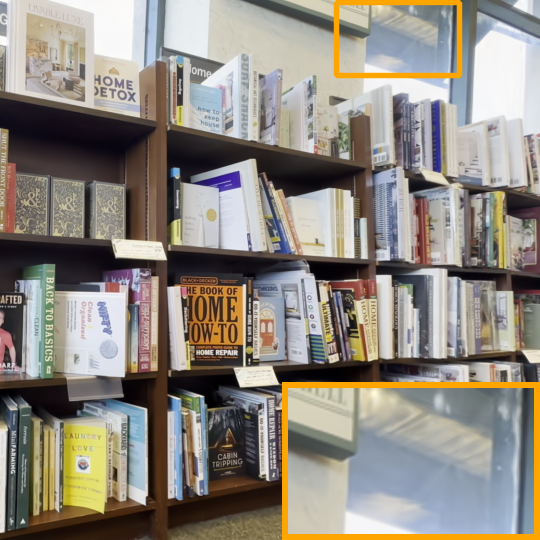} &
\includegraphics[width=0.15\textwidth]{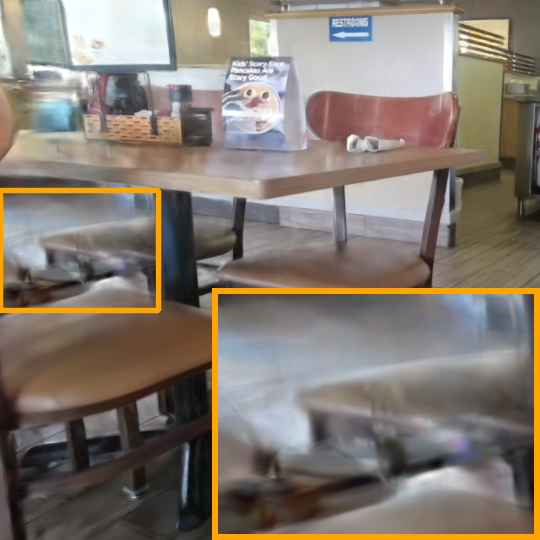} \\

\raisebox{1.5\height}{\rotatebox[origin=c]{90}{\textbf{NeST}}} &
\includegraphics[width=0.15\textwidth]{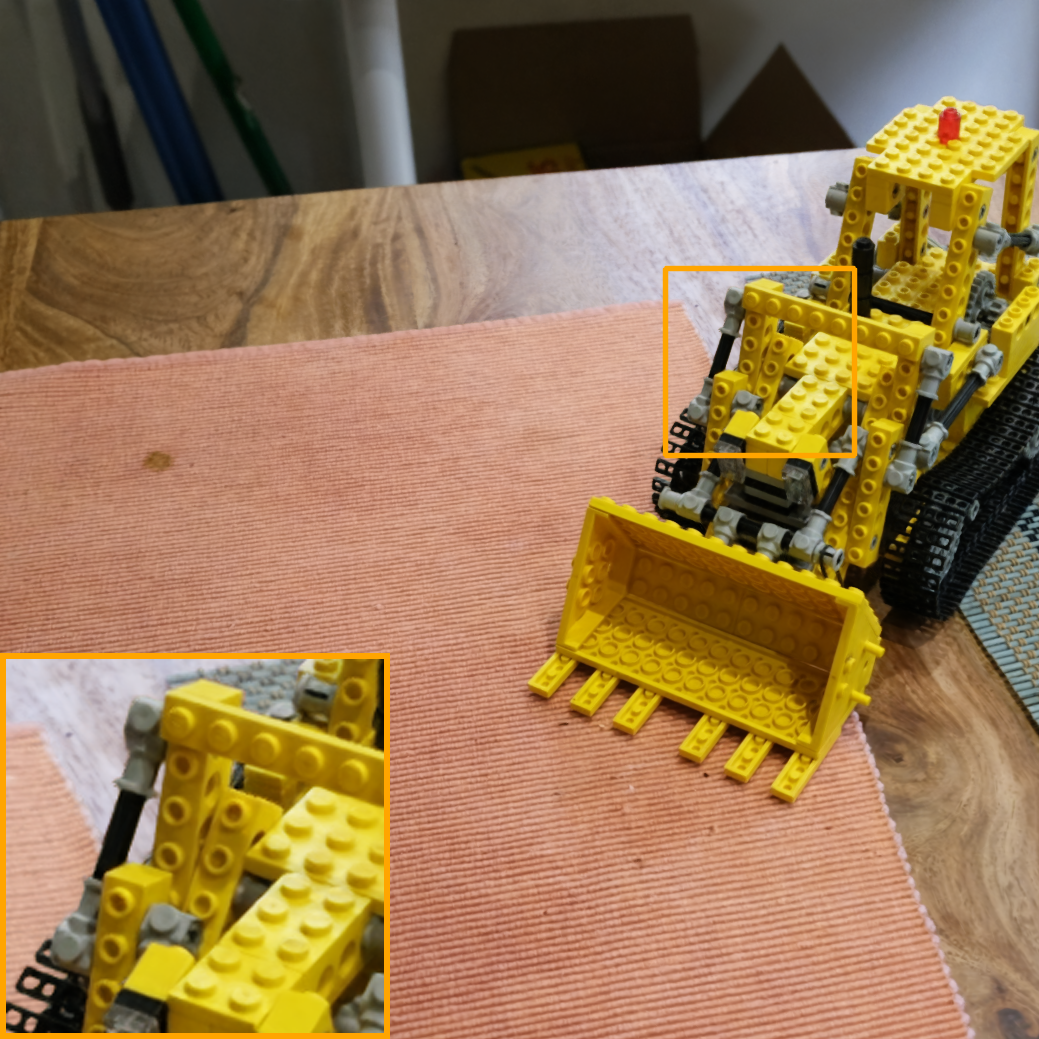} &
\includegraphics[width=0.15\textwidth]{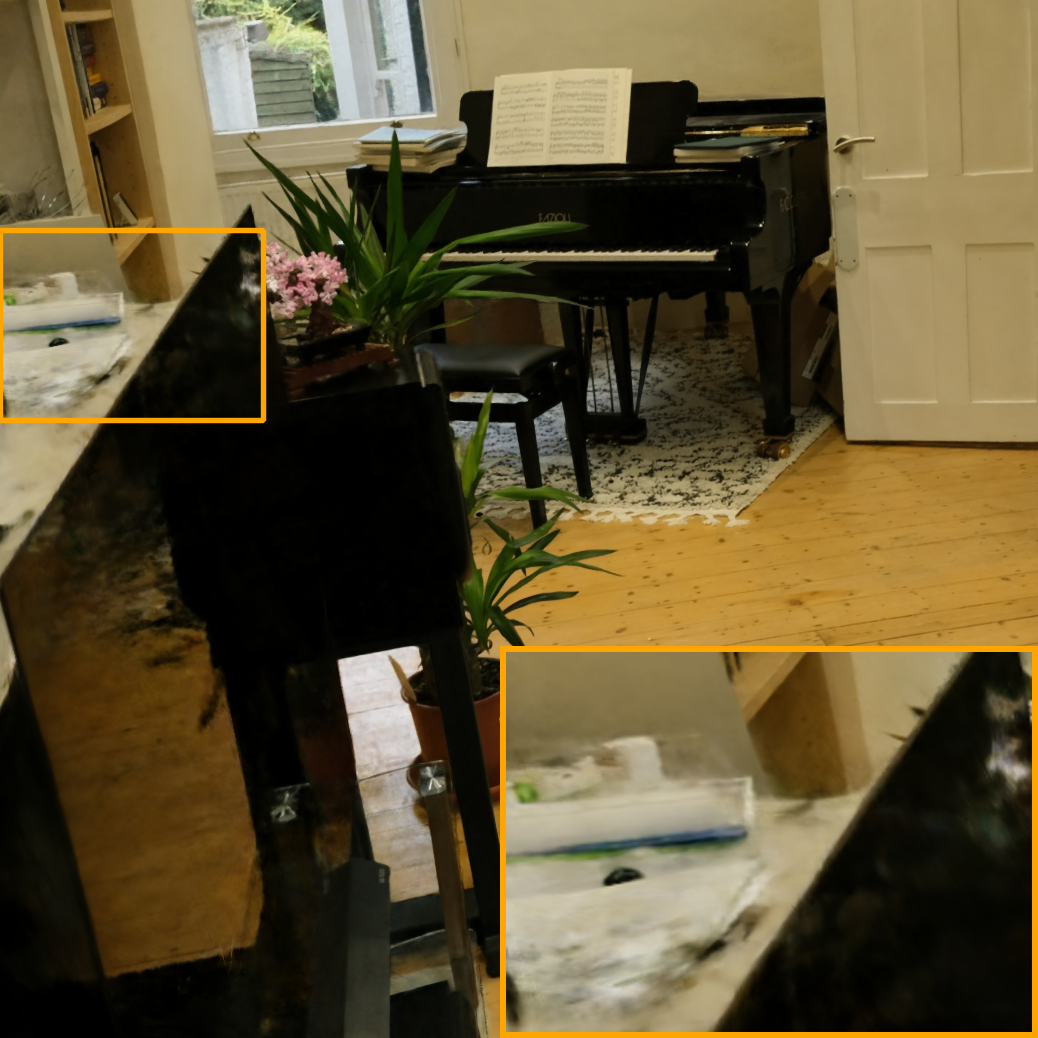} &
\includegraphics[width=0.15\textwidth]{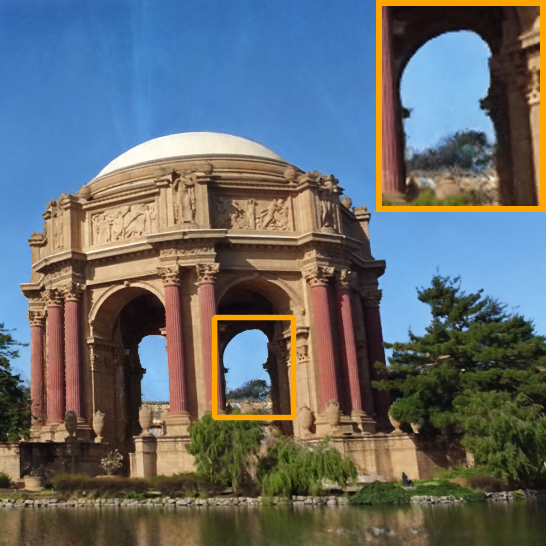} &
\includegraphics[width=0.15\textwidth]{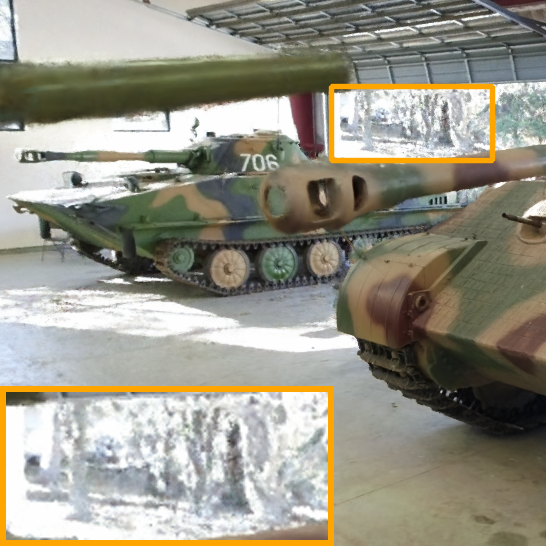} &
\includegraphics[width=0.15\textwidth]{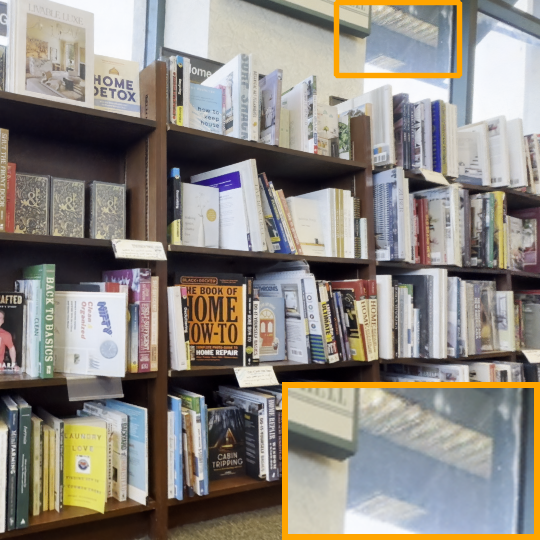} &
\includegraphics[width=0.15\textwidth]{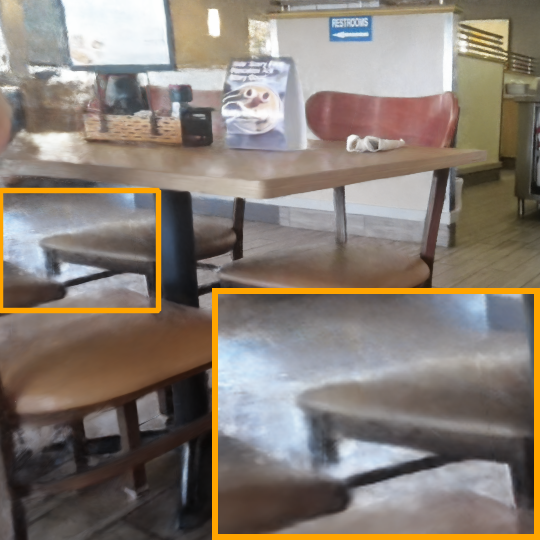} \\

\raisebox{1.5\height}{\rotatebox[origin=c]{90}{\textbf{NTS}}} &
\includegraphics[width=0.15\textwidth]{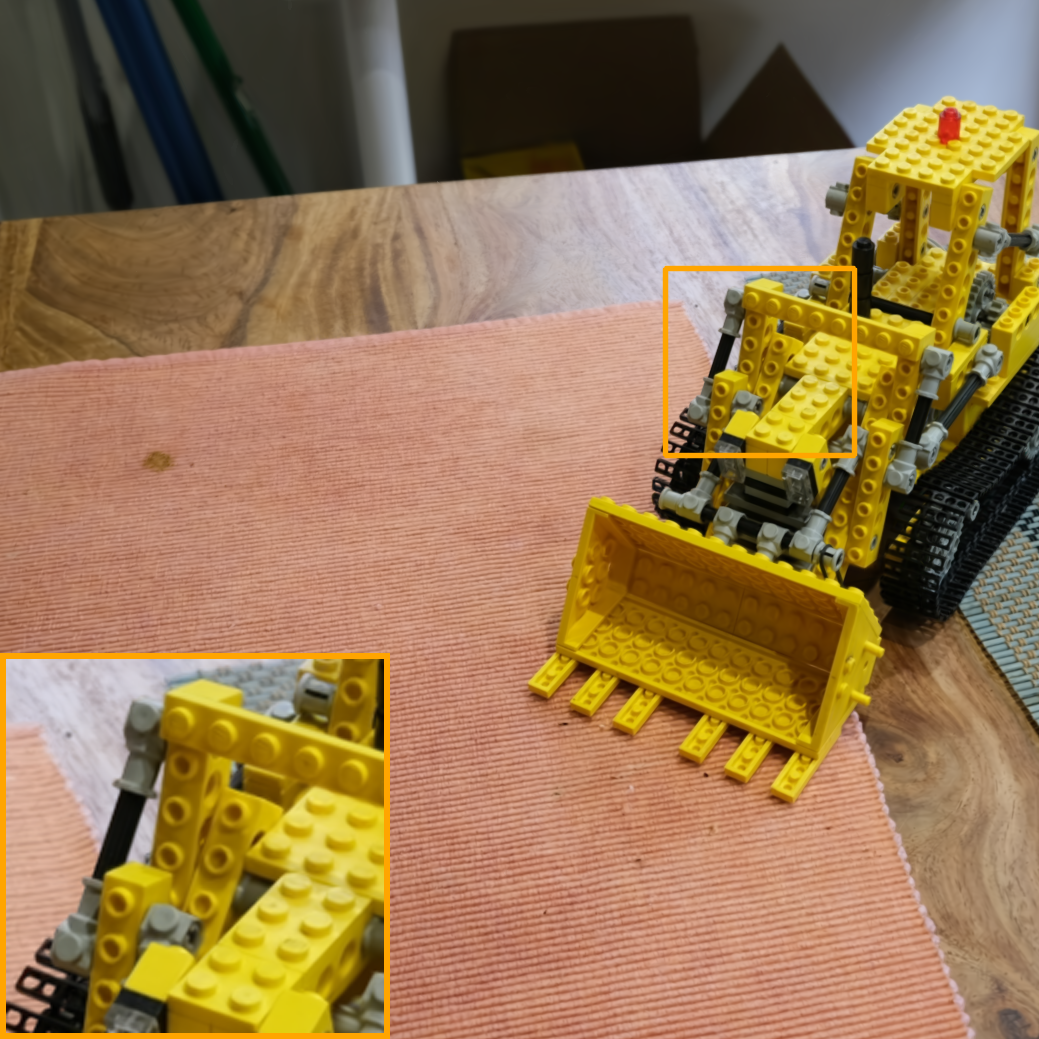} &
\includegraphics[width=0.15\textwidth]{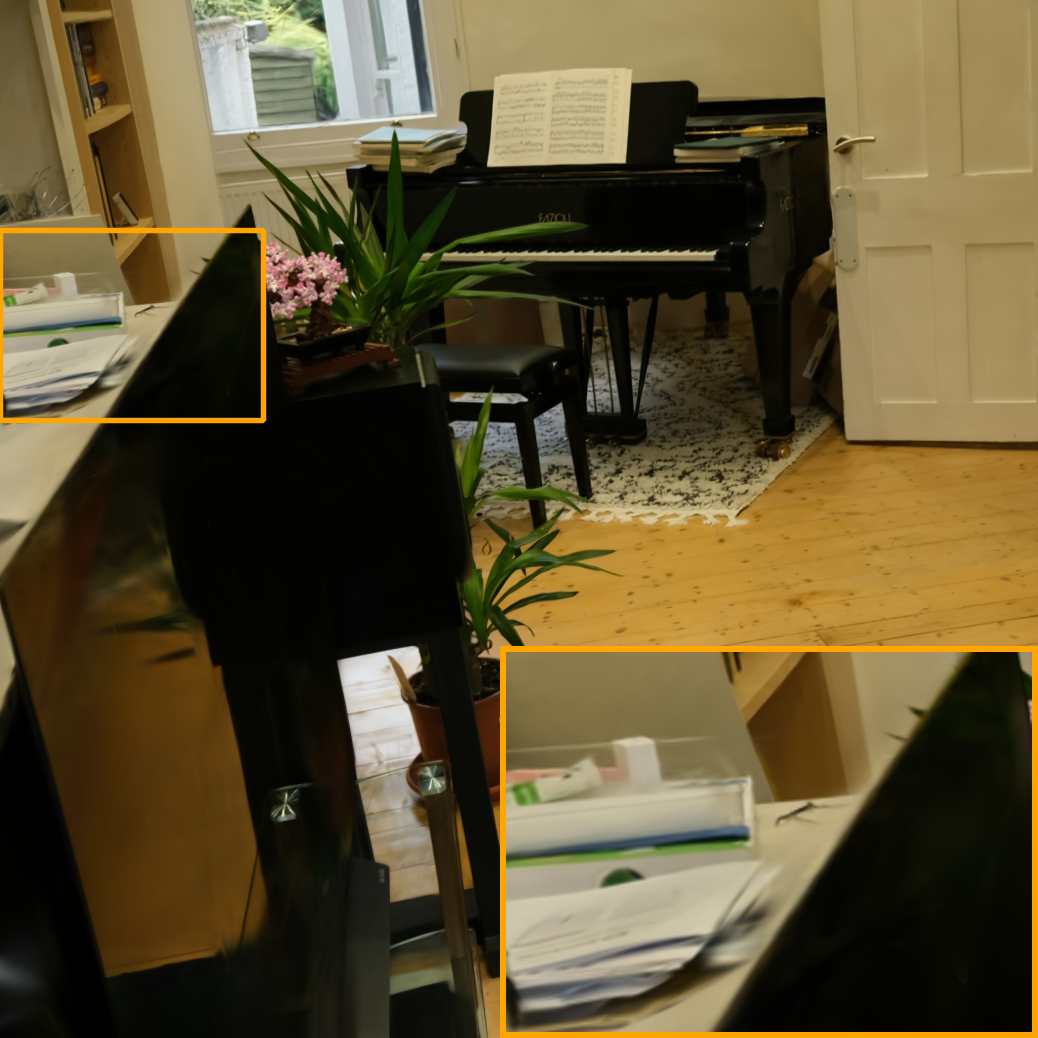} &
\includegraphics[width=0.15\textwidth]{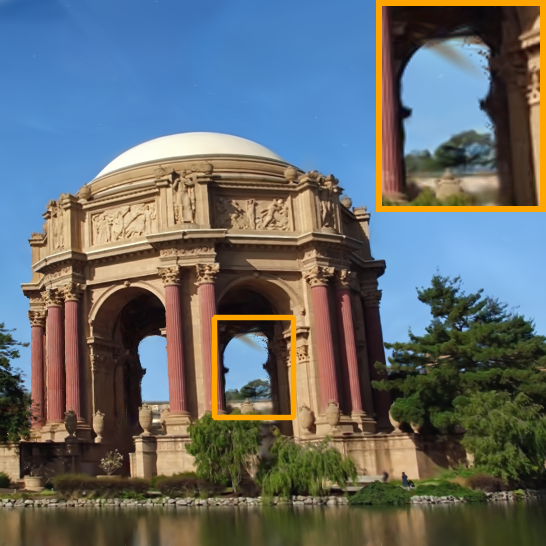} &
\includegraphics[width=0.15\textwidth]{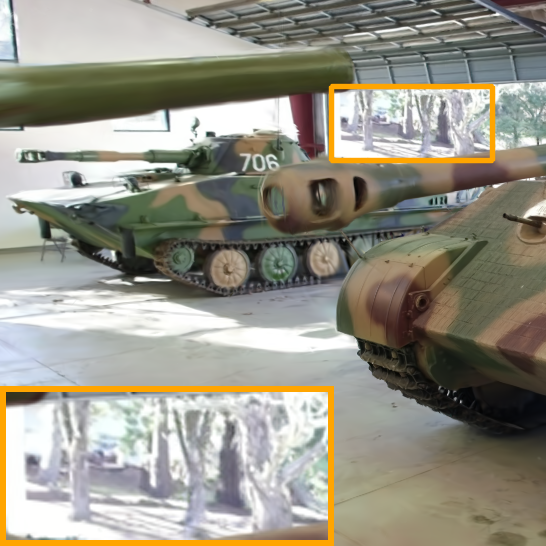} &
\includegraphics[width=0.15\textwidth]{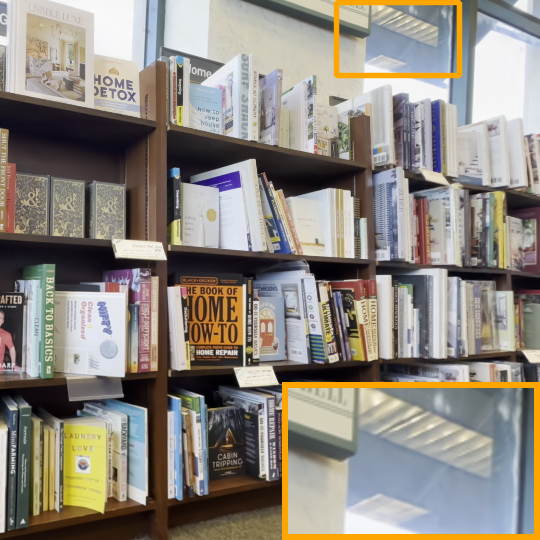} &
\includegraphics[width=0.15\textwidth]{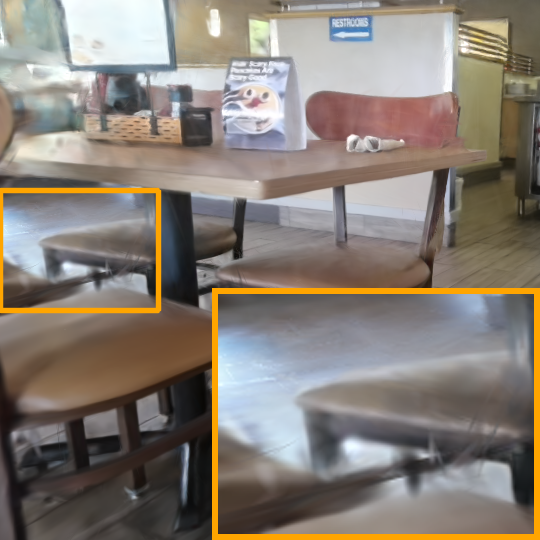} \\

\raisebox{1.5\height}{\rotatebox[origin=c]{90}{\textbf{Ours}}} &
\includegraphics[width=0.15\textwidth]{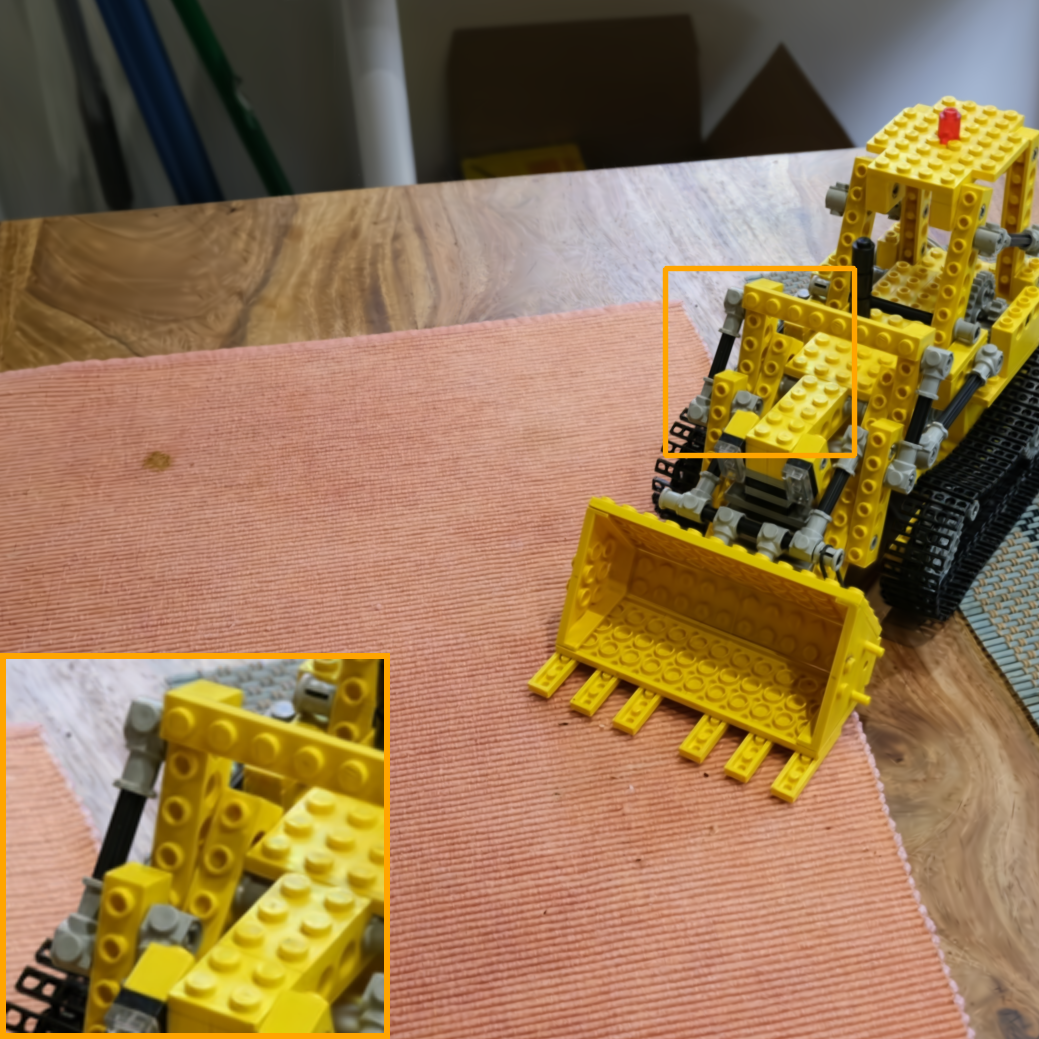} &
\includegraphics[width=0.15\textwidth]{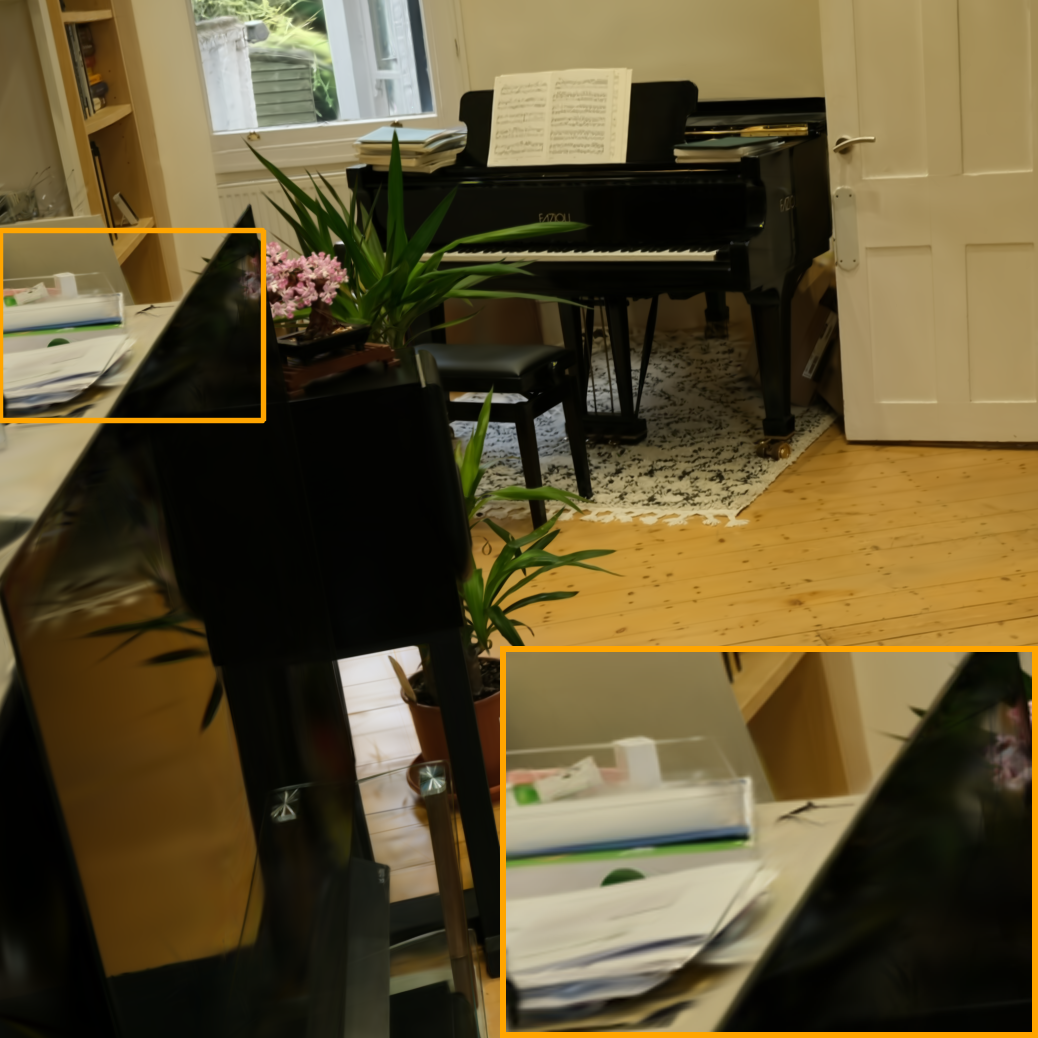} &
\includegraphics[width=0.15\textwidth]{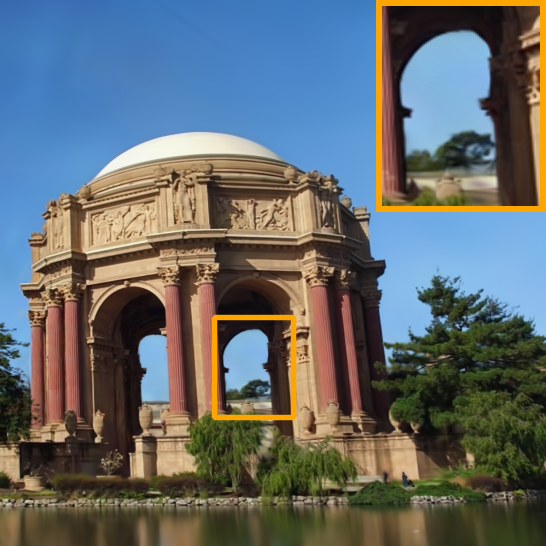} &
\includegraphics[width=0.15\textwidth]{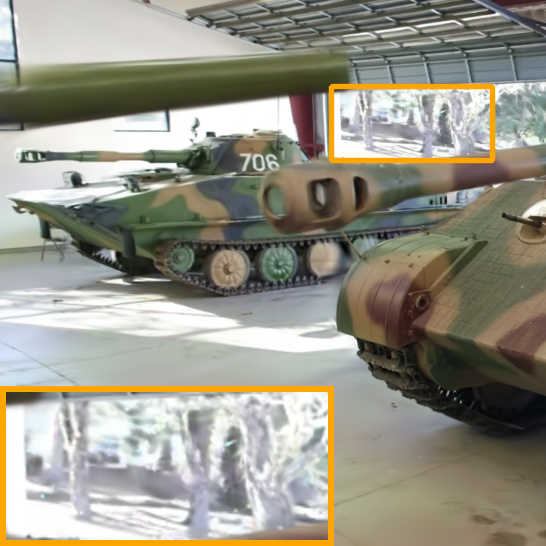} &
\includegraphics[width=0.15\textwidth]{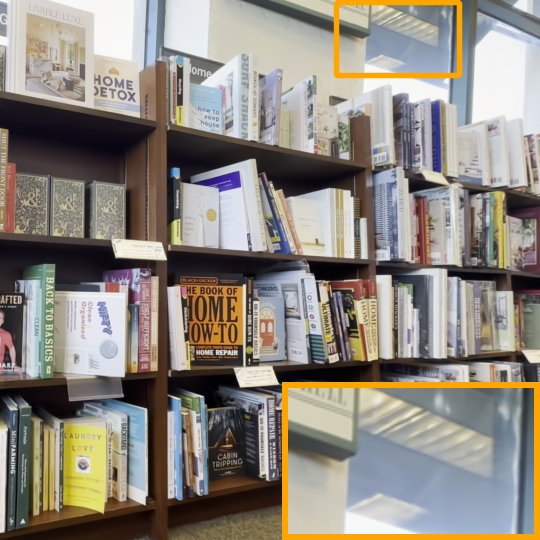} &
\includegraphics[width=0.15\textwidth]{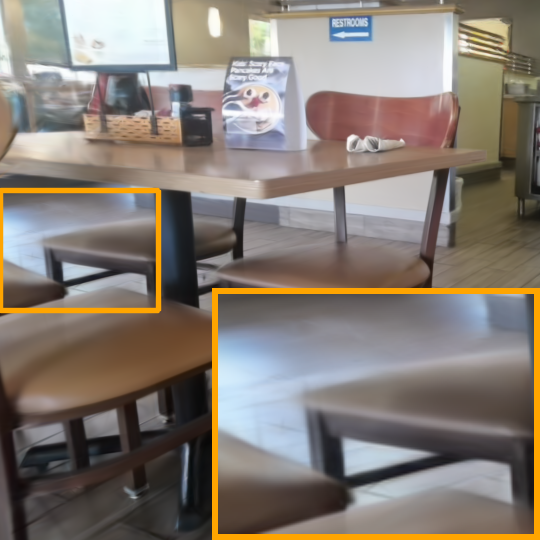} \\

\raisebox{2.5\height}{\rotatebox[origin=c]{90}{\textbf{GT}}} &
\includegraphics[width=0.15\textwidth]{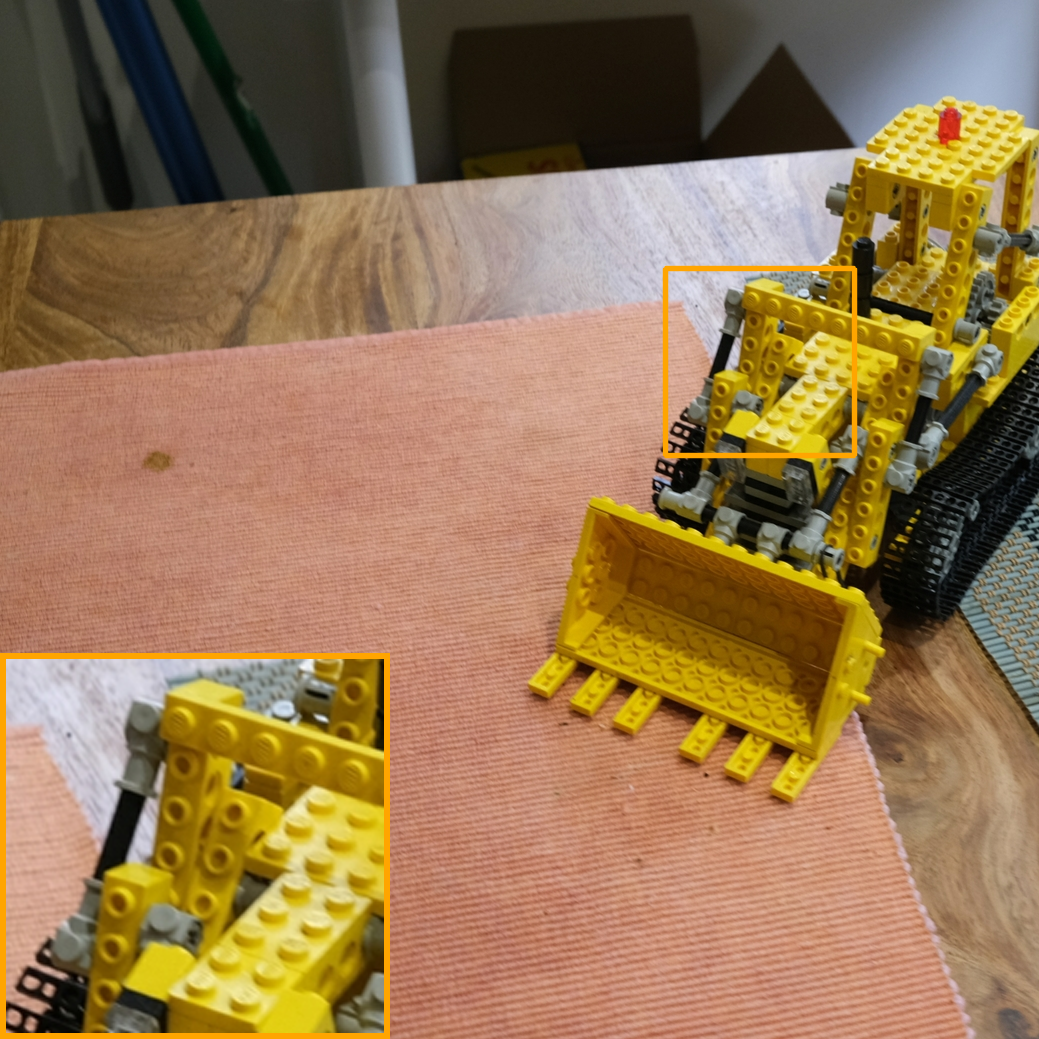} &
\includegraphics[width=0.15\textwidth]{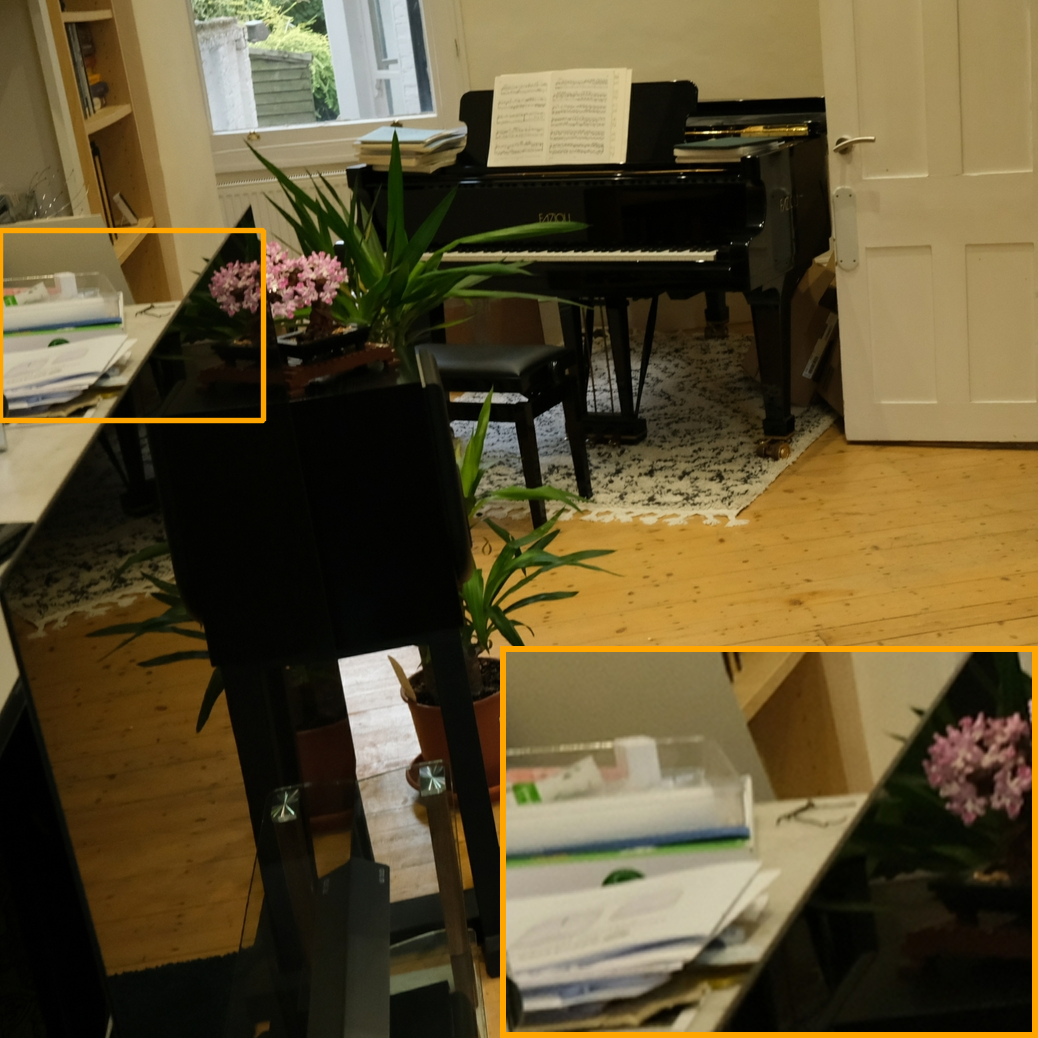} &
\includegraphics[width=0.15\textwidth]{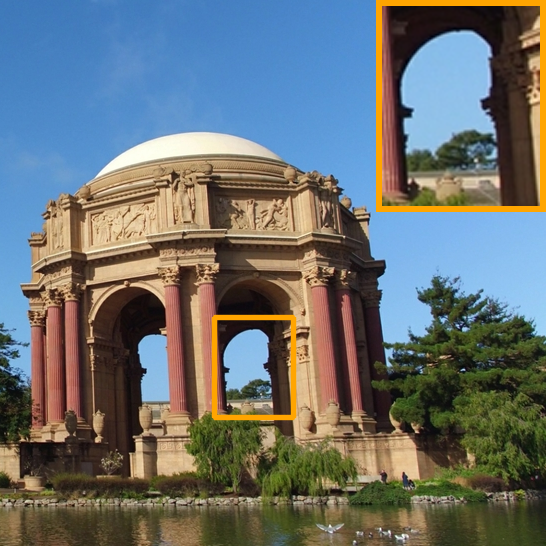} &
\includegraphics[width=0.15\textwidth]{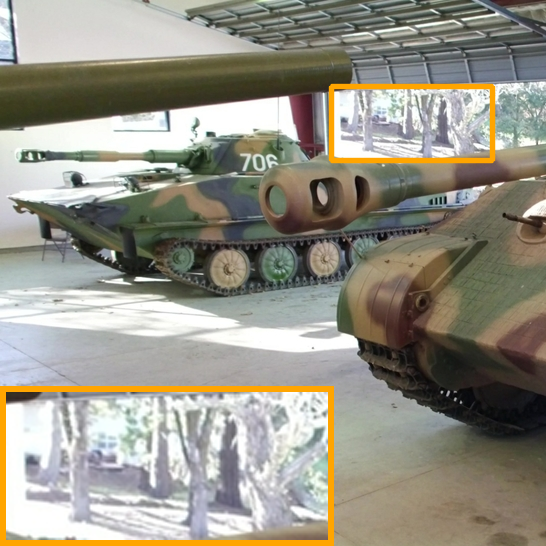} &
\includegraphics[width=0.15\textwidth]{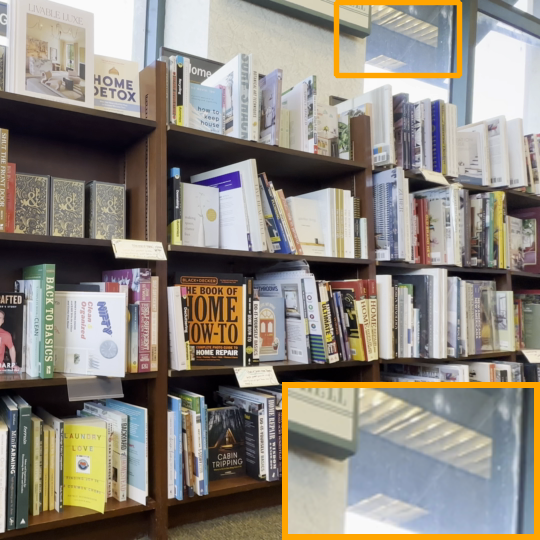} &
\includegraphics[width=0.15\textwidth]{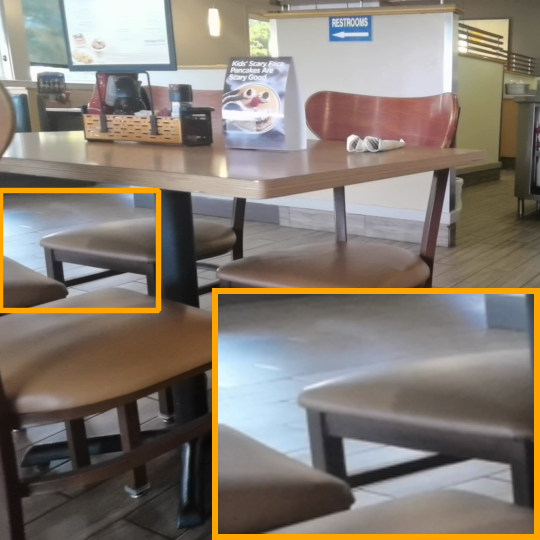} \\

\end{tabular}
}
\caption{Qualitative comparison across multiple scenes. MipNeRF360 - scenes 'kitchen' and 'room', TnT - scenes 'palace' and 'panther', DL3DV - scenes '387e' and '06da'.}
\label{fig:qual_grid}
\vspace{-1.0em}
\end{figure}

\if 10
\begin{figure}[t]
\centering
\resizebox{\textwidth}{!}{
\setlength{\tabcolsep}{0.05pt}

\begin{tabular}{c c c c c}
\textbf{GT} & \textbf{3DGS} & \textbf{VDGS} & \textbf{GSNB} & \textbf{NRF-GS (Ours)} \\

\includegraphics[width=0.19\textwidth]{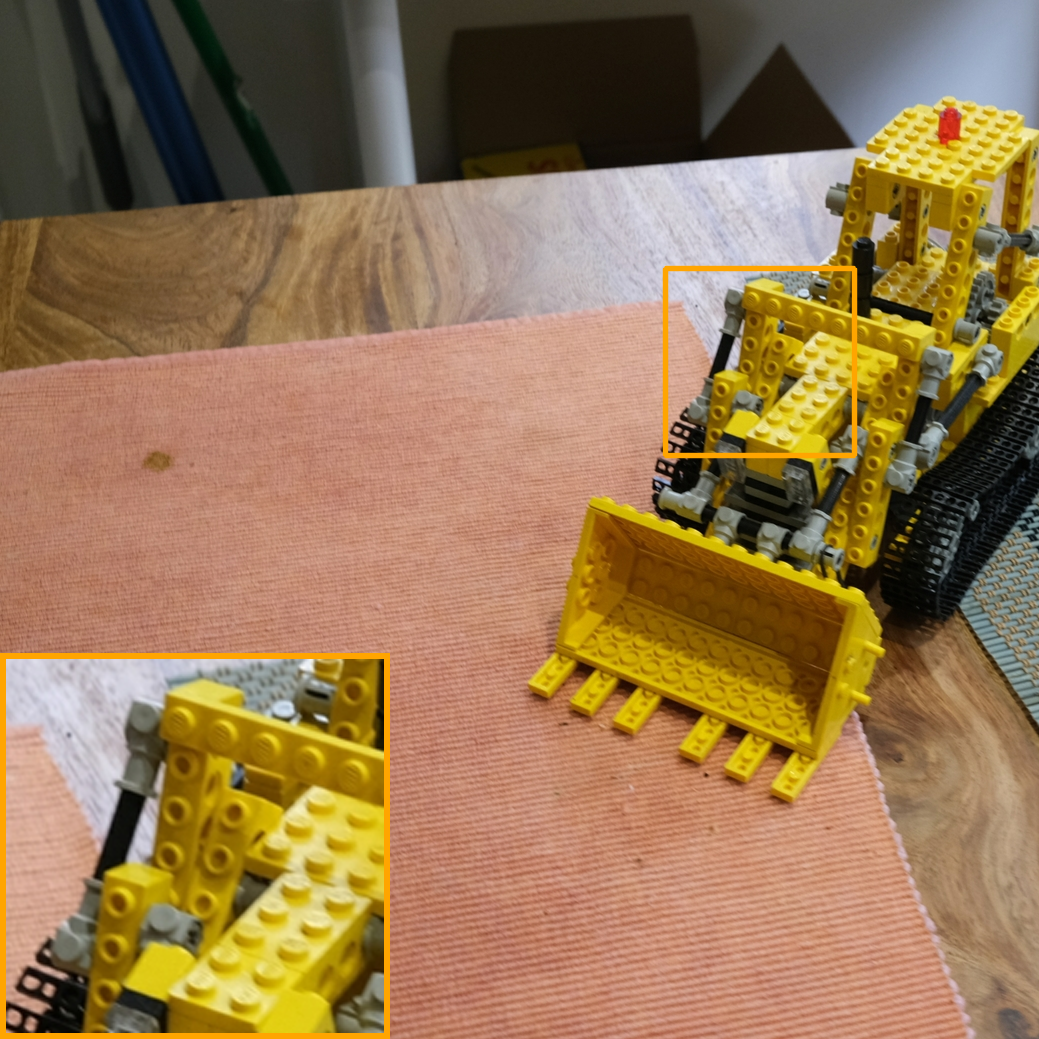} &
\includegraphics[width=0.19\textwidth]{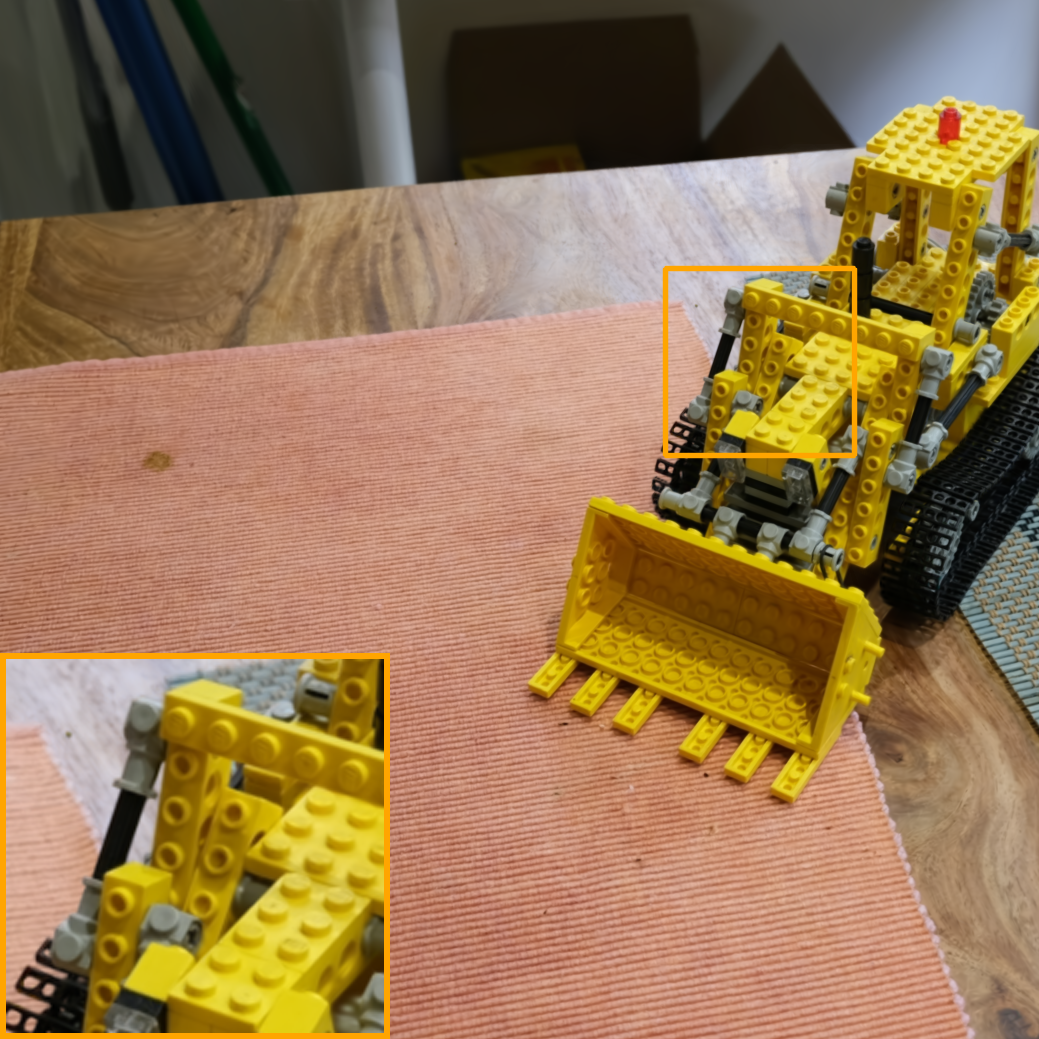} &
\includegraphics[width=0.19\textwidth]{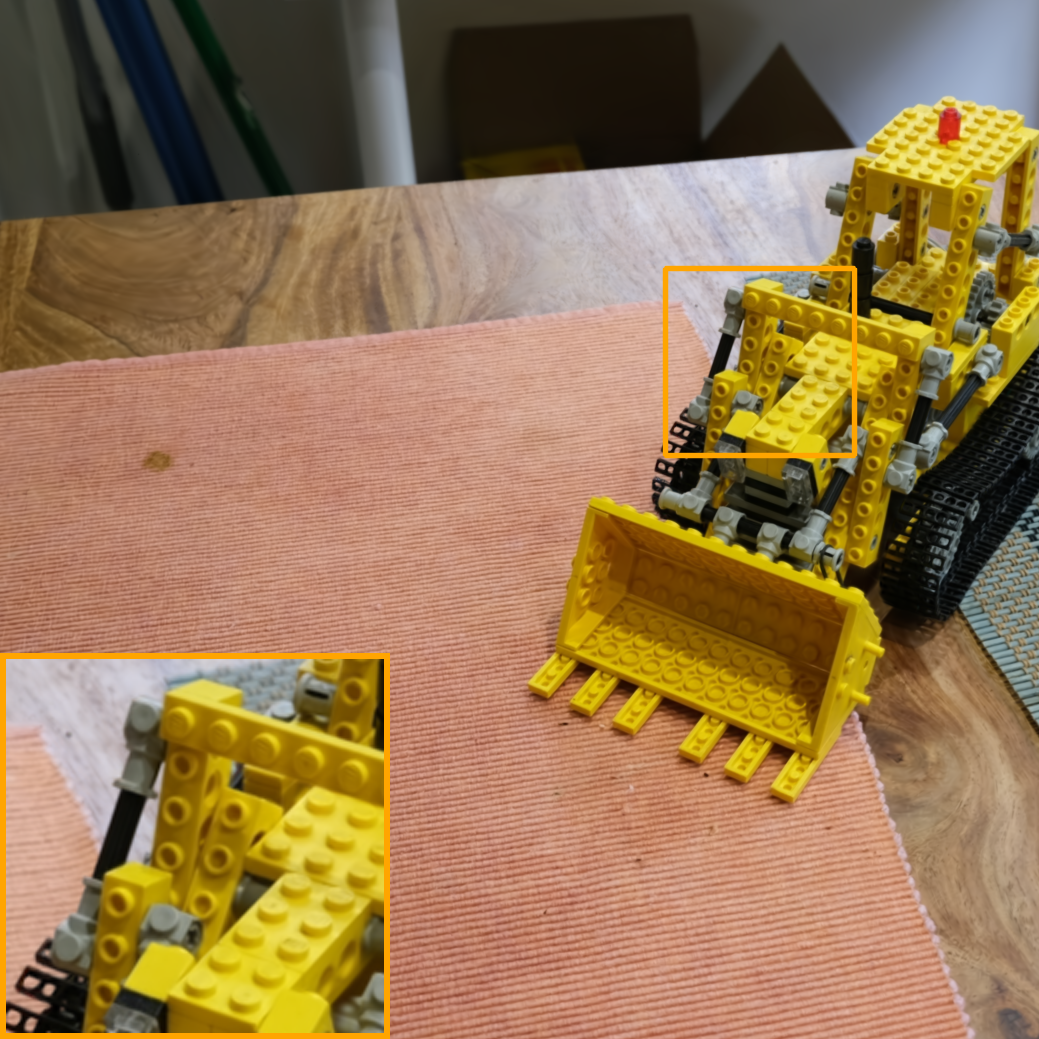} &
\includegraphics[width=0.19\textwidth]{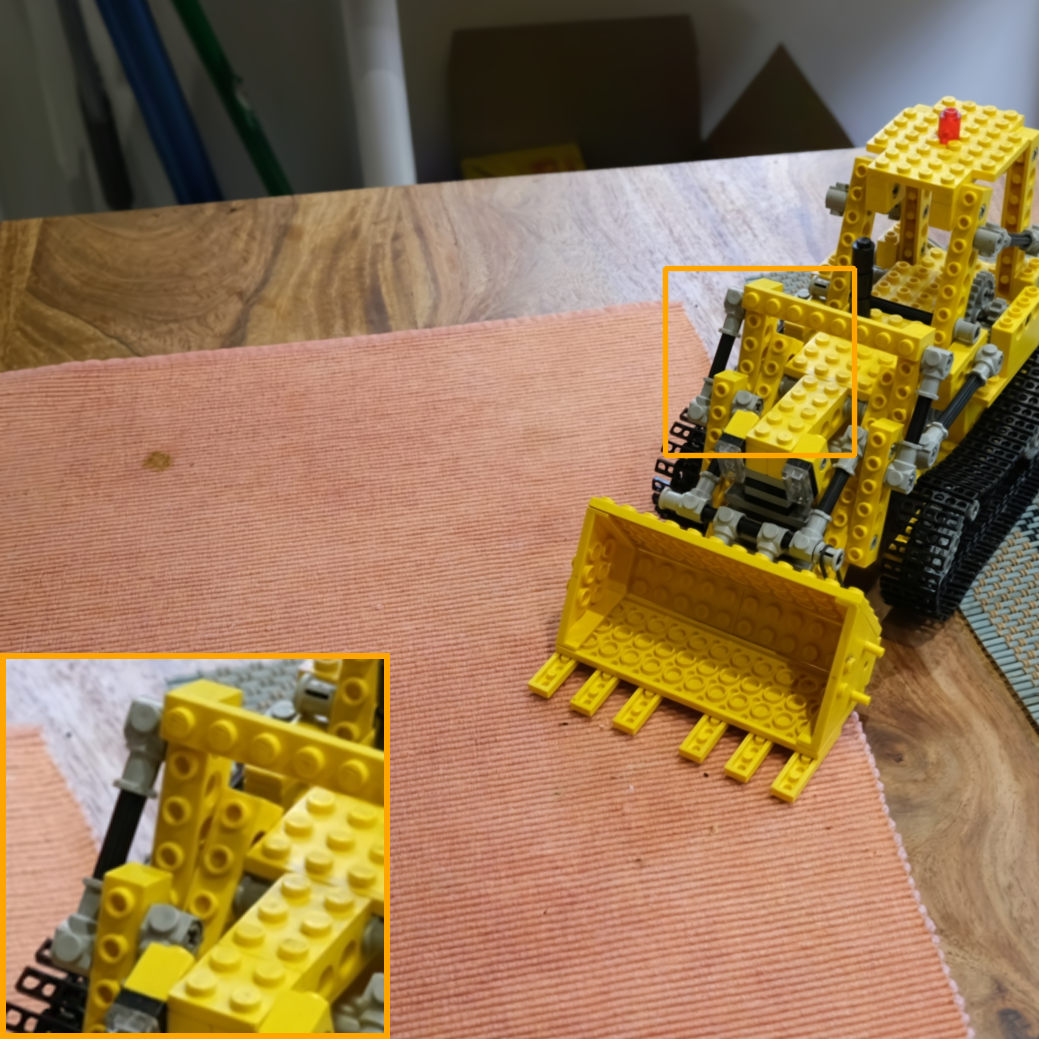} &
\includegraphics[width=0.19\textwidth]{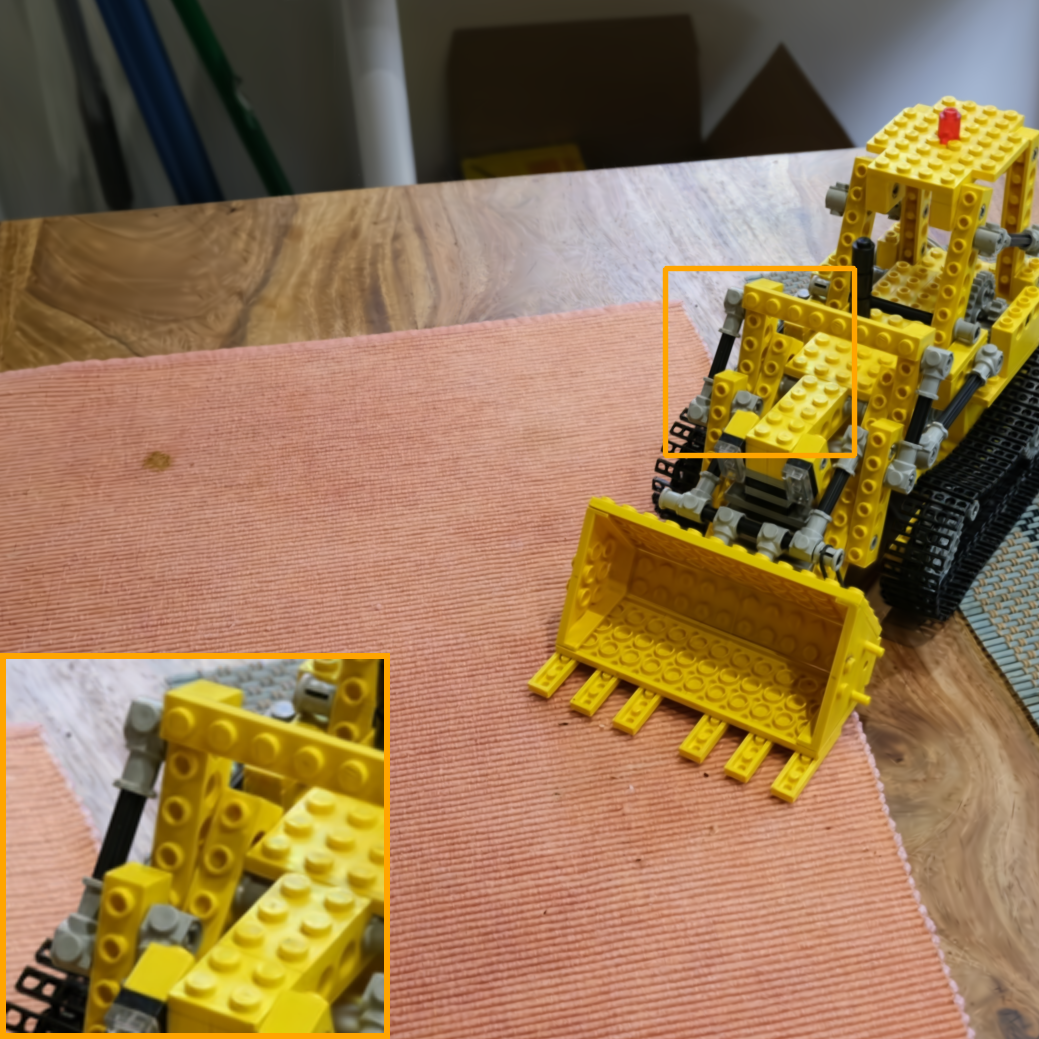} \\

\includegraphics[width=0.19\textwidth]{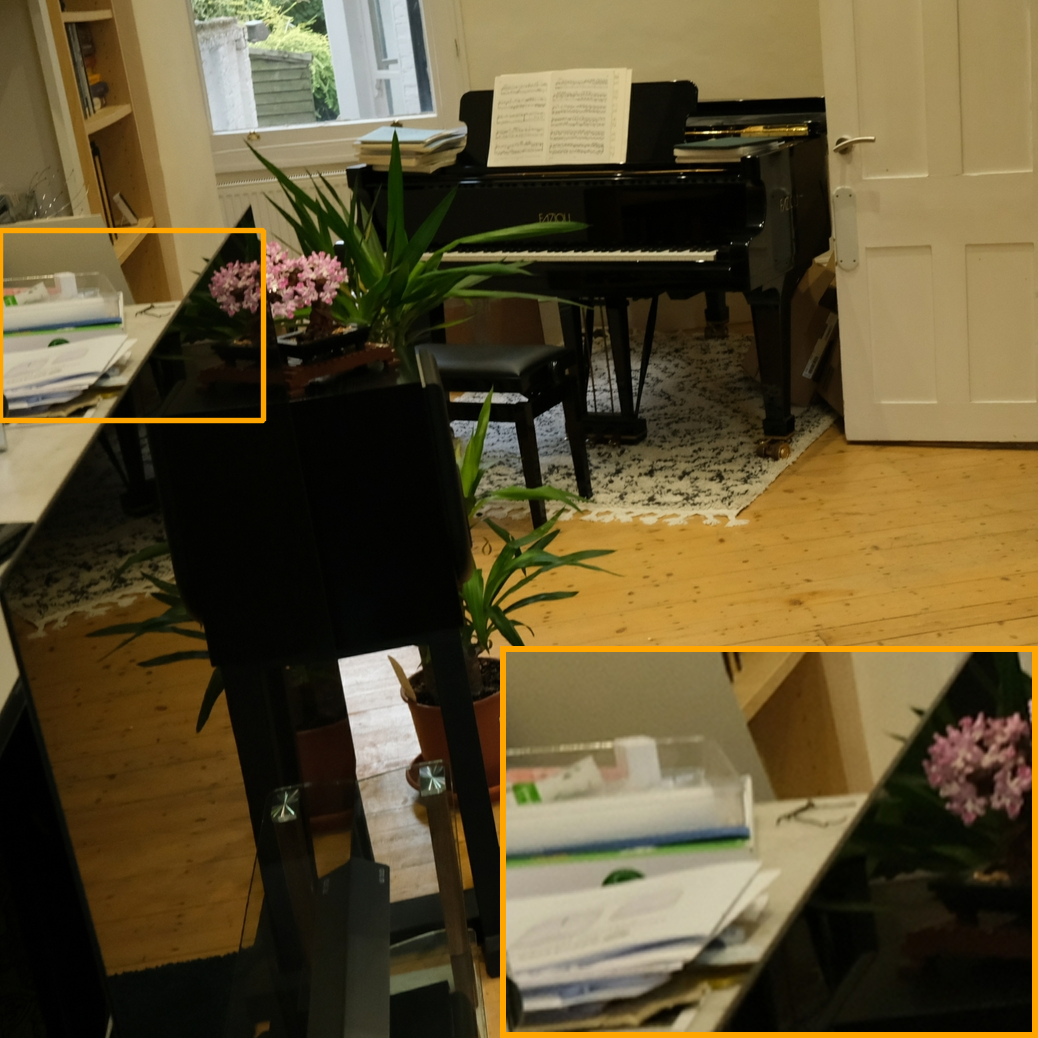} &
\includegraphics[width=0.19\textwidth]{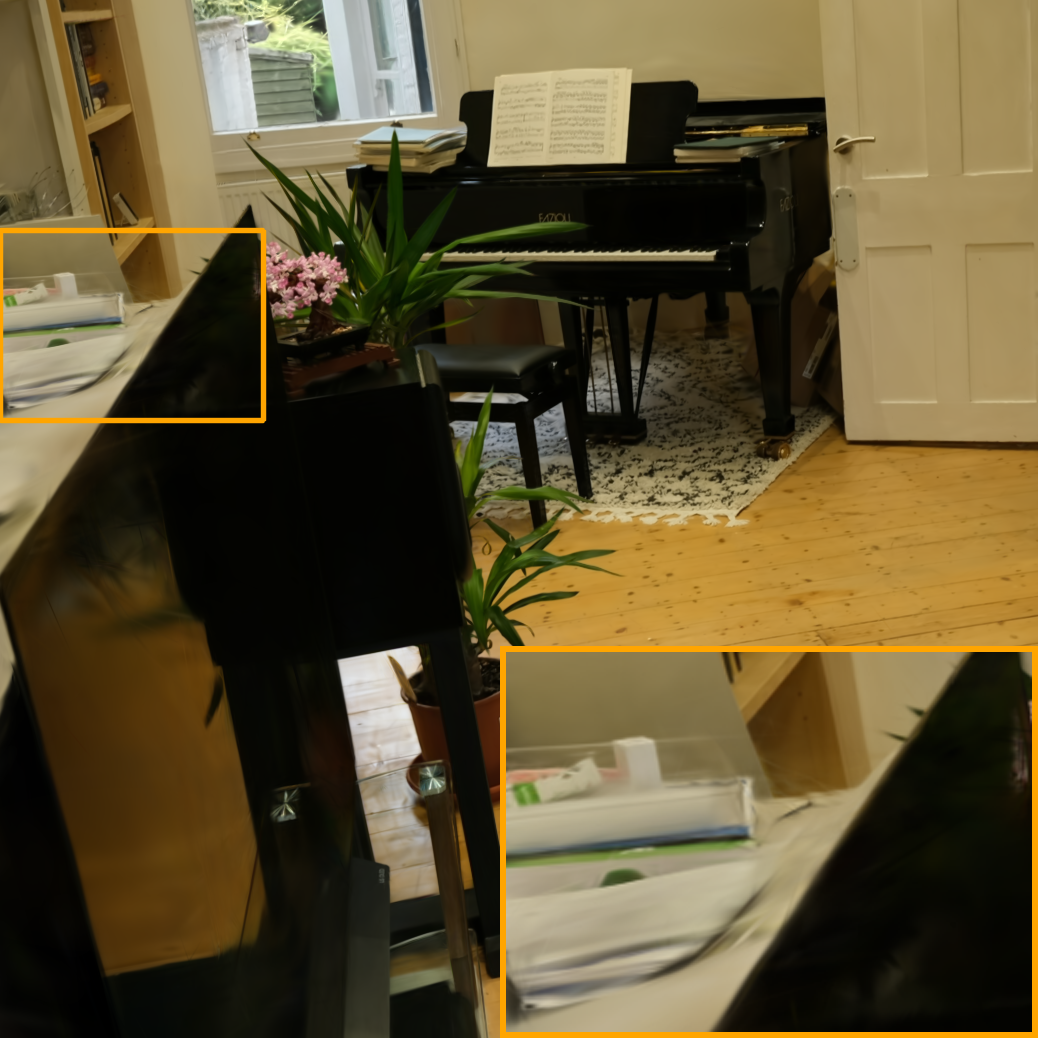} &
\includegraphics[width=0.19\textwidth]{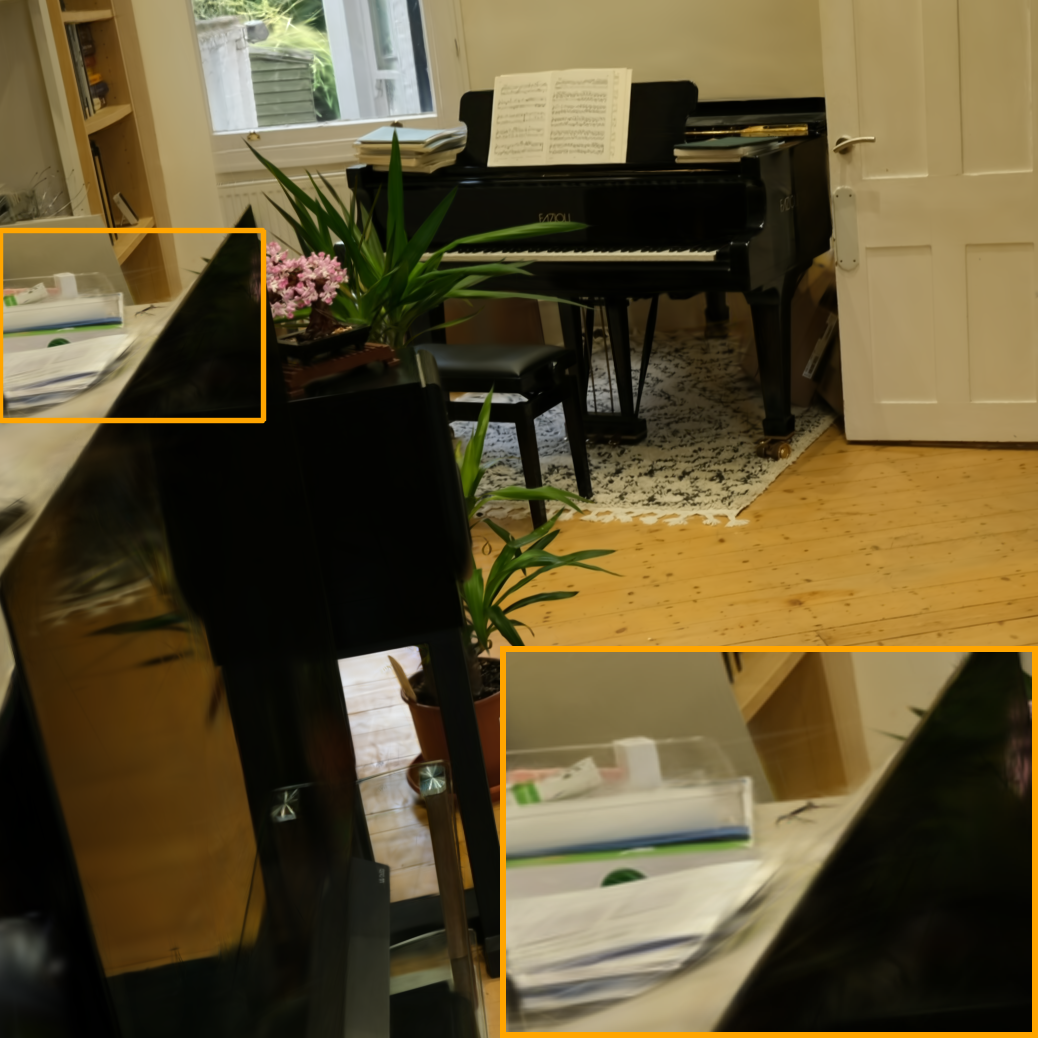} &
\includegraphics[width=0.19\textwidth]{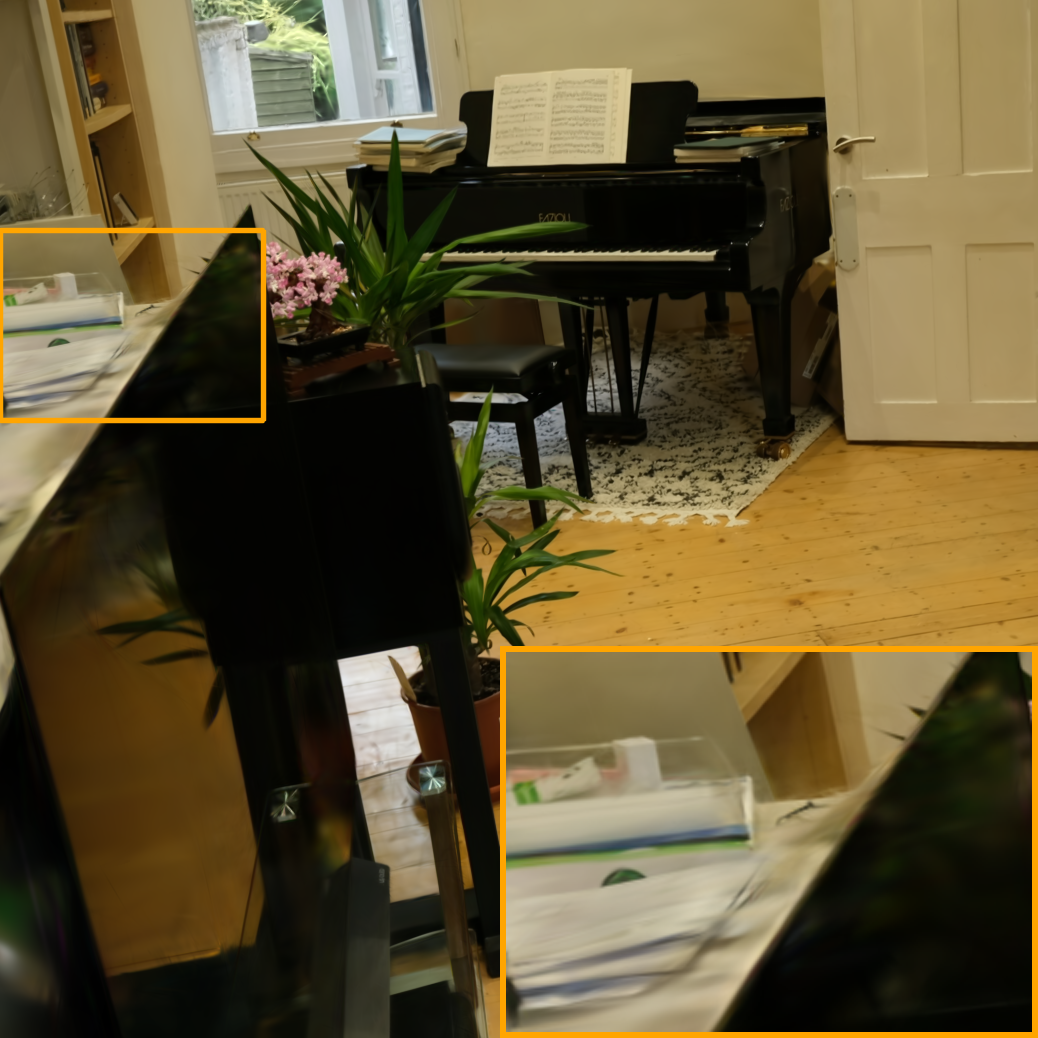} &
\includegraphics[width=0.19\textwidth]{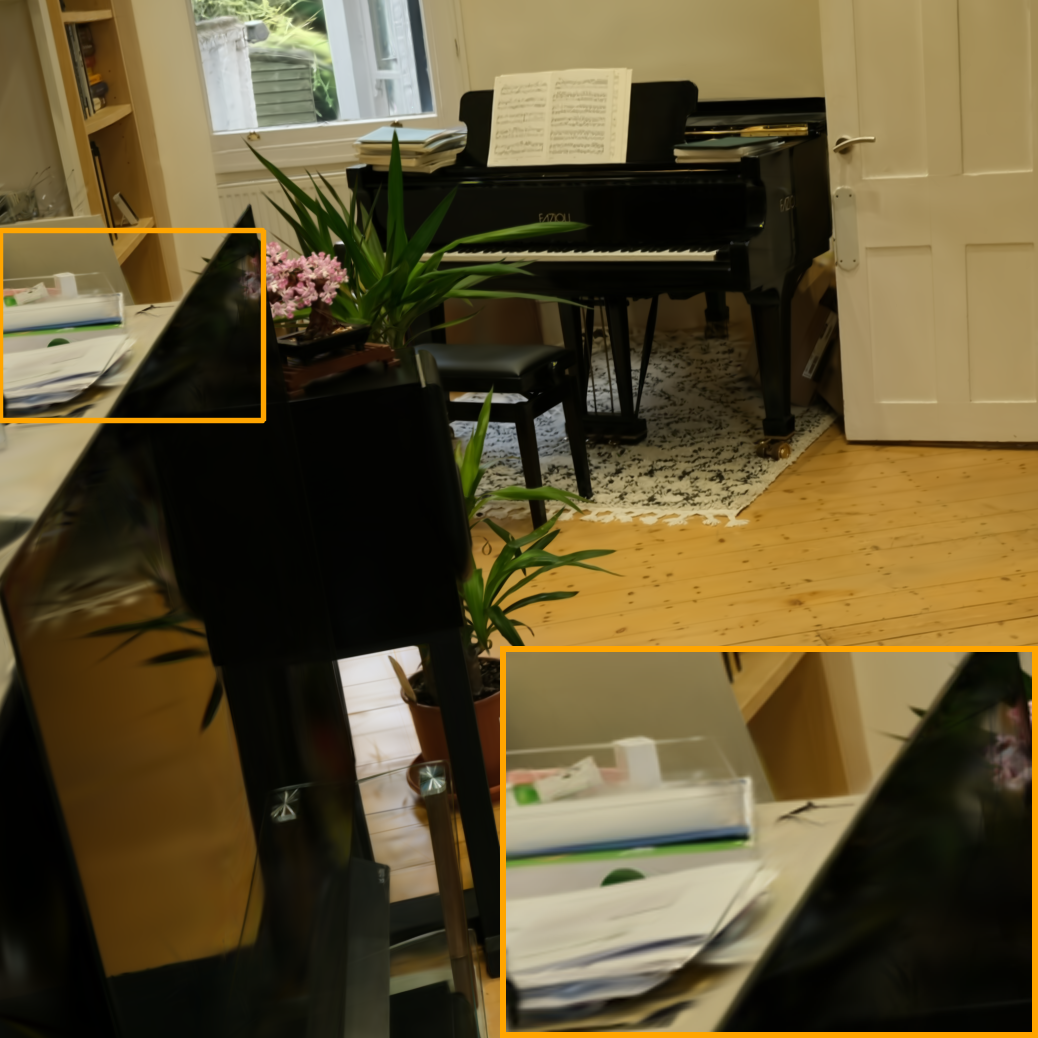} \\

\includegraphics[width=0.19\textwidth]{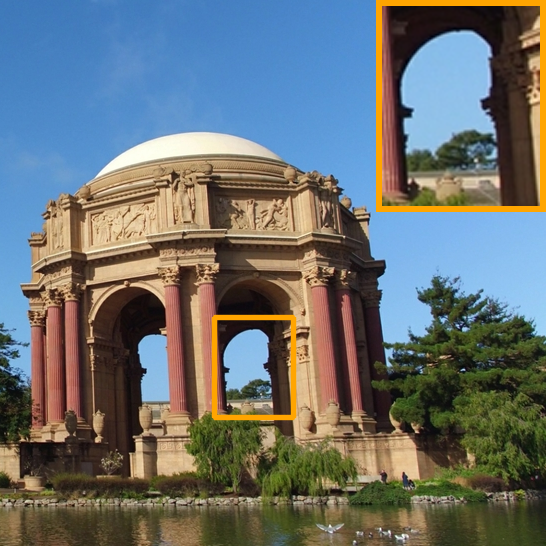} &
\includegraphics[width=0.19\textwidth]{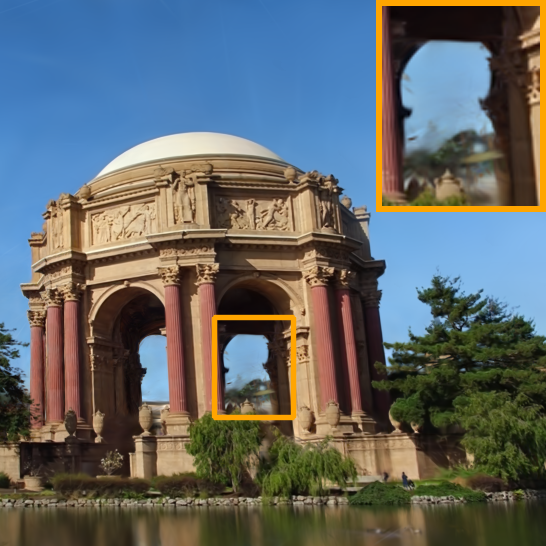} &
\includegraphics[width=0.19\textwidth]{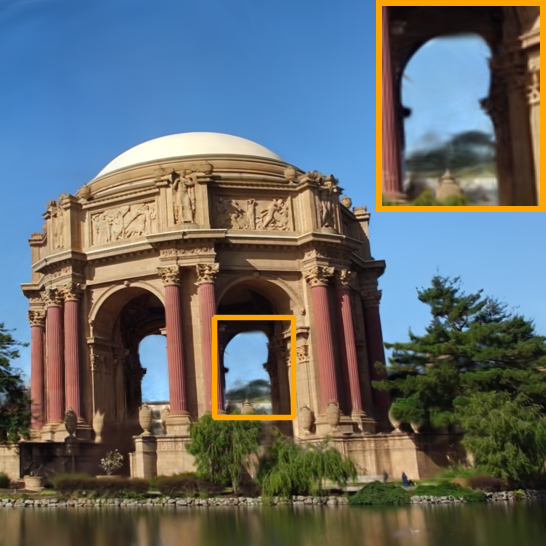} &
\includegraphics[width=0.19\textwidth]{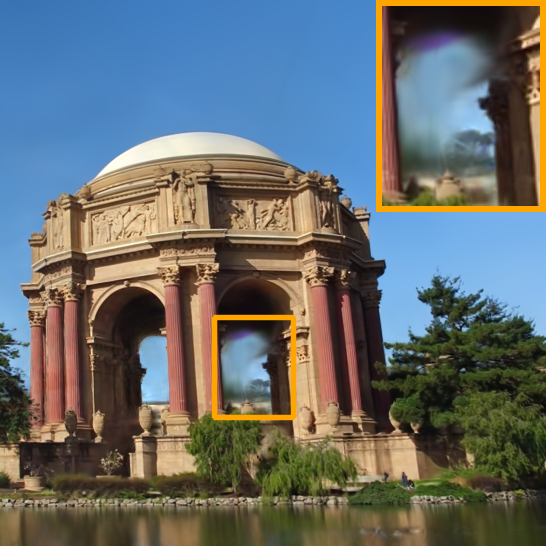} &
\includegraphics[width=0.19\textwidth]{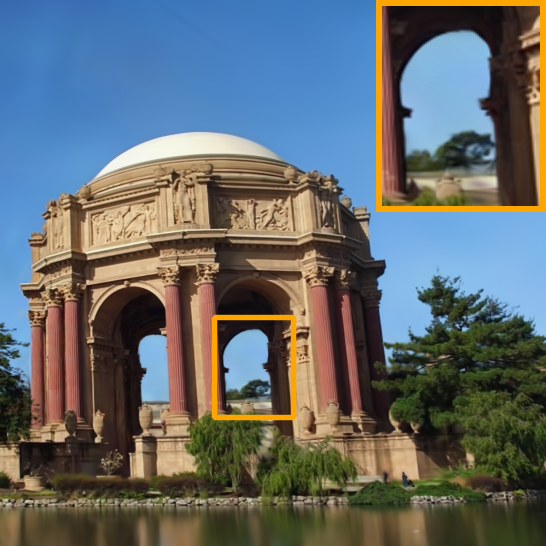} \\

\includegraphics[width=0.19\textwidth]{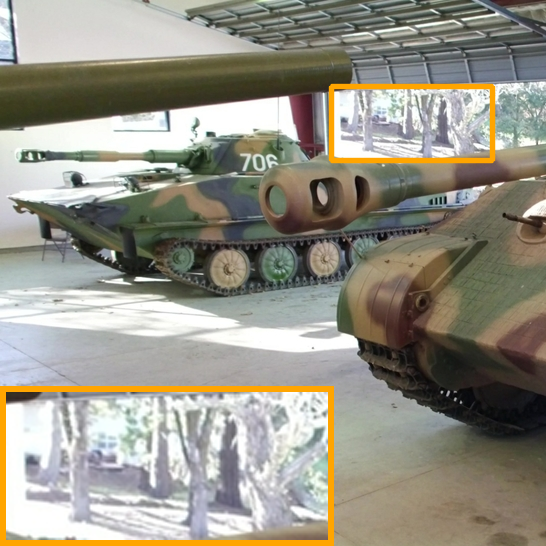} &
\includegraphics[width=0.19\textwidth]{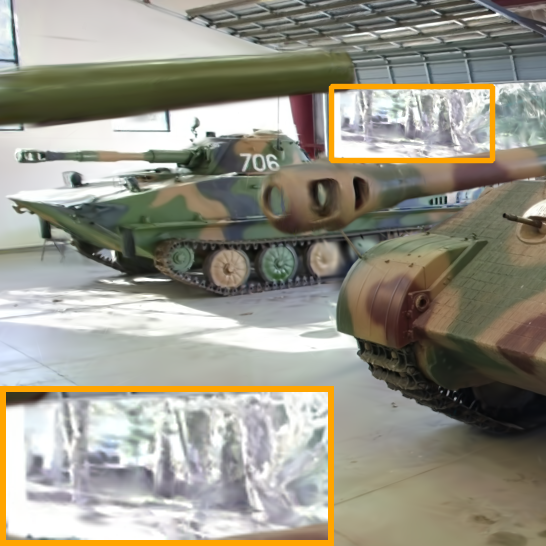} &
\includegraphics[width=0.19\textwidth]{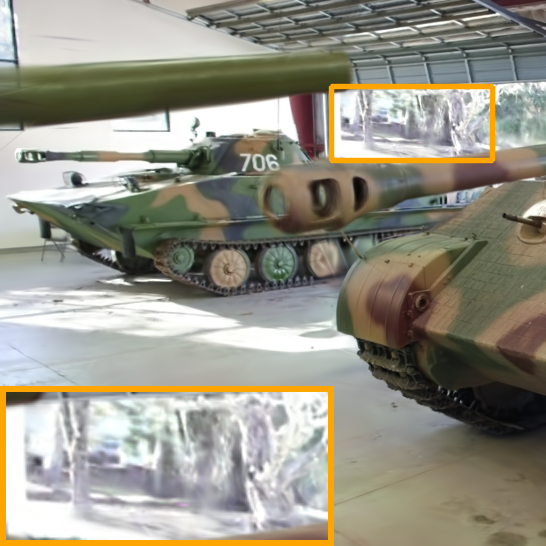} &
\includegraphics[width=0.19\textwidth]{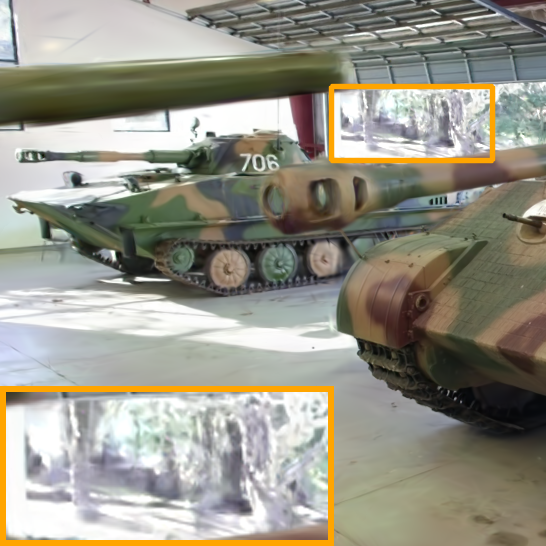} &
\includegraphics[width=0.19\textwidth]{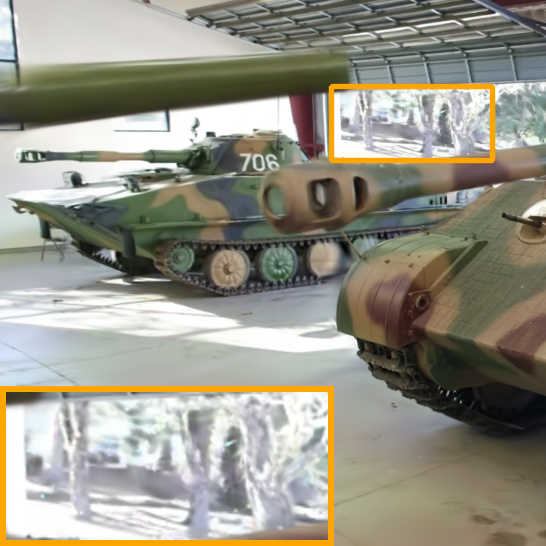} \\

\includegraphics[width=0.19\textwidth]{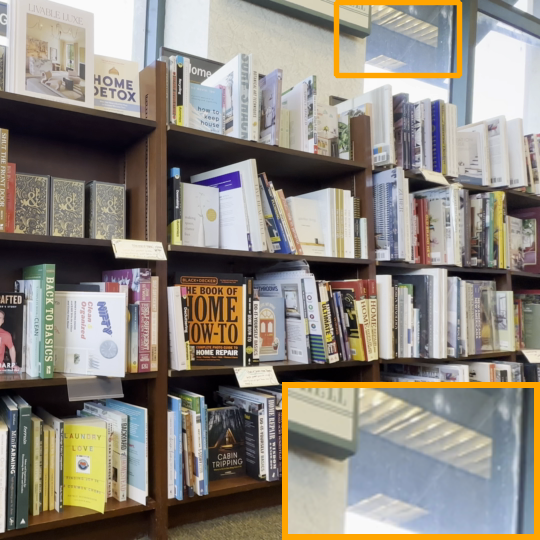} &
\includegraphics[width=0.19\textwidth]{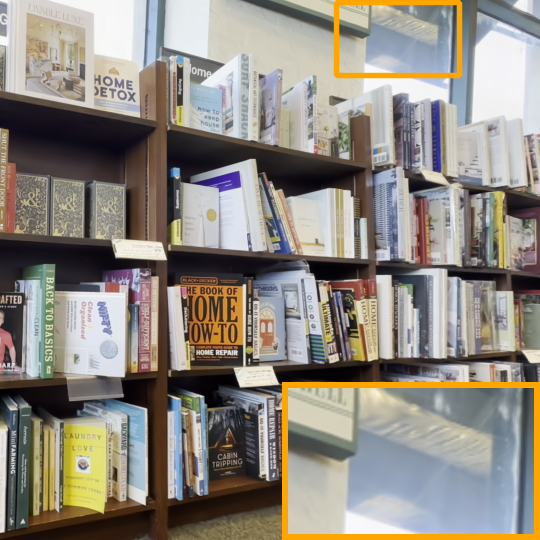} &
\includegraphics[width=0.19\textwidth]{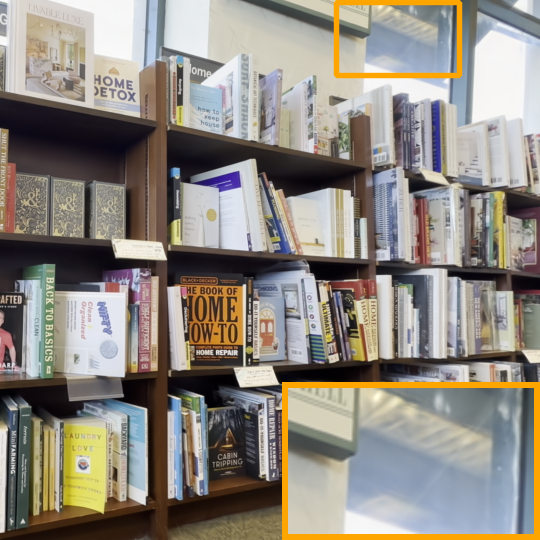} &
\includegraphics[width=0.19\textwidth]{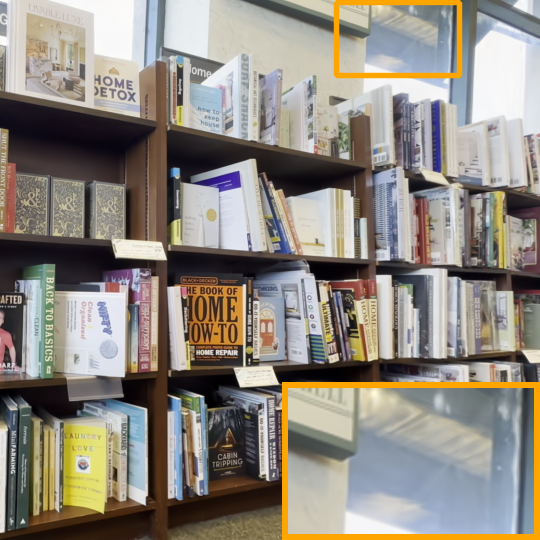} &
\includegraphics[width=0.19\textwidth]{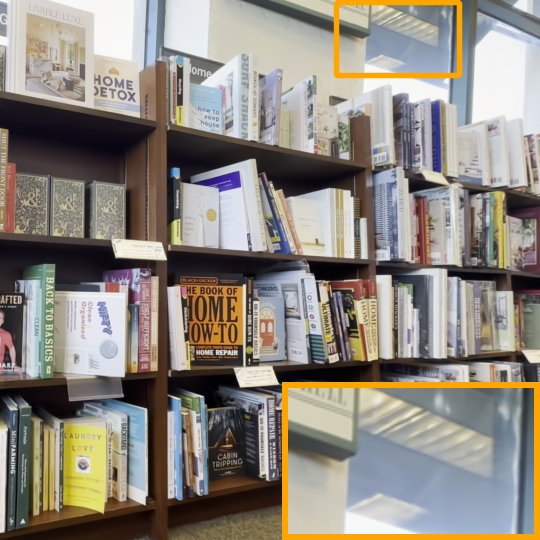} \\

\includegraphics[width=0.19\textwidth]{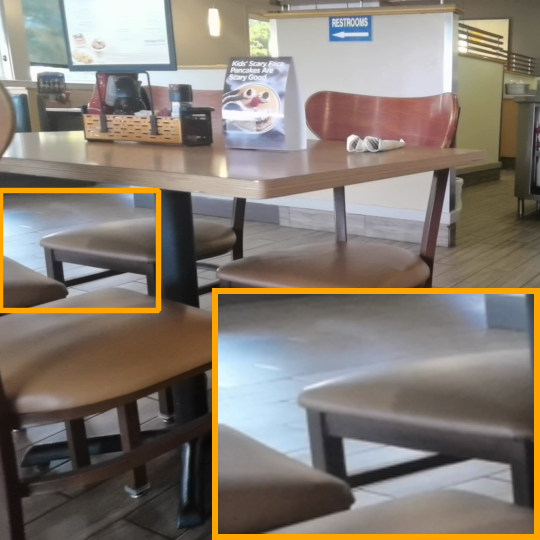} &
\includegraphics[width=0.19\textwidth]{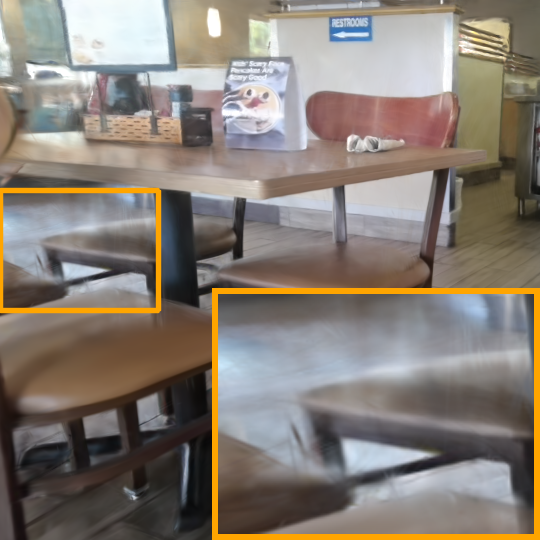} &
\includegraphics[width=0.19\textwidth]{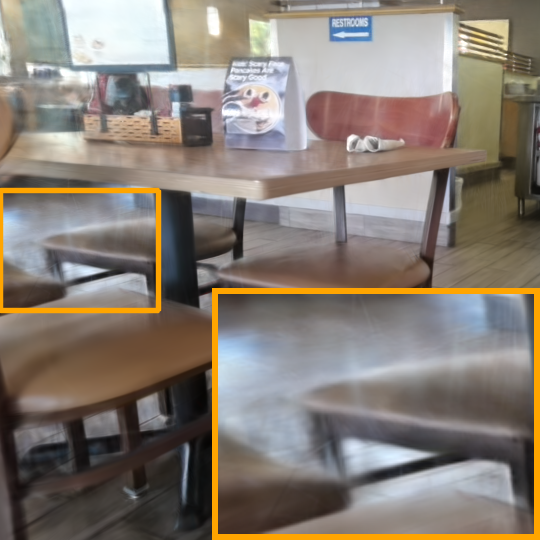} &
\includegraphics[width=0.19\textwidth]{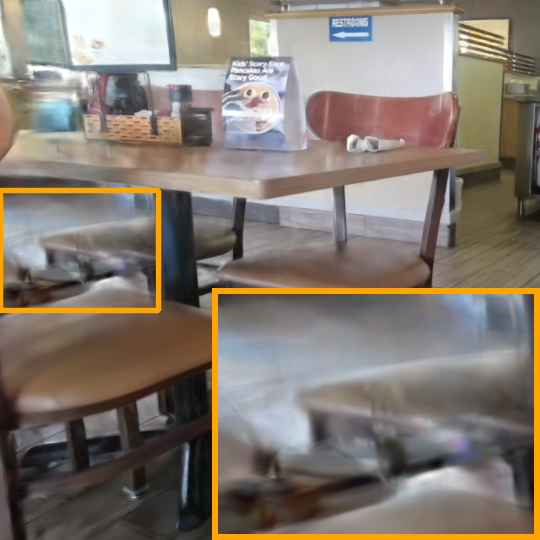} &
\includegraphics[width=0.19\textwidth]{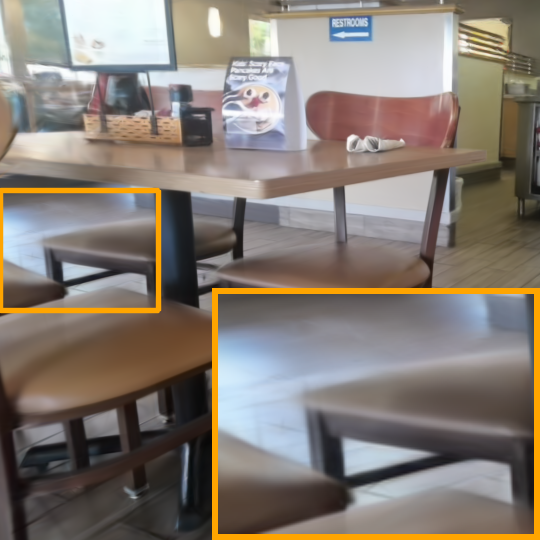} \\

\end{tabular}
}
\caption{Qualitative comparison across multiple scenes. Top to Bottom: MipNeRF360 - scenes 'kitchen' and 'room', TnT - scenes 'palace' and 'panther', DL3DV - scenes '387e' and '06da'.}
\label{fig:qual_grid}
\vspace{-1.0em}
\end{figure}
\fi

\begin{figure*}[t]
    \centering
    \setlength{\tabcolsep}{2pt}
    \begin{tabular}{ccccc}
        \includegraphics[width=0.19\textwidth]{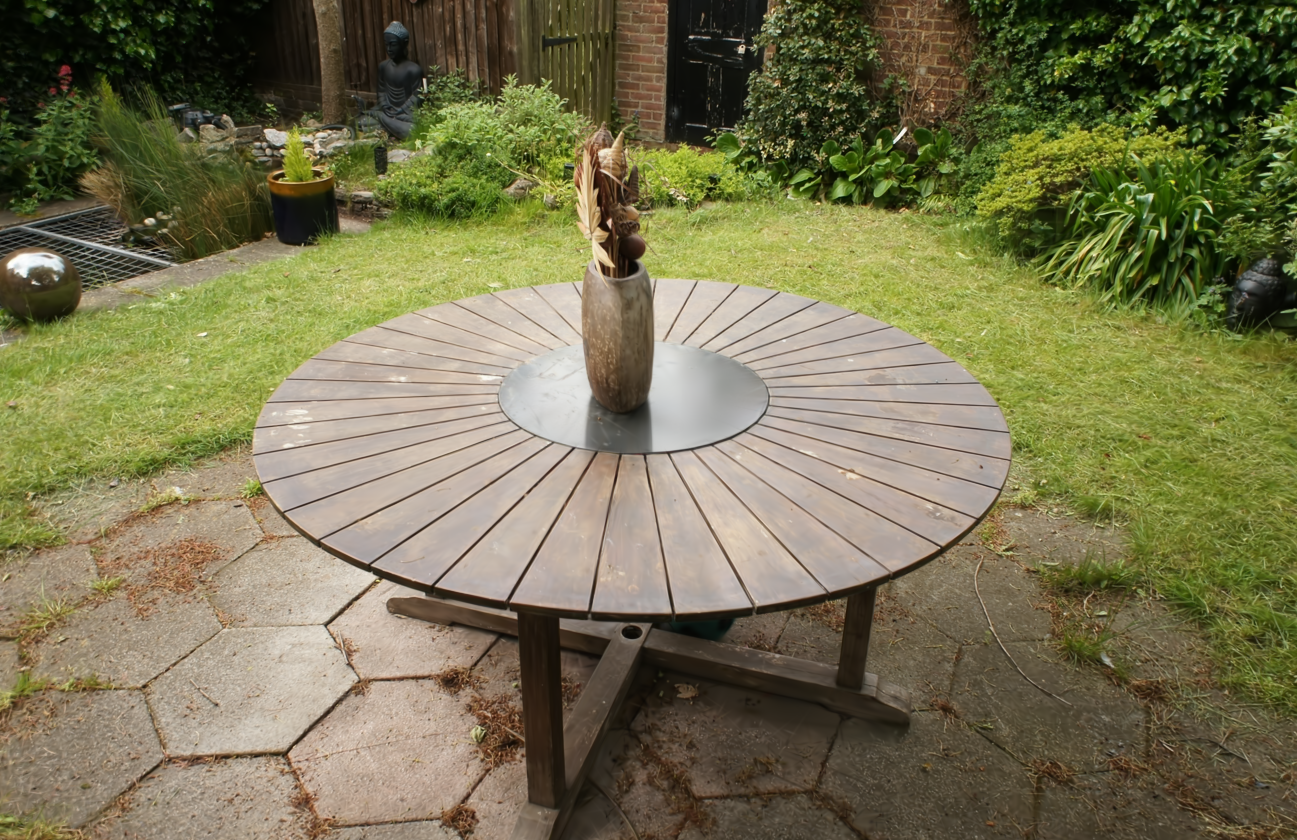} &
        \includegraphics[width=0.19\textwidth]{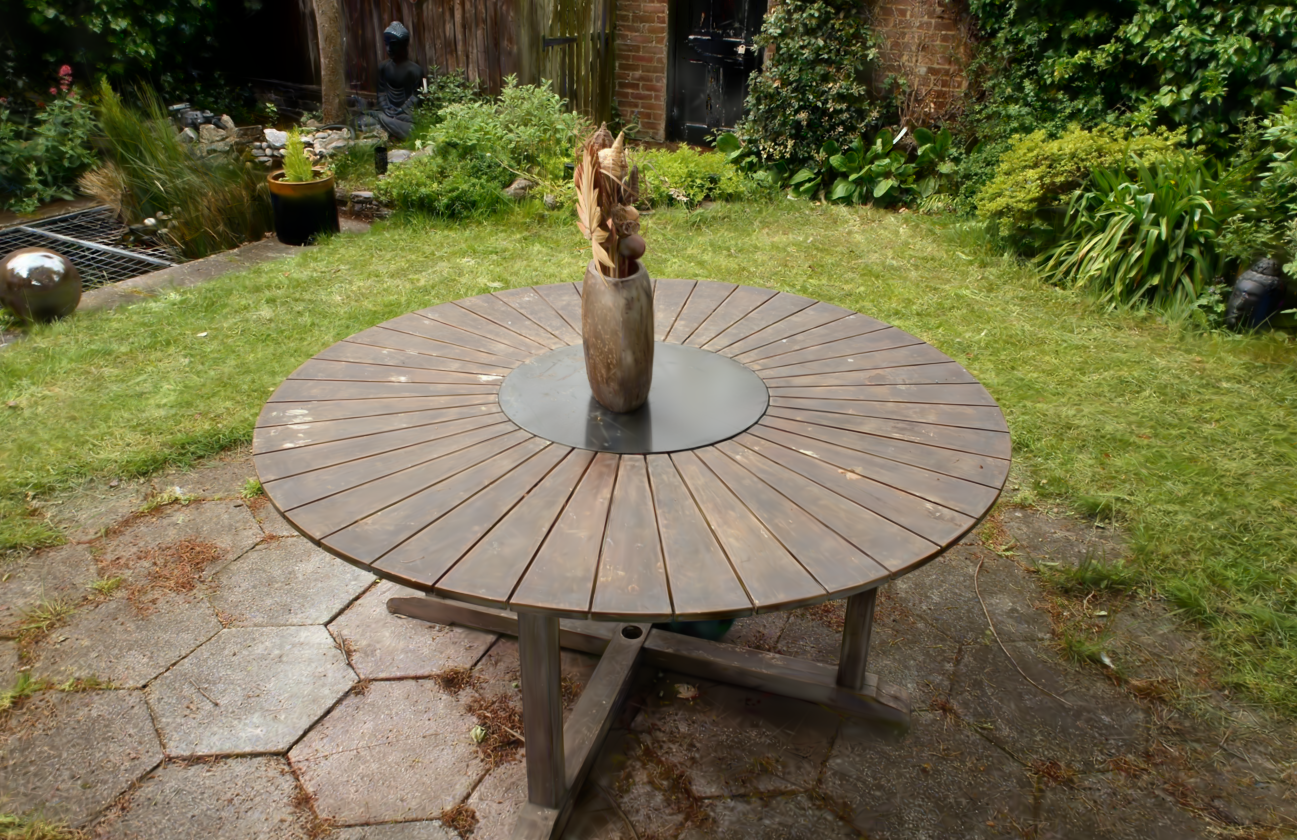} &
        \includegraphics[width=0.19\textwidth]{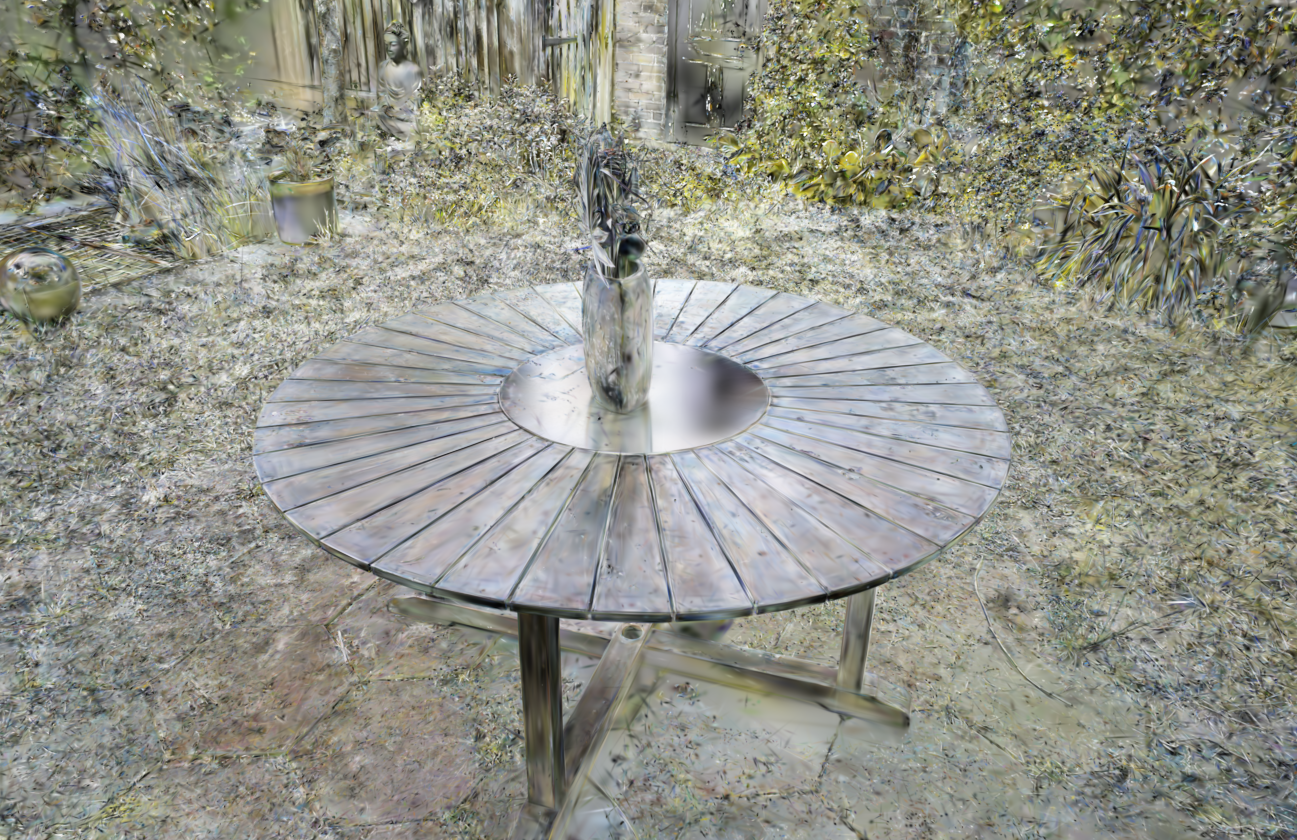} &
        \includegraphics[width=0.19\textwidth]{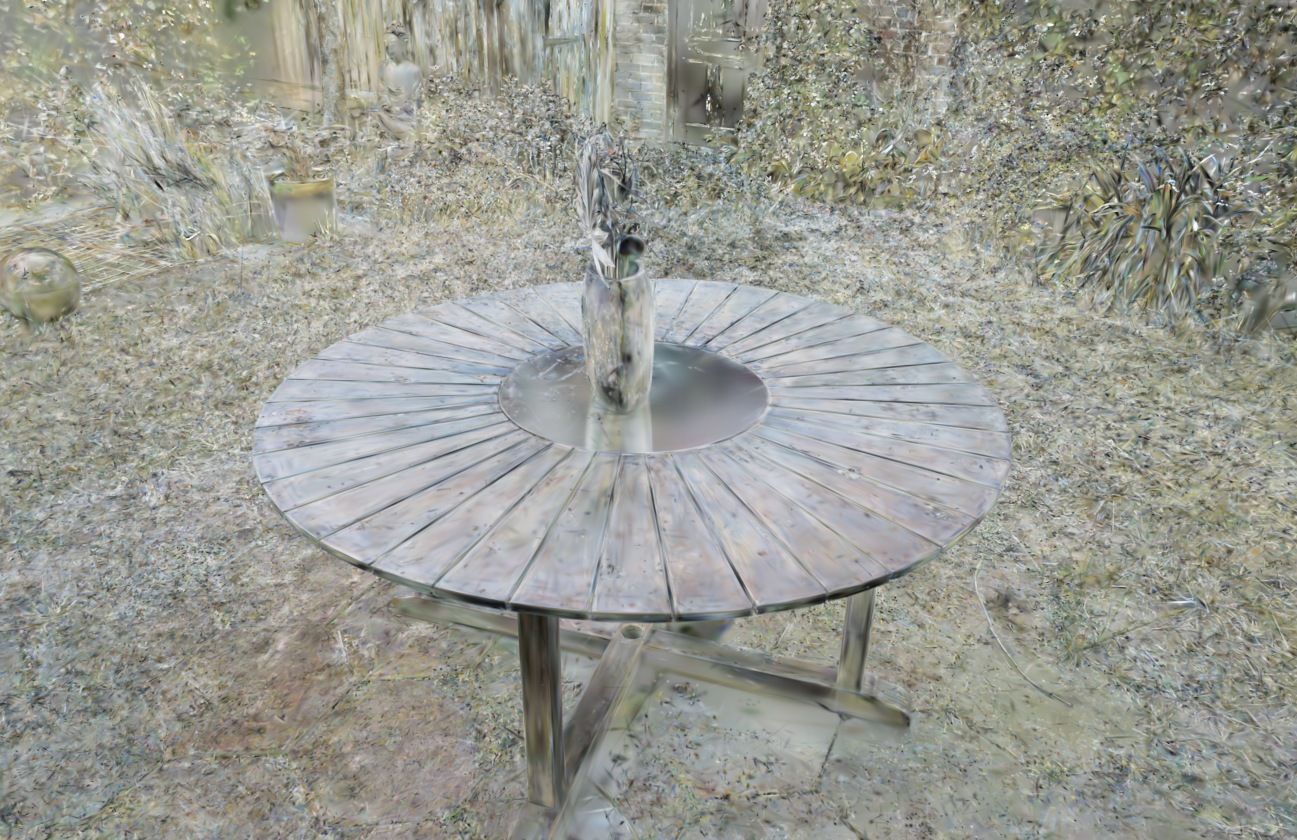} &
        \includegraphics[width=0.19\textwidth]{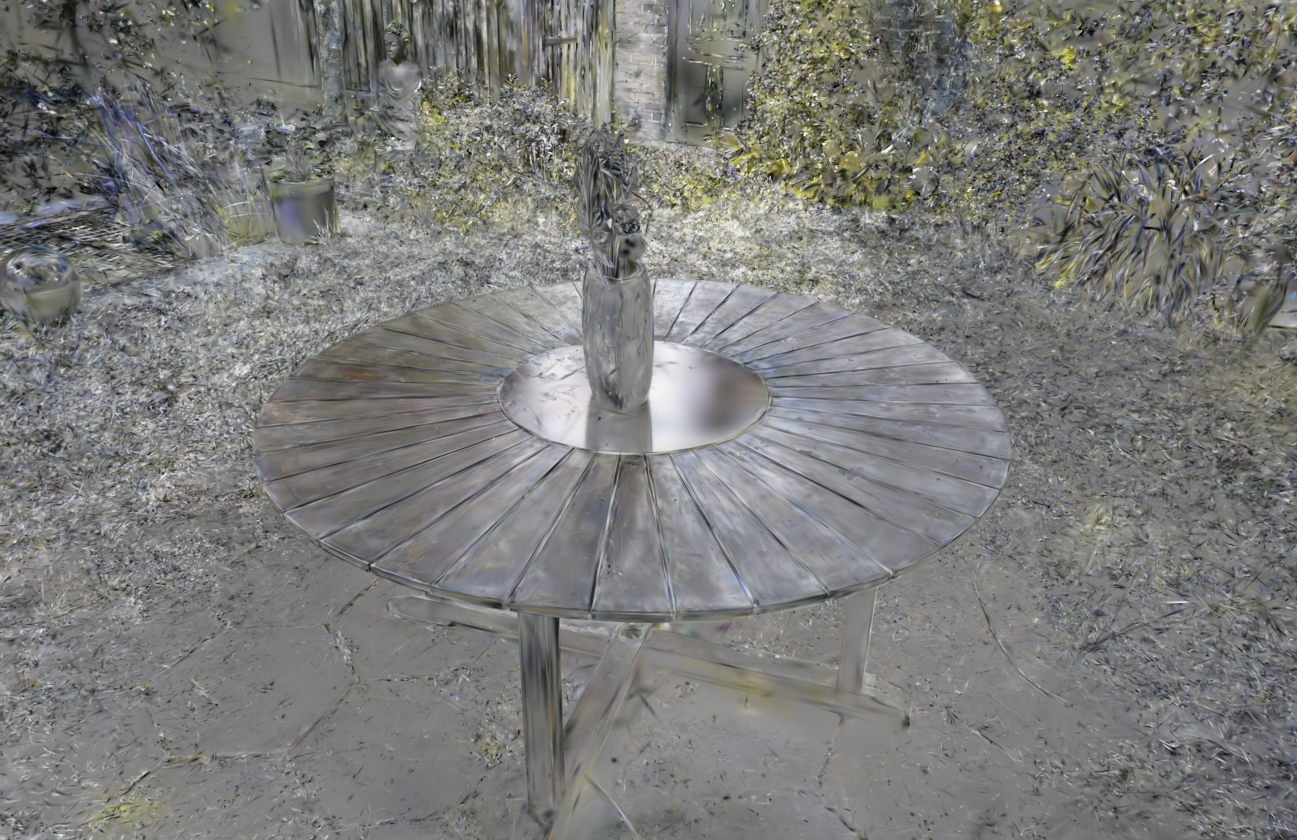} \\[-1pt]
        \footnotesize Full Render &
        \footnotesize Base Color &
        \footnotesize Residual &
        \footnotesize Low-Frequency &
        \footnotesize High-Frequency
    \end{tabular}
    \vspace{-0.5em}
    \caption{\textbf{Appearance decomposition of NRF-GS.}
    Visualization of the full rendering, base color, and learned
    neural appearance residual, including its low- and high-frequency
    components. Signed residuals are mapped to $[0,1]$, where
    neutral gray ($0.5$) indicates zero correction. The components are visualized individually and are not strictly additive due to the nonlinear activation as per Eq.~\ref{eq:res_comp}.}
    \label{fig:appearance_decomposition}
    \vspace{-1em}
\end{figure*}

\textbf{Quantitative and qualitative results.} Even with significantly lower point count, NRF-GS achieves the best PSNR across all datasets, with strong MS-SSIM and L1. Although the gains are modest, they are consistent ($\sim$0.3--0.8 dB), suggesting improved expressiveness rather than increased geometric capacity. Qualitatively (Fig.~\ref{fig:teaser_images},~\ref{fig:qual_grid}), it produces sharper reconstructions and more accurate view-dependent effects, particularly for specularities (\textbf{kitchen}, \textbf{room}, \textbf{06da}), and preserves fine details of low-parallax background in large scenes (\textbf{palace}, \textbf{panther}, \textbf{387e}). The scene-level shared network and better high-frequency modeling helps achieve the improvement with efficient splat distribution. Fig.~\ref{fig:appearance_decomposition} visualizes the contributions of the base color and the learned neural appearance residual, including its low- and high-frequency components. The signed residuals are mapped to $[0,1]$ for visualization, with neutral gray ($0.5$) indicating zero correction, darker values indicating negative corrections, and brighter values indicating positive corrections.

\textbf{Training time.} NRF-GS lies between analytic and neural baselines: slightly slower than 3DGS due to neural overhead, but faster than VDGS, GSNB, NeST and NTS (Tab.~\ref{tab:main_results}). GSNB, NeST and NTS are significantly slower and fail on some high-resolution scenes (e.g., \emph{Bicycle}, \emph{Garden}) due to memory limits; we therefore also report results excluding these scenes. During inference, on MipNeRF360** dataset, NRF-GS runs at 115 FPS vs 124 FPS for 3DGS, still remaining faster than VDGS (76 FPS), GSNB (101 FPS) and the two neural texture methods NeST (15 FPS) and NTS (5 FPS).

\section{Conclusions}

We presented \textbf{NRF-GS}, a hybrid formulation that augments 3DGS with a shared neural residual field and per-splat latent features for view-dependent appearance modeling, which shifts the directional modeling from independent analytic bases to a learned scene-level function. NRF-GS achieves comparable rendering quality while reducing the number of Gaussians by up to \textbf{50\%}, indicating that improved appearance modeling directly translates to higher representation efficiency. The frequency-decomposed design further improves robustness by separating stable low-frequency components from weighted high-frequency effects, leading to consistent performance across diverse scenes.\newline
\textbf{Limitations.} Despite these advantages, NRF-GS introduces additional computational overhead due to per-splat neural evaluations, making it slower than purely analytic approaches like 3DGS, but still faster than other neural baselines. Additionally, it inherits standard assumptions from 3DGS, including accurate camera poses and initialization, and does not explicitly optimize for runtime or memory efficiency beyond reducing the number of splats. For our method, LPIPS exhibits a modest, dataset-dependent trade-off, which may arise from the compact shared appearance which can slightly smooth fine details and boundaries. Such small structural differences have limited impact on PSNR or L1 metrics while being more strongly penalized by the deep feature activations used by LPIPS~\cite{zhang2018unreasonable}.
\newline
\textbf{Future works.} These limitations point to promising directions for future work, jointly optimizing appearance modeling with pruning, compression, and more efficient neural evaluation strategies. Also, the learned appearance features per splat could be further explored for more downstream tasks.

\section{Acknowledgments}

This work has been funded by the German Research Foundation (DFG), project funding for "Polymorphic Scene Representation for Enhanced Instant Scene Reconstruction" (Project No. 510825780, Ref. KO 2960/20-1)

\newpage

\bibliographystyle{plain}  
\bibliography{references}

\newpage


\appendix

\section{Technical appendices and supplementary material}

\subsection{Additional Ablation discussion}
Tab.~\ref{tab:single_vs_ours} shows the overfitting scenario discussed in Sec.~\ref{subsec:results_abl}. Note the for Single branch ablation, the unbounded scenes (namely \emph{bicycle}, \emph{garden}, \emph{flowers}, \emph{stump} and \emph{treehill}) have a worse test score than \textbf{Ours} split branch design, although having better train metrics, showing the overfitting issue without splitting and weighted high-frequency formulation.

\begin{table*}[hp]
\centering
\caption{Per-scene comparison of Single-branch vs Ours (NRF-GS) on MipNeRF360. Train/Test metrics are reported as sub-columns.}
\label{tab:single_vs_ours}
\resizebox{\textwidth}{!}{
\begin{tabular}{llcccccccc}
\toprule
\multirow{2}{*}{Scene} & \multirow{2}{*}{Method} 
& \multicolumn{2}{c}{MS-SSIM$\uparrow$}
& \multicolumn{2}{c}{LPIPS$\downarrow$}
& \multicolumn{2}{c}{L1$\downarrow$}
& \multicolumn{2}{c}{PSNR$\uparrow$} \\
& 
& Train & Test & Train & Test & Train & Test & Train & Test \\
\midrule

\multirow{2}{*}{Bicycle}
& Single & 0.864 & 0.715 & 0.162 & 0.234 & 0.026 & 0.047 & 27.34 & 23.71 \\
& Ours   & 0.843 & 0.759 & 0.170 & 0.203 & 0.029 & 0.041 & 26.11 & 25.06 \\

\midrule
\multirow{2}{*}{Bonsai}
& Single & 0.950 & 0.942 & 0.138 & 0.144 & 0.012 & 0.013 & 34.54 & 33.23 \\
& Ours   & 0.950 & 0.944 & 0.136 & 0.137 & 0.012 & 0.014 & 34.10 & 33.24 \\

\midrule
\multirow{2}{*}{Counter}
& Single & 0.926 & 0.906 & 0.151 & 0.173 & 0.013 & 0.018 & 32.03 & 29.81 \\
& Ours   & 0.924 & 0.913 & 0.153 & 0.158 & 0.014 & 0.017 & 31.59 & 29.83 \\

\midrule
\multirow{2}{*}{Flowers}
& Single & 0.760 & 0.576 & 0.260 & 0.377 & 0.038 & 0.061 & 24.24 & 21.02 \\
& Ours   & 0.748 & 0.599 & 0.261 & 0.349 & 0.040 & 0.057 & 23.70 & 21.41 \\

\midrule
\multirow{2}{*}{Garden}
& Single & 0.915 & 0.852 & 0.075 & 0.100 & 0.018 & 0.028 & 31.05 & 27.20 \\
& Ours   & 0.907 & 0.864 & 0.070 & 0.082 & 0.020 & 0.026 & 30.20 & 27.67 \\

\midrule
\multirow{2}{*}{Kitchen}
& Single & 0.942 & 0.925 & 0.092 & 0.107 & 0.014 & 0.016 & 33.42 & 31.83 \\
& Ours   & 0.942 & 0.927 & 0.090 & 0.096 & 0.014 & 0.016 & 33.51 & 32.10 \\

\midrule
\multirow{2}{*}{Room}
& Single & 0.937 & 0.919 & 0.153 & 0.178 & 0.011 & 0.016 & 34.94 & 32.00 \\
& Ours   & 0.936 & 0.920 & 0.156 & 0.168 & 0.011 & 0.015 & 34.74 & 32.14 \\

\midrule
\multirow{2}{*}{Stump}
& Single & 0.902 & 0.716 & 0.126 & 0.206 & 0.018 & 0.040 & 31.06 & 25.06 \\
& Ours   & 0.890 & 0.762 & 0.135 & 0.178 & 0.019 & 0.034 & 30.10 & 26.39 \\

\midrule
\multirow{2}{*}{Treehill}
& Single & 0.787 & 0.612 & 0.255 & 0.380 & 0.034 & 0.048 & 24.54 & 22.63 \\
& Ours   & 0.788 & 0.636 & 0.252 & 0.350 & 0.033 & 0.045 & 24.15 & 22.84 \\

\bottomrule
\end{tabular}
}
\end{table*}

\subsection{Higher order SH degree comparison}
We compare our method against higher-order SH which helps distinguish the benefit of the proposed neural residual field from the more general effect of simply increasing analytical basis i.e. higher SH degrees. We therefore evaluated additional degree-4 and degree-5 SH modes for standard 3DGS baseline under the same Mip-NeRF360* configuration in Tab.~\ref{tab:sh_degree_comparison} (NOTE: Mip-NeRF360* excludes scene 'bicycle' and 'garden' since SH degree 5 went Out-of-Memory for these experiments on our system). The results show that increasing the SH degree provides only a small improvement in reconstruction-quality and does not substantially reduce the number of Gaussians. At the same time, SH4 and SH5 model size and VRAM is increased with lower FPS. In comparison, NRF-GS achieves higher quality than both higher-order SH baselines while using fewer Gaussians, comparable training time and faster inference speed. The overall results indicate that simply increasing the SH degree does not reproduce the compactness achieved by the proposed neural residual field. NRF-GS therefore provides a more favorable overall quality--compactness--runtime trade-off.

\begin{table*}[t]
\centering
\caption{
Comparison with higher-order spherical harmonics (SH) on the
MipNeRF360* subset (excluding \emph{Bicycle} and \emph{Garden}).
Points are reported in millions.
Best and second-best values are highlighted.
}
\label{tab:sh_degree_comparison}
\resizebox{\textwidth}{!}{
\begin{tabular}{lccccccccc}
\toprule
Method
& Pts (M)$\downarrow$
& Time (m)$\downarrow$
& SSIM$\uparrow$
& LPIPS$\downarrow$
& L1$\downarrow$
& PSNR$\uparrow$
& Model (GB)$\downarrow$
& VRAM (GB)$\downarrow$
& FPS$\uparrow$ \\
\midrule
SH3
& 2.605
& \cellcolor{best}\textbf{24.22}
& 0.812
& 0.199
& 0.030
& 27.74
& \cellcolor{second}0.596
& \cellcolor{second}8.146
& \cellcolor{best}\textbf{115.10\phantom{0}} \\
SH4
& 2.593
& \cellcolor{second}27.50
& \cellcolor{second}0.815
& \cellcolor{second}0.196
& \cellcolor{second}0.028
& 27.78
& 0.866
& 9.678
& \cellcolor{second}106.70\phantom{0} \\
SH5
& \cellcolor{second}2.580
& 31.75
& \cellcolor{second}0.815
& \cellcolor{best}\textbf{0.195}
& \cellcolor{second}0.028
& \cellcolor{second}27.81
& 1.192
& 11.660\phantom{0}
& 97.14 \\
\textbf{Ours}
& \cellcolor{best}\textbf{1.361}
& 28.83
& \cellcolor{best}\textbf{0.815}
& 0.205
& \cellcolor{best}\textbf{0.028}
& \cellcolor{best}\textbf{28.27}
& \cellcolor{best}\textbf{0.267}
& \cellcolor{best}\textbf{7.576}
& 105.20\phantom{0} \\
\bottomrule
\end{tabular}
}
\end{table*}

\subsection{Additional Results}

Tab.~\ref{tab:main_results_std} shows the average test results on all datasets along with per scene standard deviations.

\begin{table*}[hp]
\centering
\caption{
Average test-set results (mean $\pm$ std) on MipNeRF360, DL3DV, and Tanks and Temples. 
\emph{MipNeRF360*} excludes \emph{Bicycle} and \emph{Garden} scenes where GSNB runs out-of-memory. 
Points are reported in millions.
}
\label{tab:main_results_std}
\resizebox{\textwidth}{!}{
\begin{tabular}{llcccccc}
\toprule
Dataset & Method & Points (M)$\downarrow$ & Time (min)$\downarrow$ & MS-SSIM$\uparrow$ & LPIPS$\downarrow$ & L1$\downarrow$ & PSNR$\uparrow$ \\
\midrule
\multirow{3}{*}{MipNeRF360}
& 3DGS   
& 3.359 $\pm$ 1.986
& 26.46 $\pm$ 05.02
& 0.813 $\pm$ 0.130
& 0.184 $\pm$ 0.094
& 0.030 $\pm$ 0.015
& 27.40 $\pm$ 3.95 \\
& VDGS   
& 3.548 $\pm$ 1.869
& 41.27 $\pm$ 09.95
& 0.813 $\pm$ 0.132
& 0.186 $\pm$ 0.096
& 0.030 $\pm$ 0.015
& 27.65 $\pm$ 4.09 \\
& NRF-GS (ours) 
& 1.656 $\pm$ 1.015
& 31.38 $\pm$ 07.12
& 0.814 $\pm$ 0.131
& 0.191 $\pm$ 0.097
& 0.029 $\pm$ 0.016
& 27.86 $\pm$ 4.27 \\
\midrule
\multirow{4}{*}{MipNeRF360*}
& 3DGS 
& 2.605 $\pm$ 1.503 
& 24.22 $\pm$ 02.66 
& 0.813 $\pm$ 0.147 
& 0.200 $\pm$ 0.098 
& 0.029 $\pm$ 0.017 
& 27.74 $\pm$ 4.46 \\
& VDGS 
& 2.828 $\pm$ 1.372 
& 36.69 $\pm$ 04.58 
& 0.814 $\pm$ 0.149 
& 0.200 $\pm$ 0.101 
& 0.029 $\pm$ 0.017 
& 28.04 $\pm$ 4.59 \\
& GSNB 
& 2.667 $\pm$ 1.315 
& 59.06 $\pm$ 20.68 
& 0.800 $\pm$ 0.162 
& 0.201 $\pm$ 0.102 
& 0.032 $\pm$ 0.020 
& 27.71 $\pm$ 5.13 \\
& NRF-GS (ours) 
& 1.361 $\pm$ 0.914 
& 28.83 $\pm$ 05.35 
& 0.814 $\pm$ 0.148 
& 0.205 $\pm$ 0.102 
& 0.028 $\pm$ 0.017 
& 28.28 $\pm$ 4.78 \\
\midrule
\multirow{4}{*}{DL3DV}
& 3DGS   
& 1.158 $\pm$ 0.316
& 12.74 $\pm$ 02.46
& 0.916 $\pm$ 0.057
& 0.088 $\pm$ 0.053
& 0.021 $\pm$ 0.009
& 29.48 $\pm$ 3.19 \\
& VDGS   
& 1.404 $\pm$ 0.465
& 21.15 $\pm$ 05.45
& 0.918 $\pm$ 0.058
& 0.089 $\pm$ 0.055
& 0.021 $\pm$ 0.009
& 29.92 $\pm$ 3.35 \\
& GSNB   
& 1.032 $\pm$ 0.274
& 25.28 $\pm$ 05.16
& 0.920 $\pm$ 0.056
& 0.086 $\pm$ 0.054
& 0.020 $\pm$ 0.010
& 30.60 $\pm$ 3.54 \\
& NRF-GS (ours) 
& 0.595 $\pm$ 0.246
& 14.52 $\pm$ 03.25
& 0.925 $\pm$ 0.052
& 0.089 $\pm$ 0.053
& 0.018 $\pm$ 0.008
& 30.82 $\pm$ 3.23 \\
\midrule
\multirow{4}{*}{TnT}
& 3DGS   
& 1.868 $\pm$ 1.064
& 15.35 $\pm$ 04.36
& 0.831 $\pm$ 0.052
& 0.165 $\pm$ 0.058
& 0.046 $\pm$ 0.016
& 23.85 $\pm$ 2.52 \\
& VDGS   
& 2.113 $\pm$ 1.009
& 26.37 $\pm$ 07.83
& 0.834 $\pm$ 0.051
& 0.163 $\pm$ 0.054
& 0.045 $\pm$ 0.015
& 24.12 $\pm$ 2.51 \\
& GSNB   
& 1.705 $\pm$ 0.902
& 35.75 $\pm$ 14.93
& 0.830 $\pm$ 0.051
& 0.164 $\pm$ 0.054
& 0.047 $\pm$ 0.017
& 23.92 $\pm$ 2.73 \\
& NRF-GS (ours) 
& 1.040 $\pm$ 0.483
& 19.91 $\pm$ 05.34
& 0.845 $\pm$ 0.045
& 0.164 $\pm$ 0.052
& 0.040 $\pm$ 0.014
& 24.68 $\pm$ 2.41 \\
\bottomrule
\end{tabular}
}
\end{table*}

\begin{figure}[hp]
\centering
\setlength{\tabcolsep}{0.5pt}

\resizebox{\textwidth}{!}{
\begin{tabular}{c c c c c c}

& \textbf{Kitchen}
& \textbf{Bonsai}
& \textbf{Lighthouse}
& \textbf{5c3a}
& \textbf{669c} \\

\raisebox{1.0\height}{\rotatebox{90}{\textbf{3DGS}}} &
\includegraphics[width=0.19\textwidth]{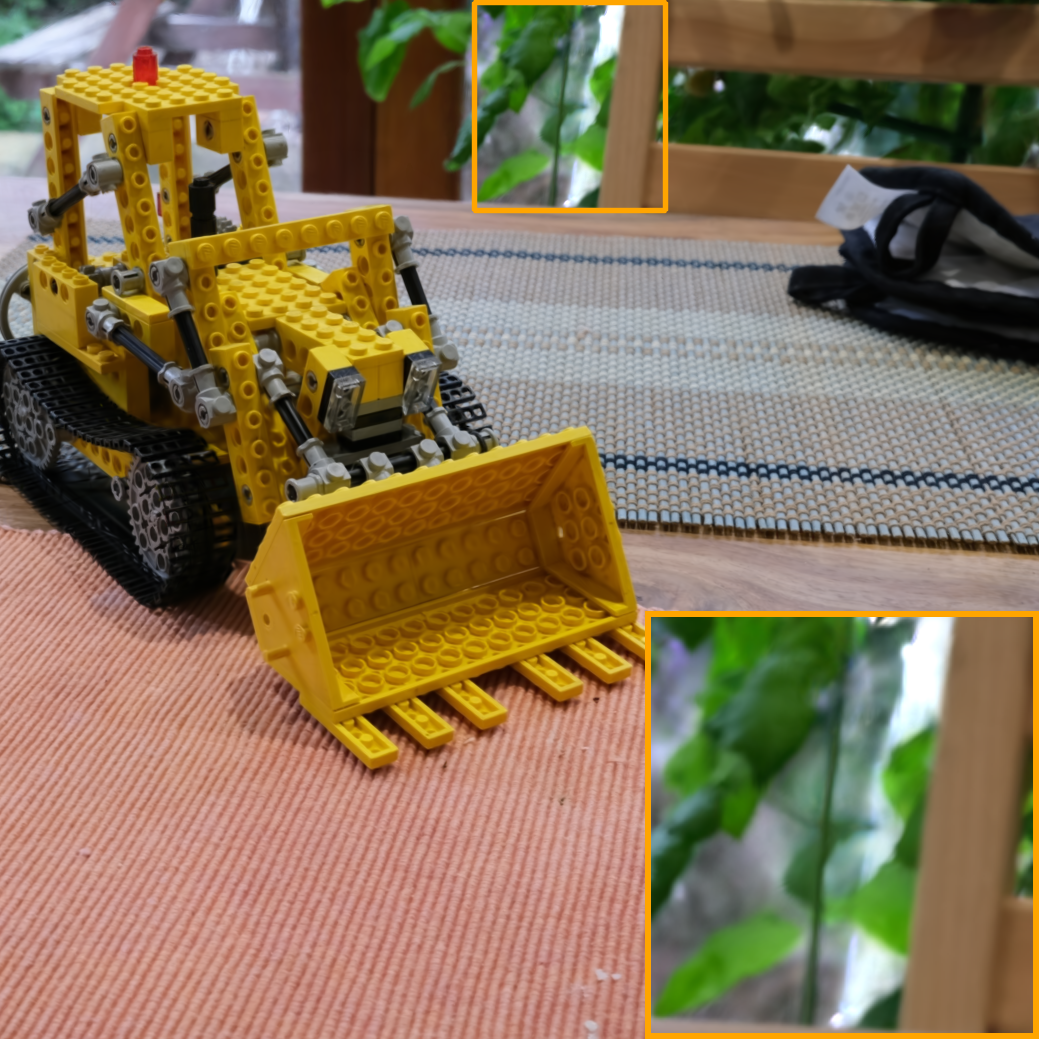} &
\includegraphics[width=0.19\textwidth]{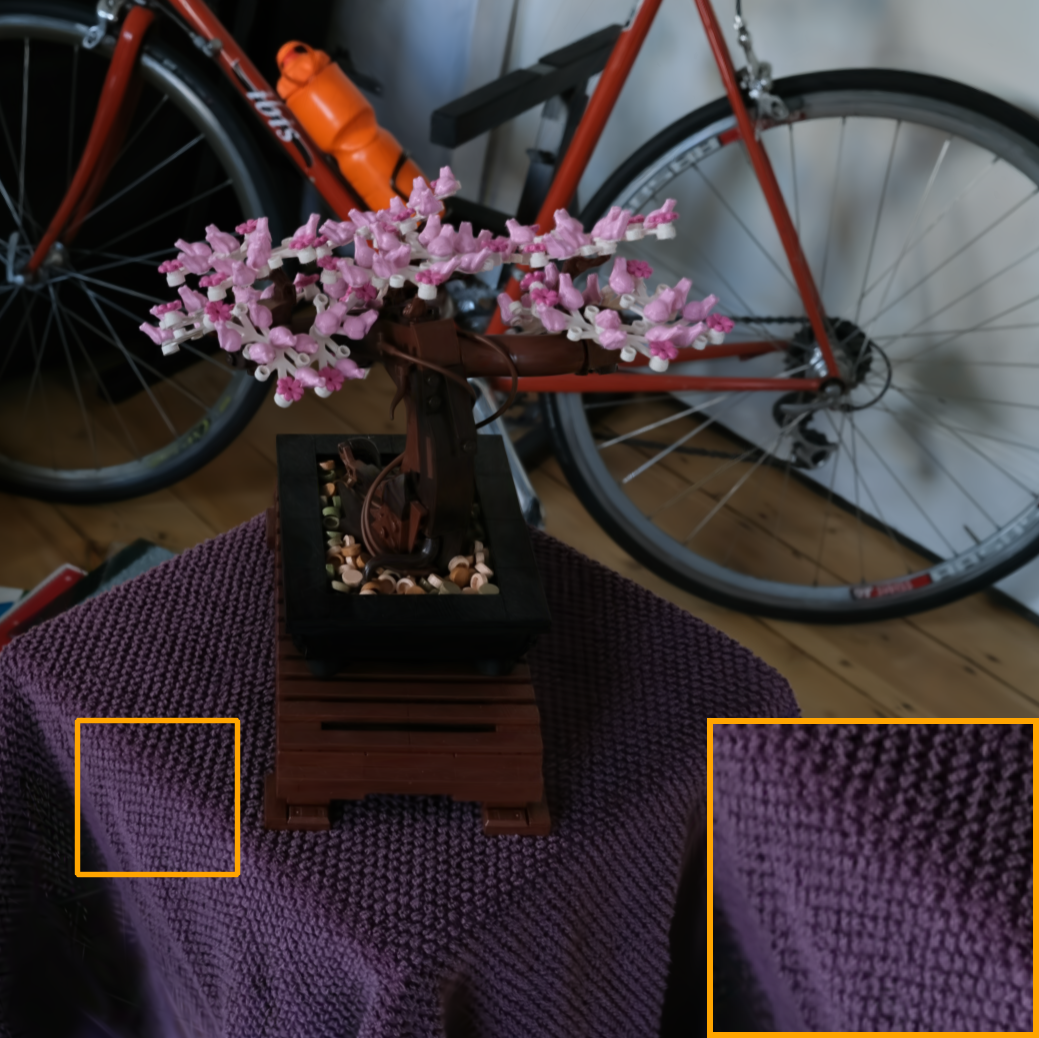} &
\includegraphics[width=0.19\textwidth]{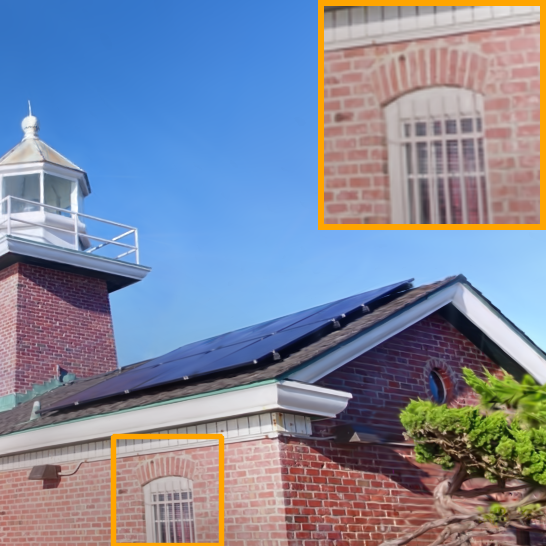} &
\includegraphics[width=0.19\textwidth]{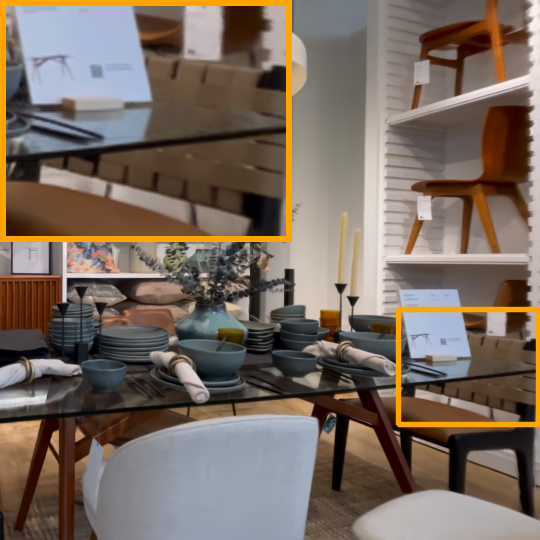} &
\includegraphics[width=0.19\textwidth]{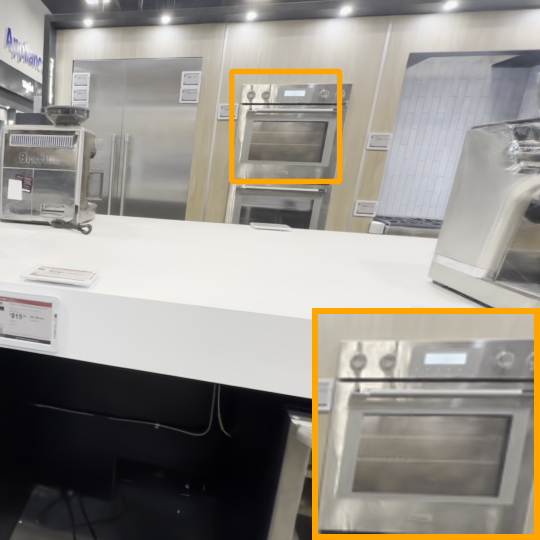} \\

\raisebox{1.0\height}{\rotatebox{90}{\textbf{VDGS}}} &
\includegraphics[width=0.19\textwidth]{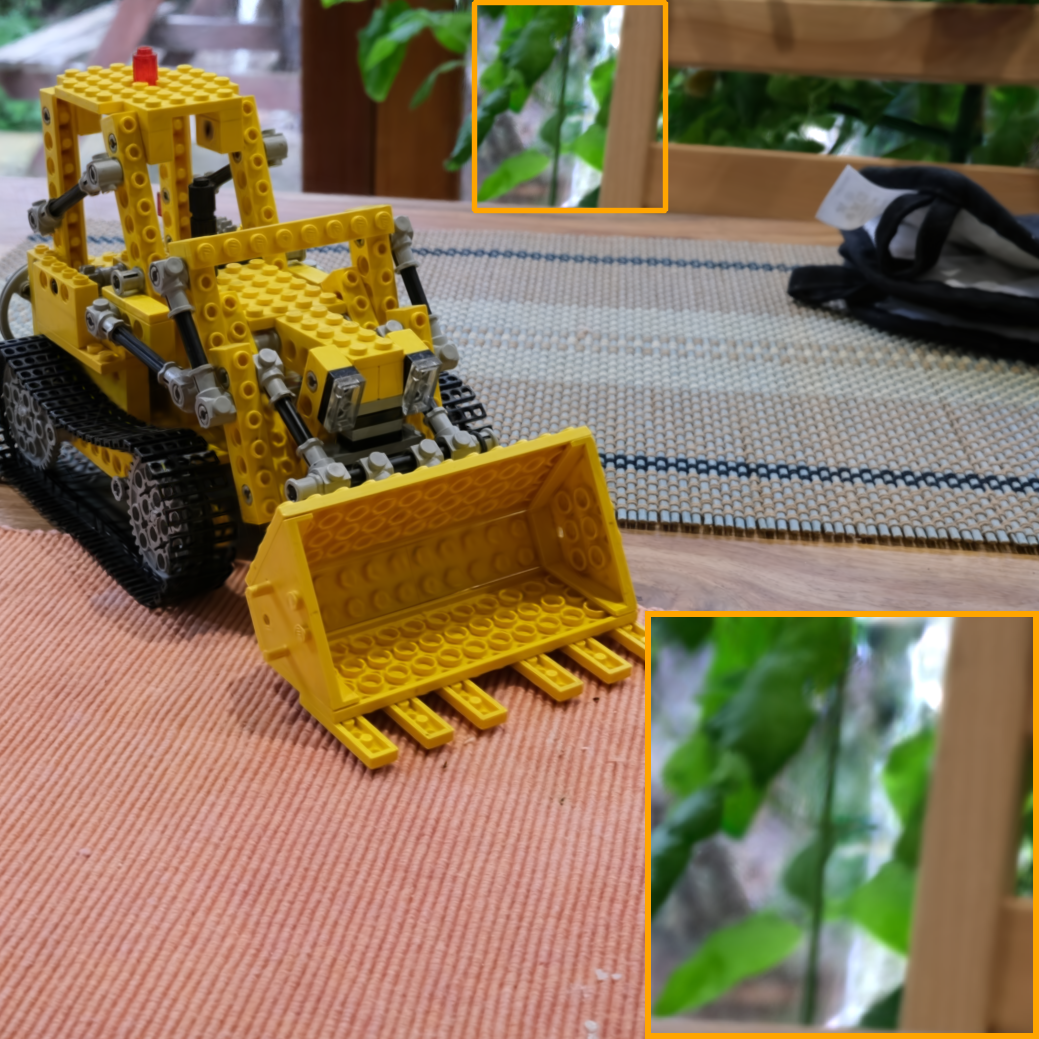} &
\includegraphics[width=0.19\textwidth]{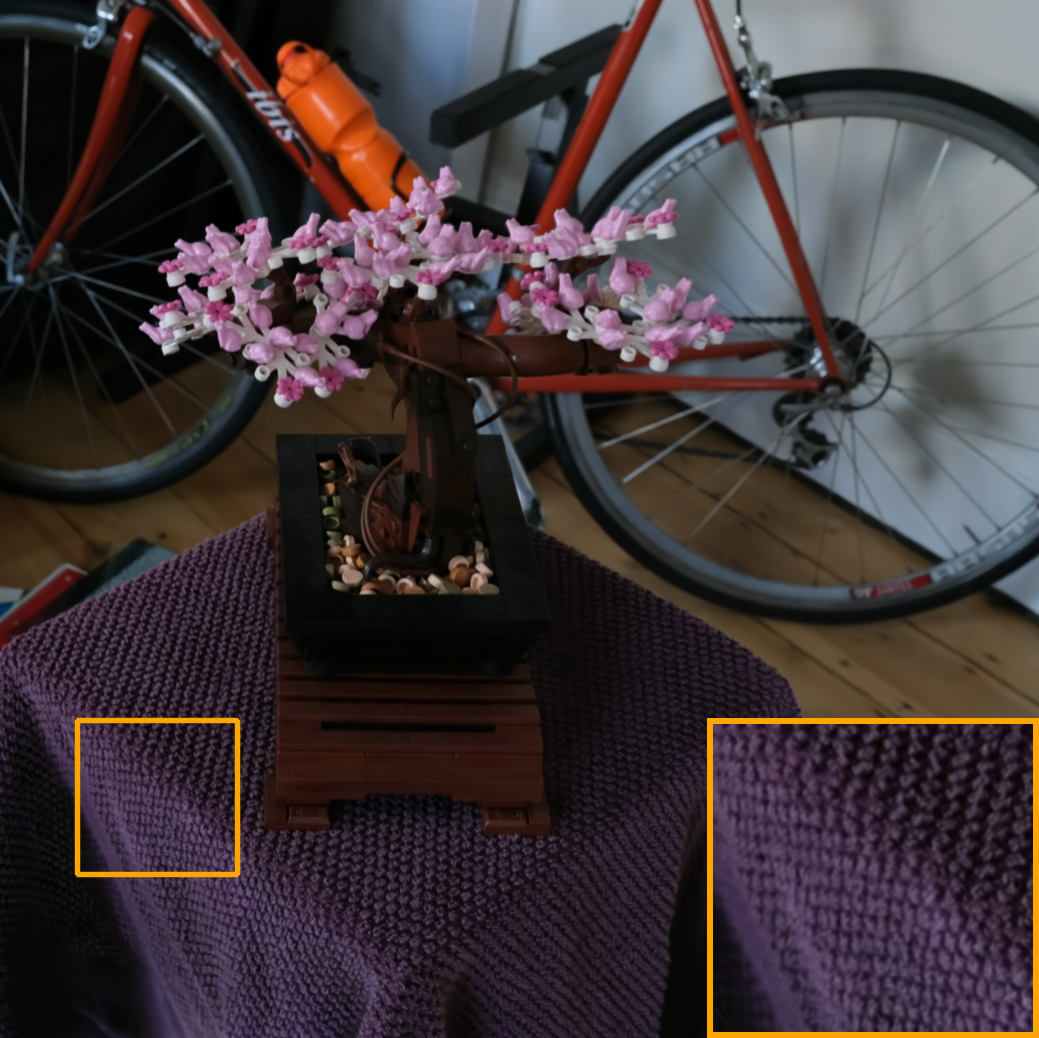} &
\includegraphics[width=0.19\textwidth]{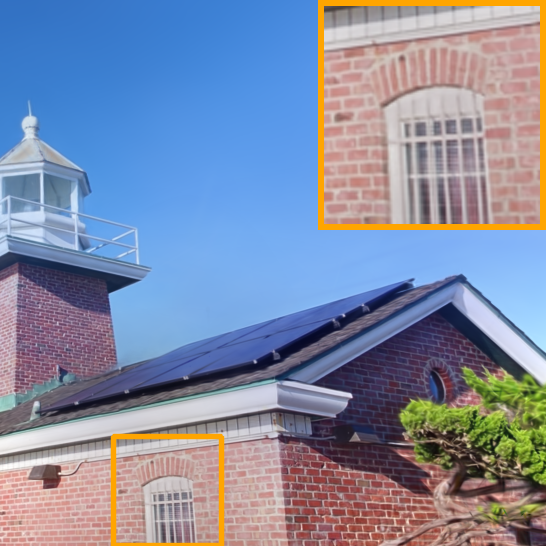} &
\includegraphics[width=0.19\textwidth]{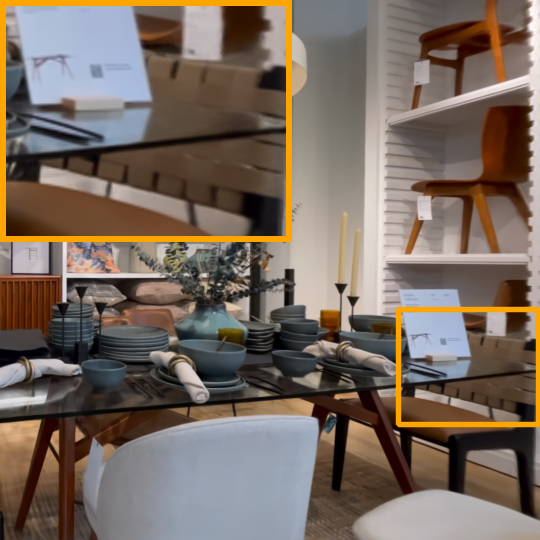} &
\includegraphics[width=0.19\textwidth]{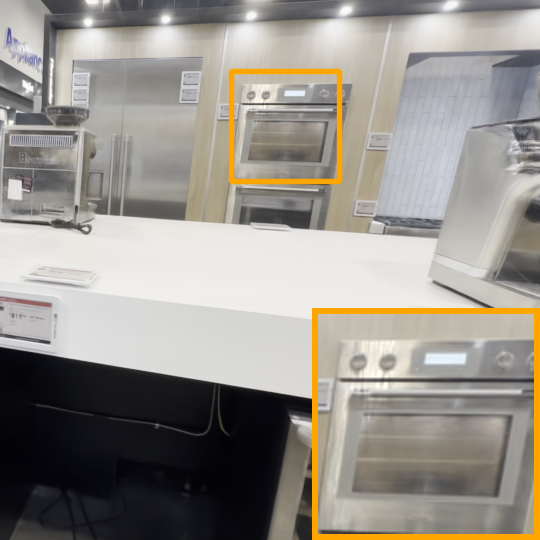} \\

\raisebox{1.0\height}{\rotatebox{90}{\textbf{GSNB}}} &
\includegraphics[width=0.19\textwidth]{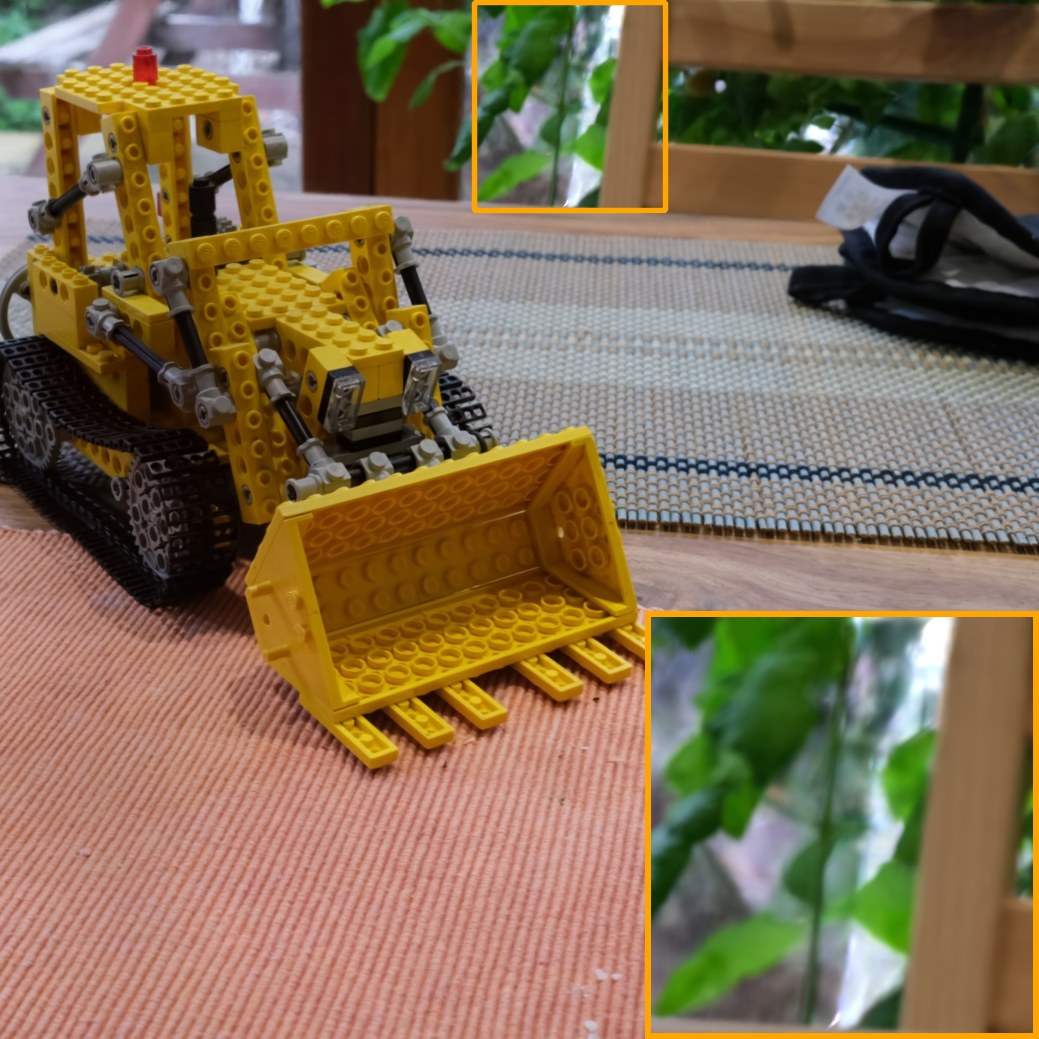} &
\includegraphics[width=0.19\textwidth]{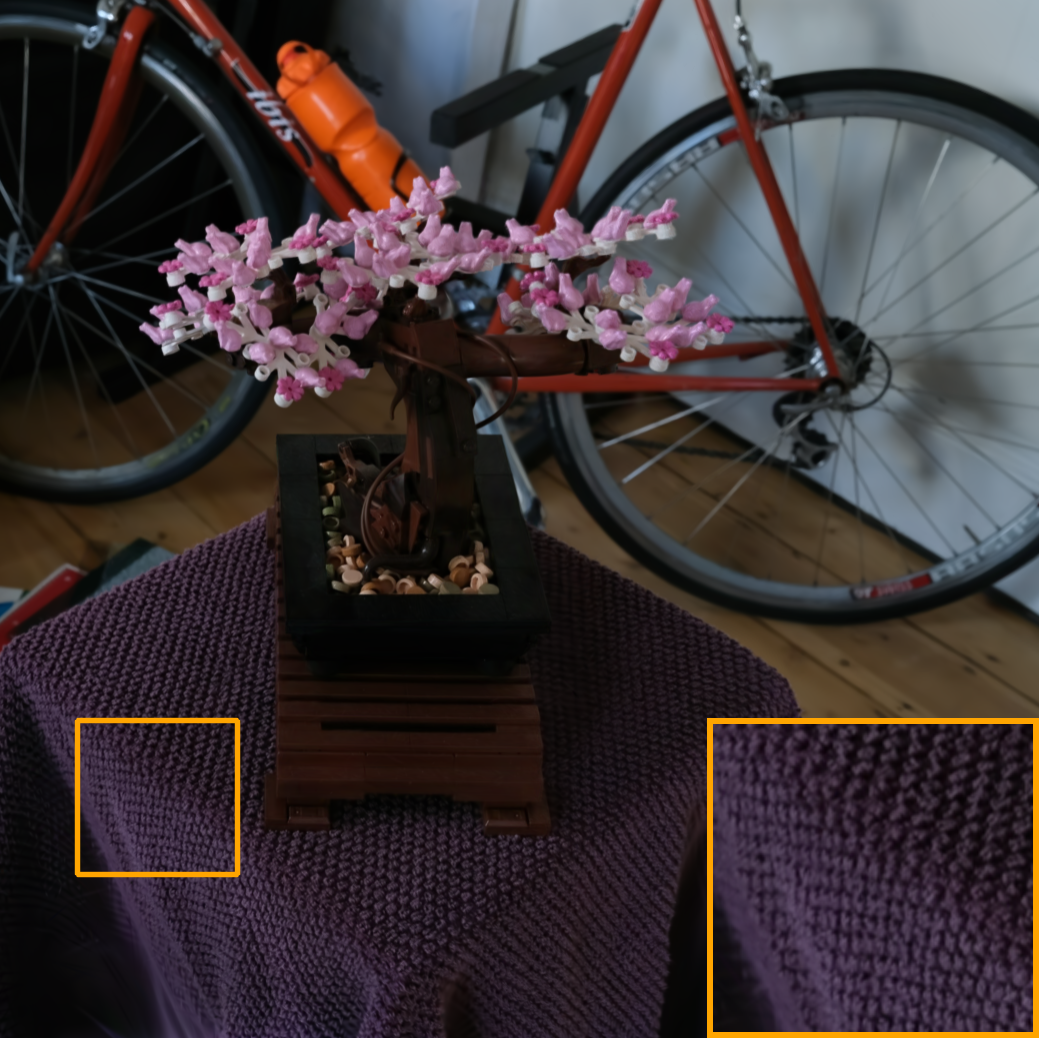} &
\includegraphics[width=0.19\textwidth]{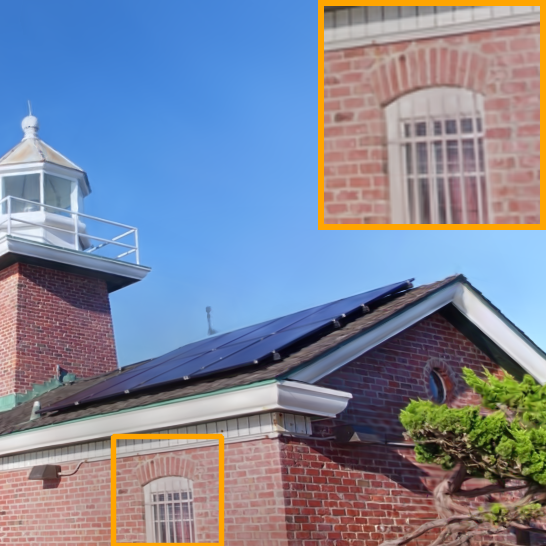} &
\includegraphics[width=0.19\textwidth]{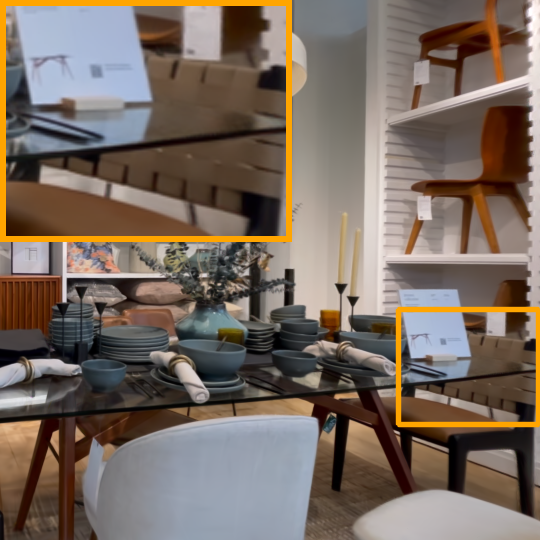} &
\includegraphics[width=0.19\textwidth]{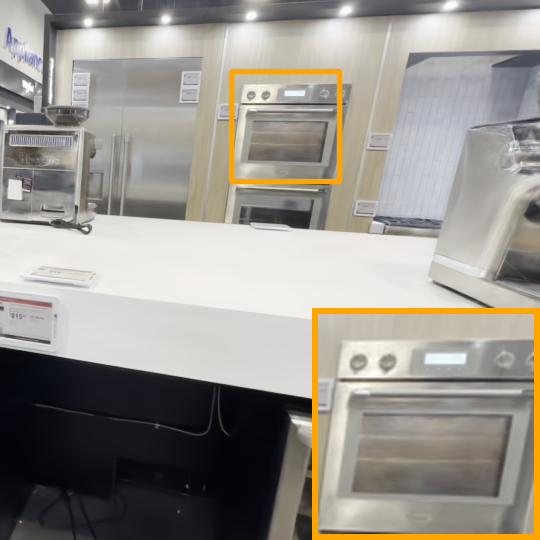} \\

\raisebox{1.0\height}{\rotatebox{90}{\textbf{NeST}}} &
\includegraphics[width=0.19\textwidth]{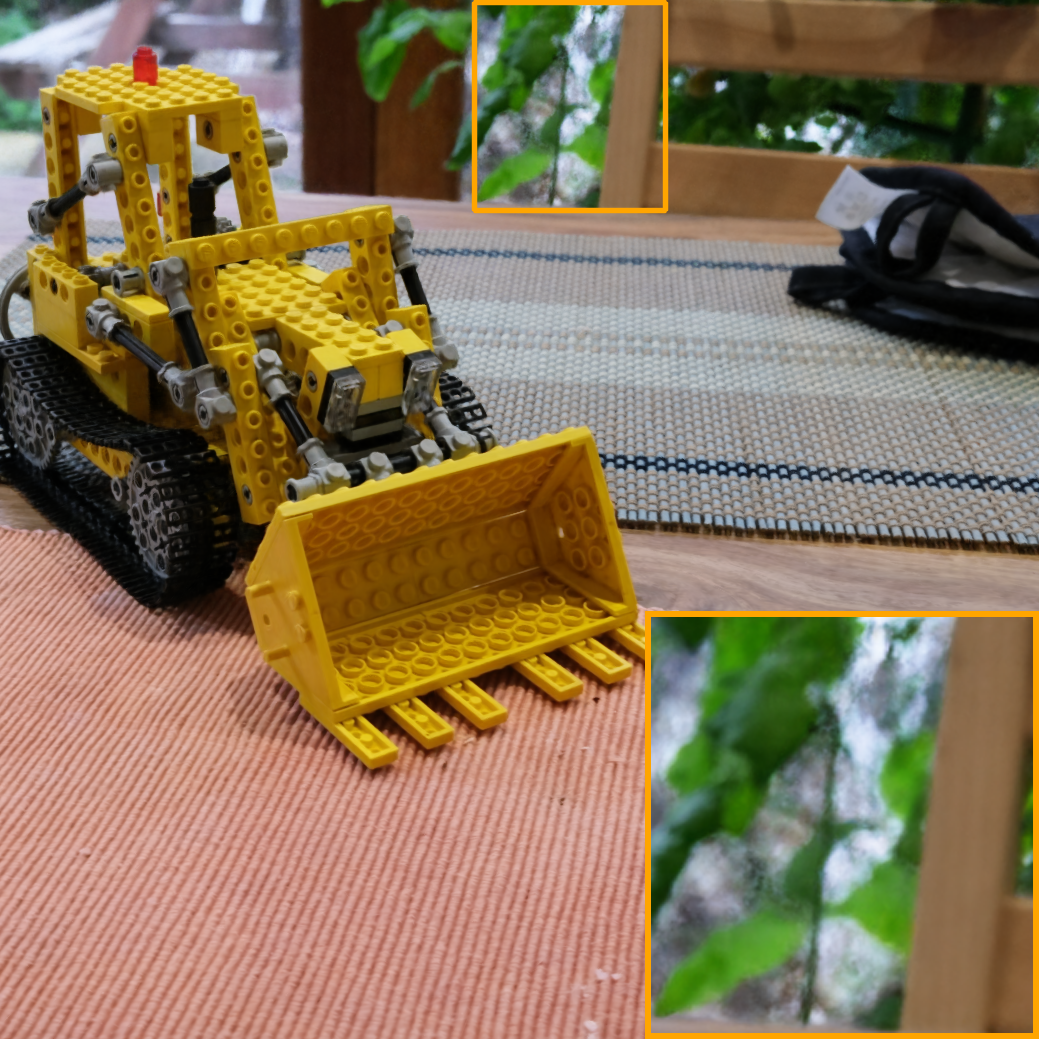} &
\includegraphics[width=0.19\textwidth]{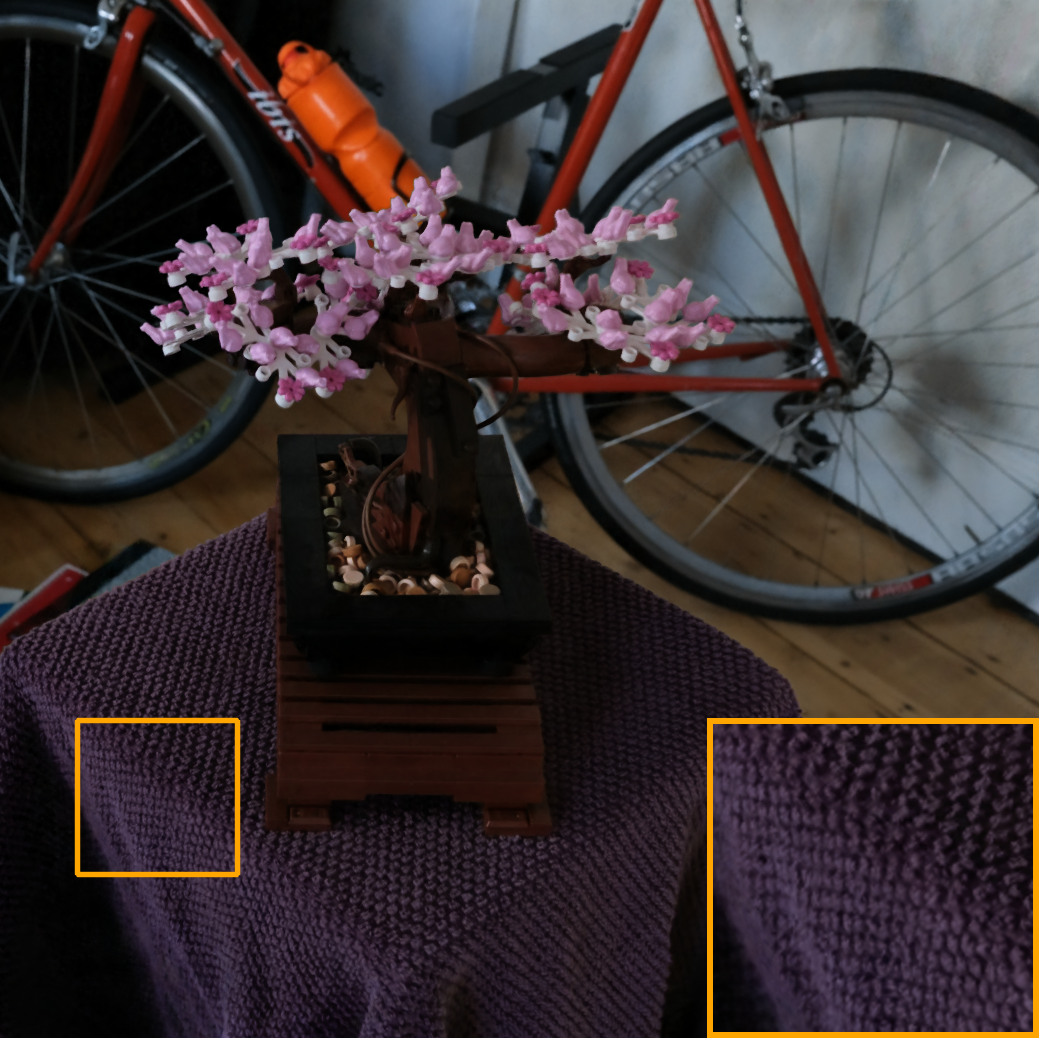} &
\includegraphics[width=0.19\textwidth]{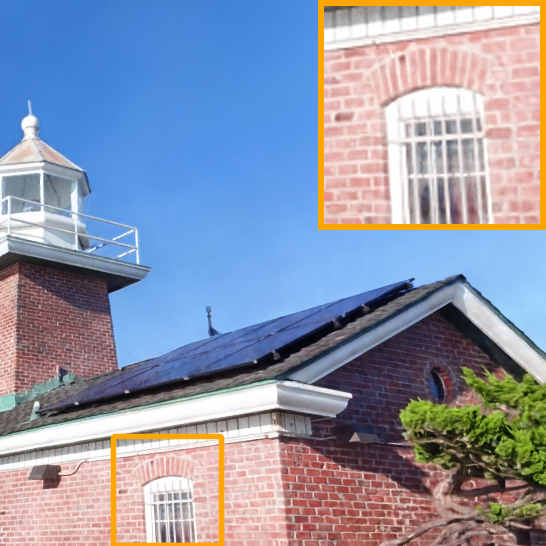} &
\includegraphics[width=0.19\textwidth]{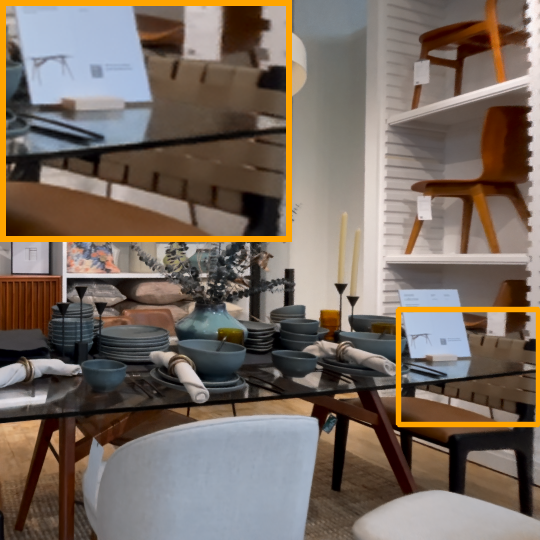} &
\includegraphics[width=0.19\textwidth]{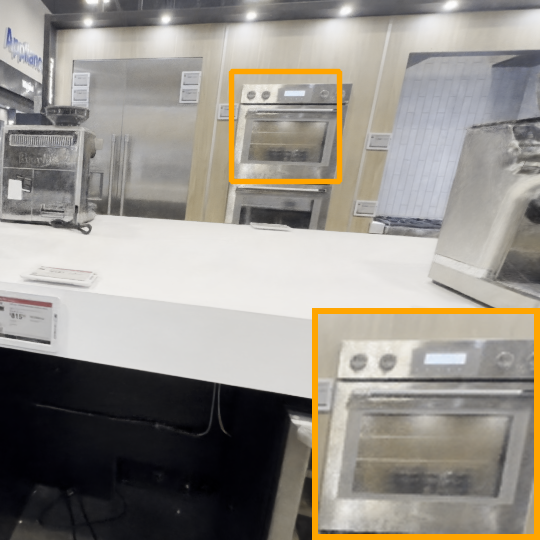} \\

\raisebox{1.0\height}{\rotatebox{90}{\textbf{NTS}}} &
\includegraphics[width=0.19\textwidth]{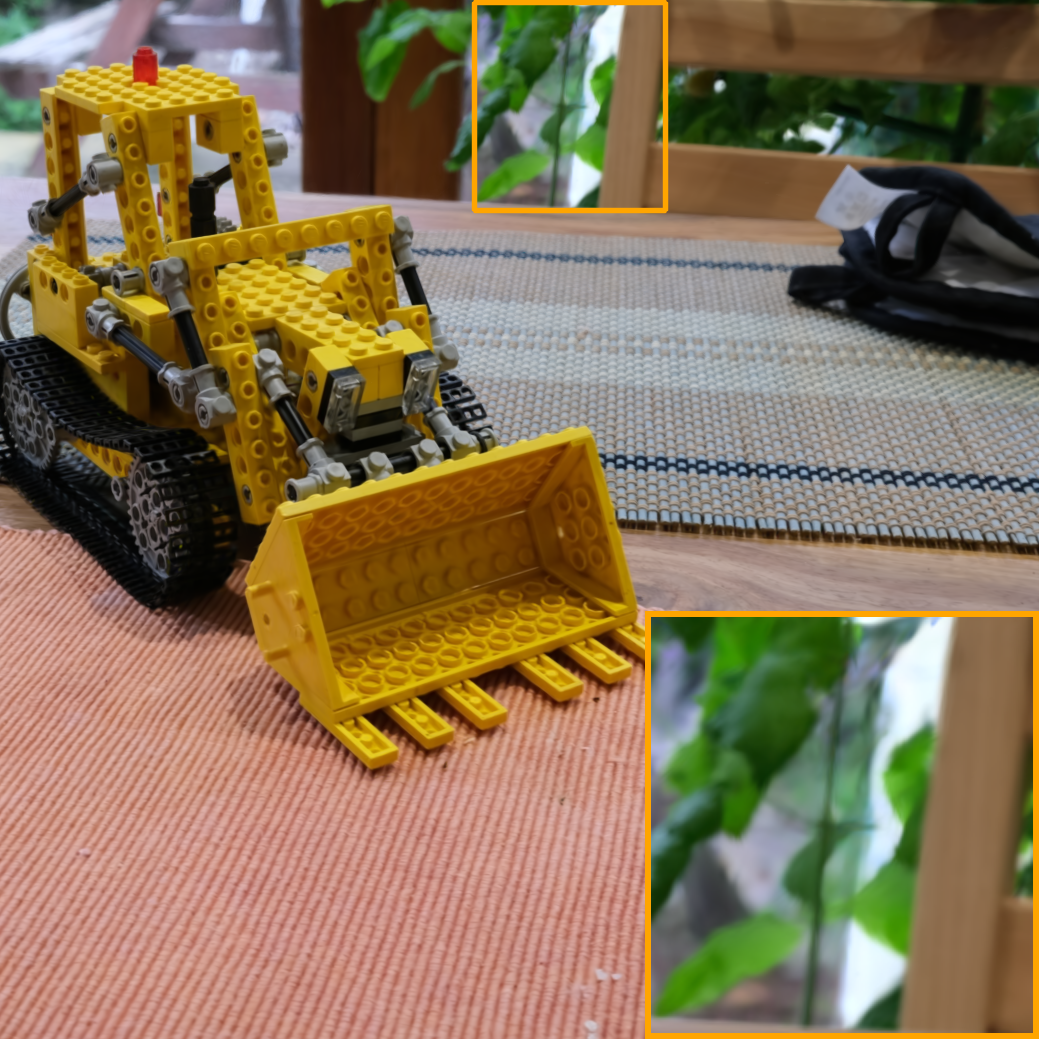} &
\includegraphics[width=0.19\textwidth]{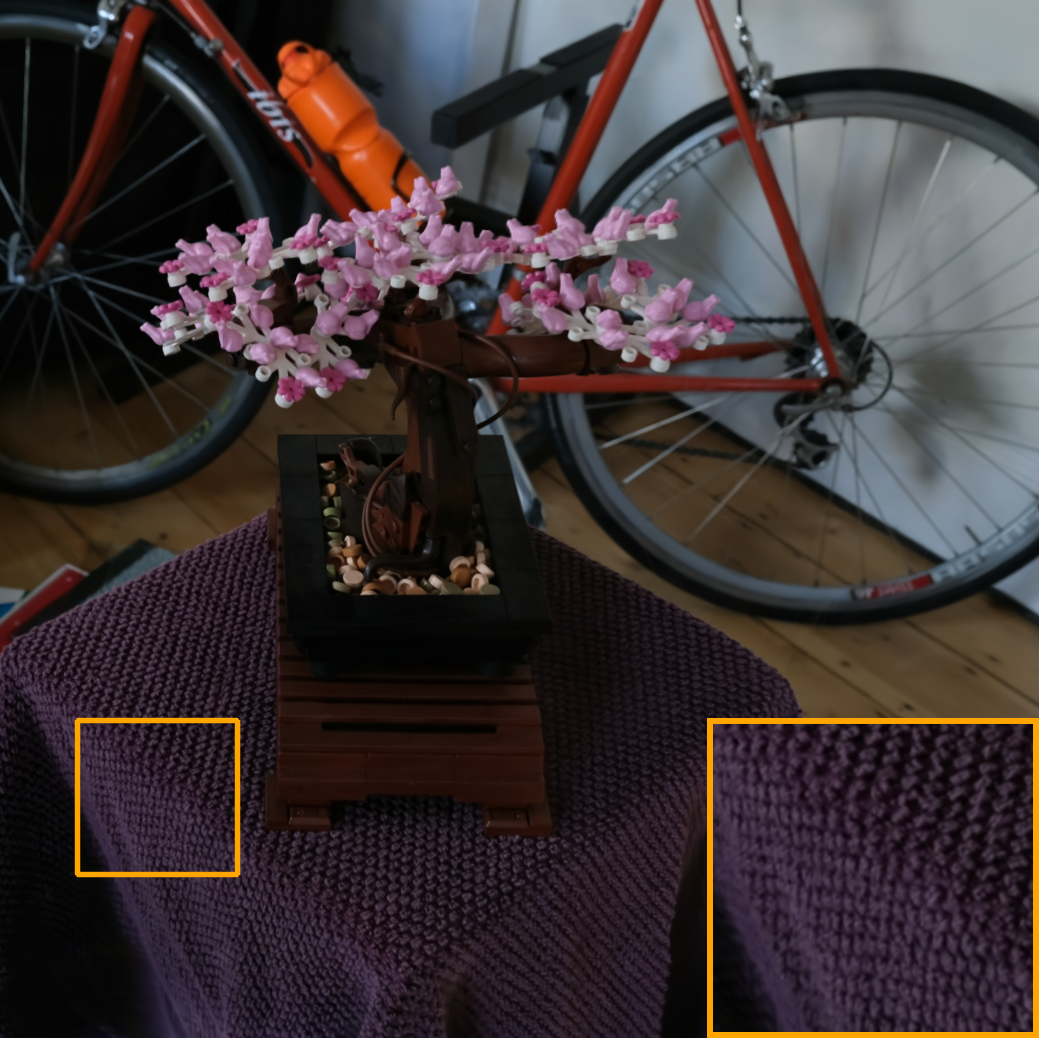} &
\includegraphics[width=0.19\textwidth]{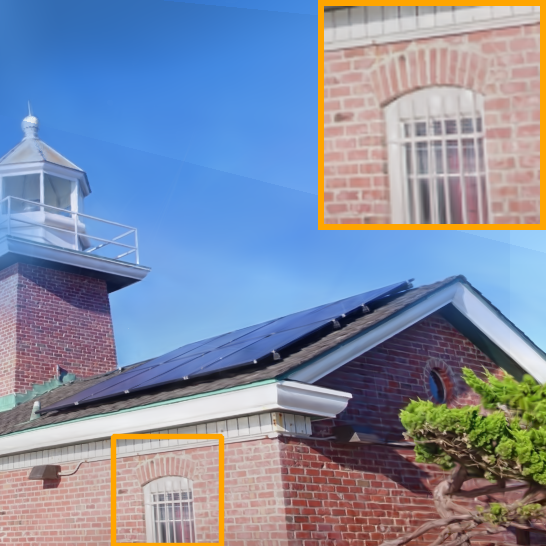} &
\includegraphics[width=0.19\textwidth]{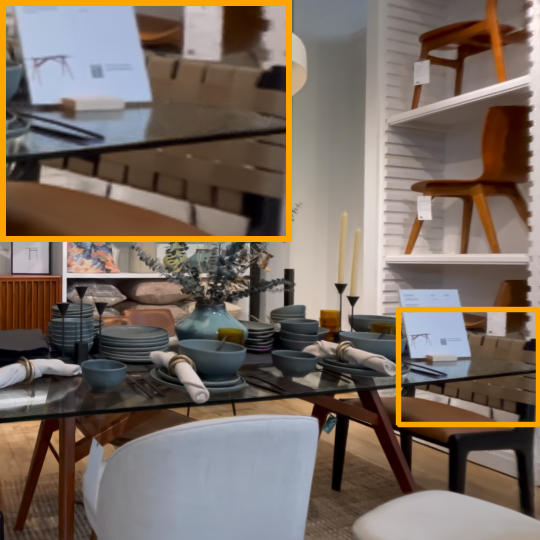} &
\includegraphics[width=0.19\textwidth]{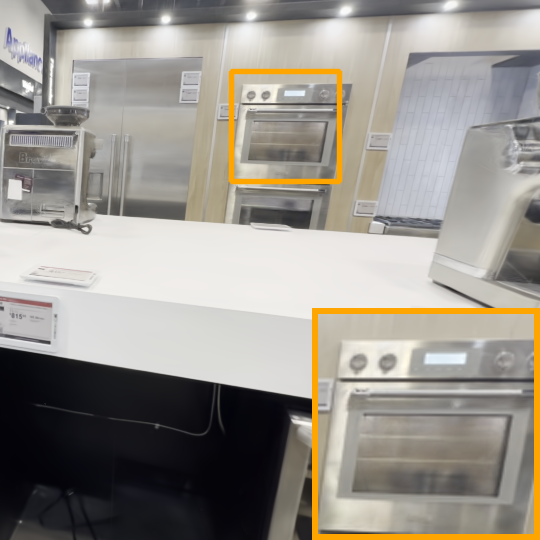} \\

\raisebox{1.0\height}{\rotatebox{90}{\textbf{Ours}}} &
\includegraphics[width=0.19\textwidth]{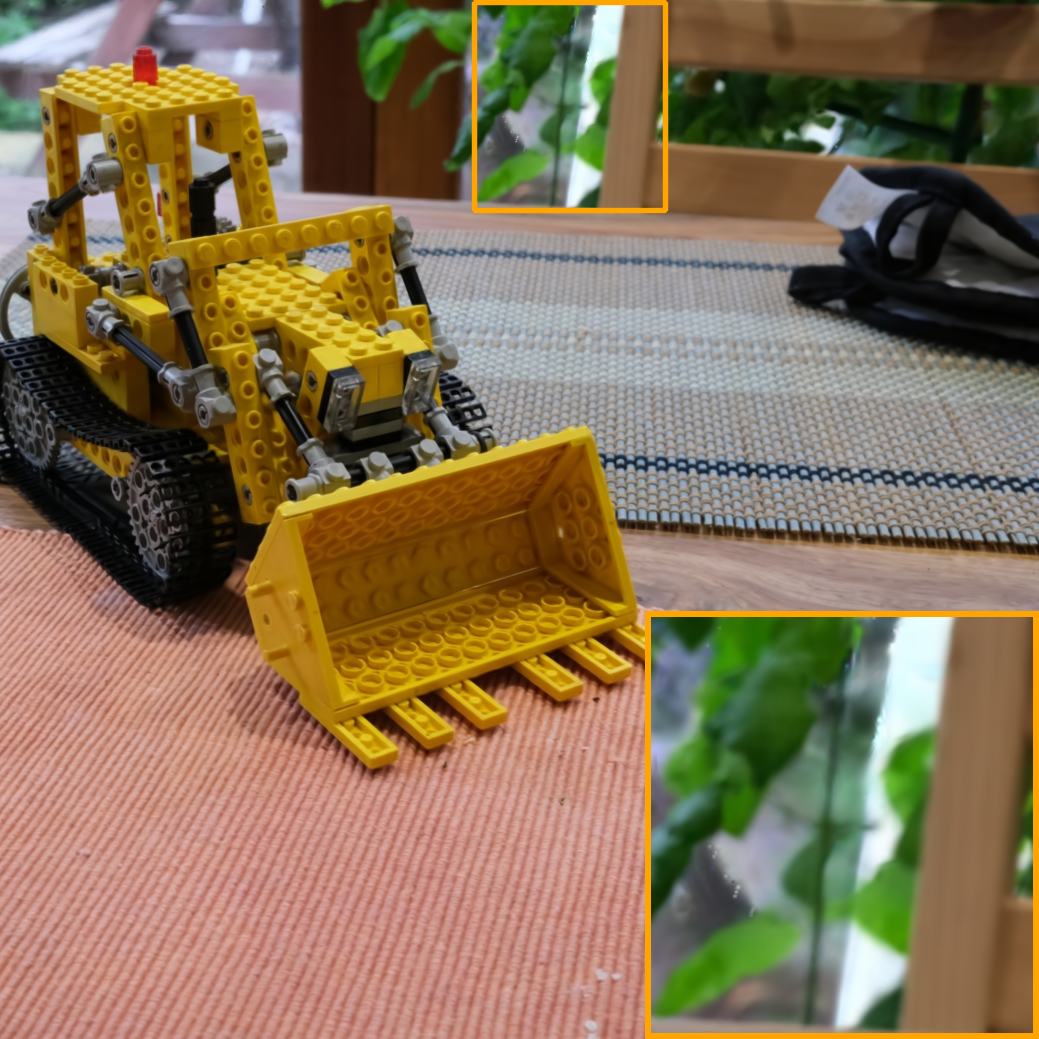} &
\includegraphics[width=0.19\textwidth]{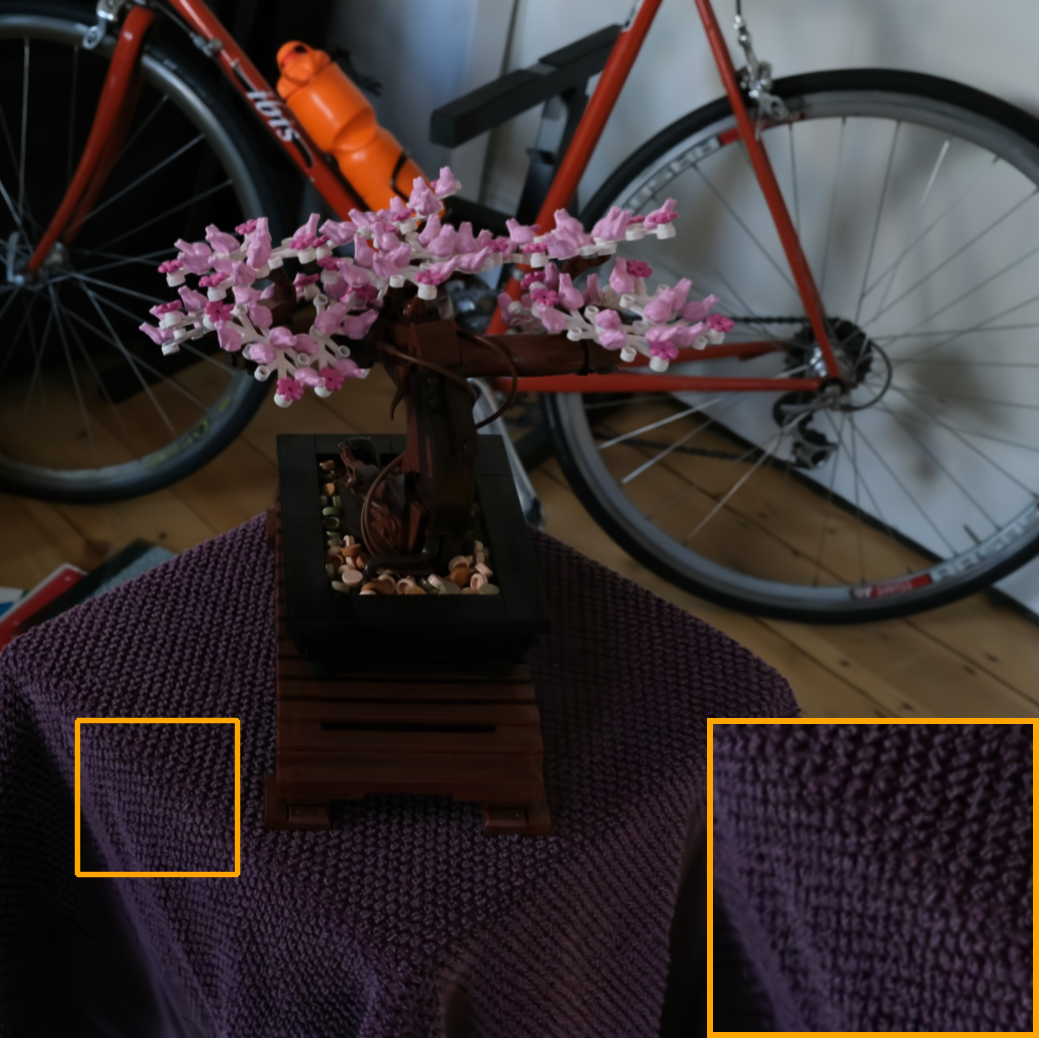} &
\includegraphics[width=0.19\textwidth]{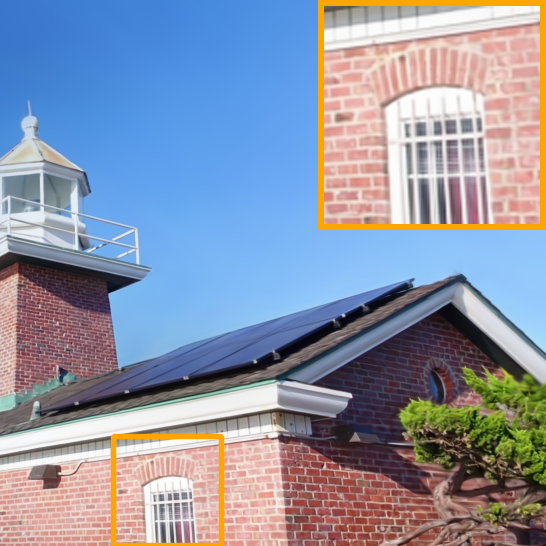} &
\includegraphics[width=0.19\textwidth]{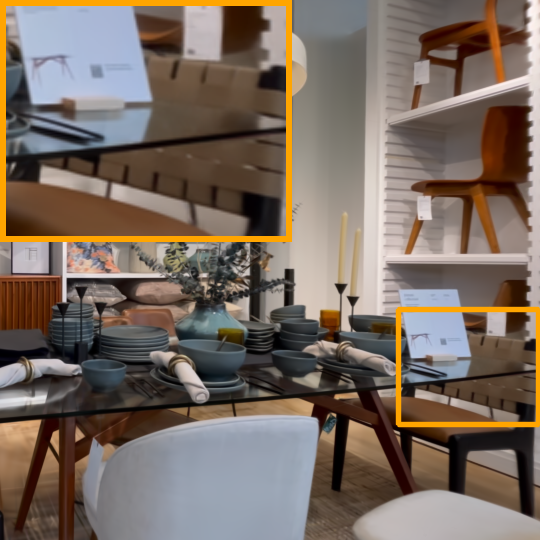} &
\includegraphics[width=0.19\textwidth]{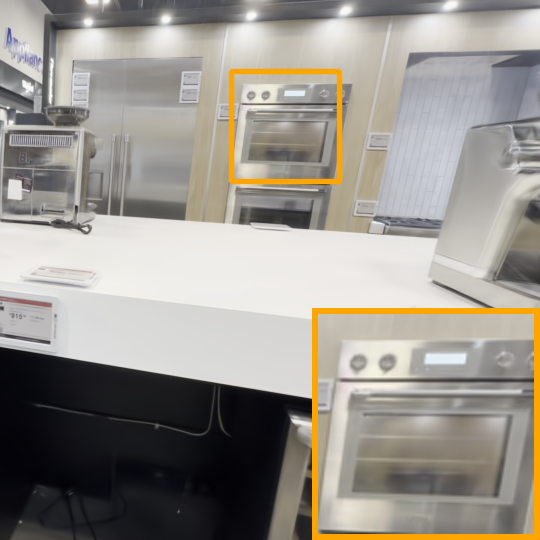} \\

\raisebox{2.0\height}{\rotatebox{90}{\textbf{GT}}} &
\includegraphics[width=0.19\textwidth]{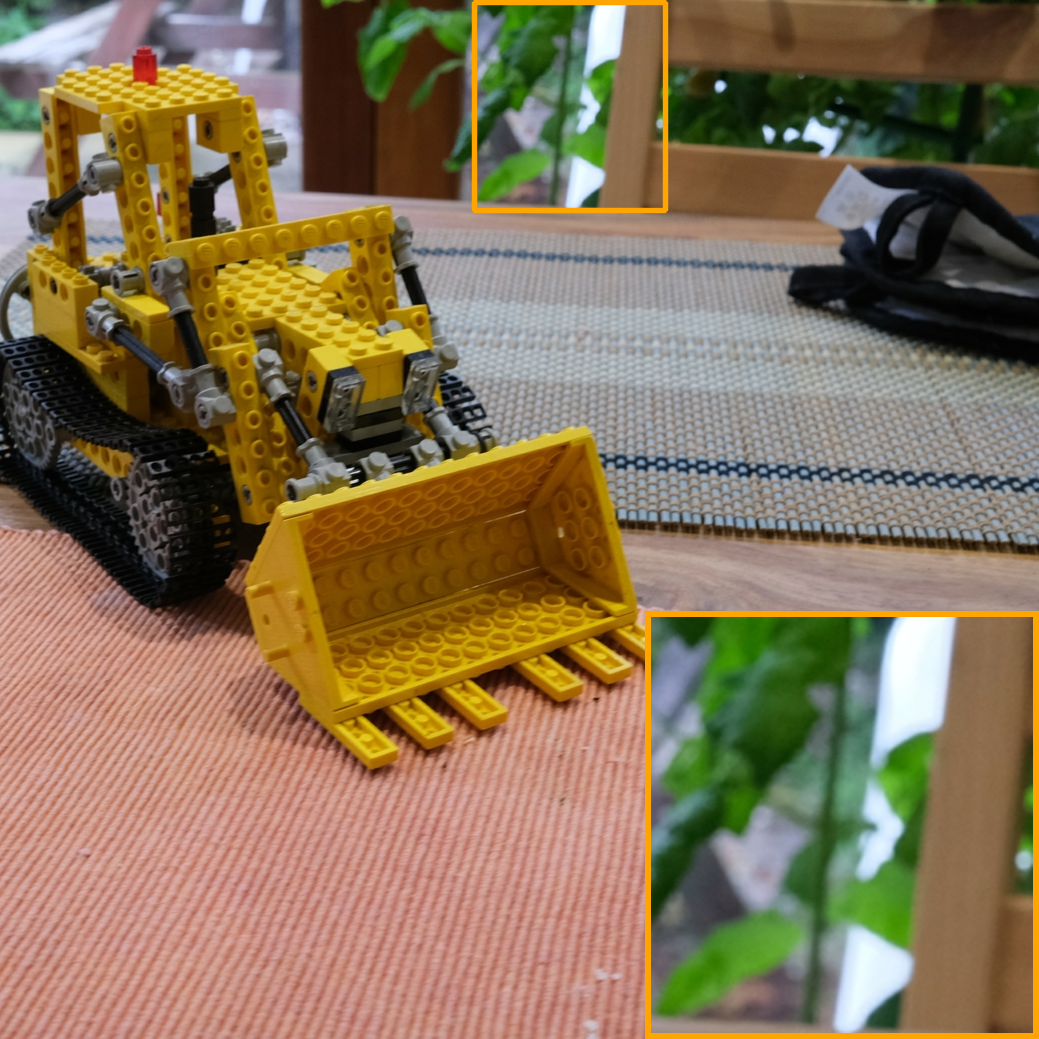} &
\includegraphics[width=0.19\textwidth]{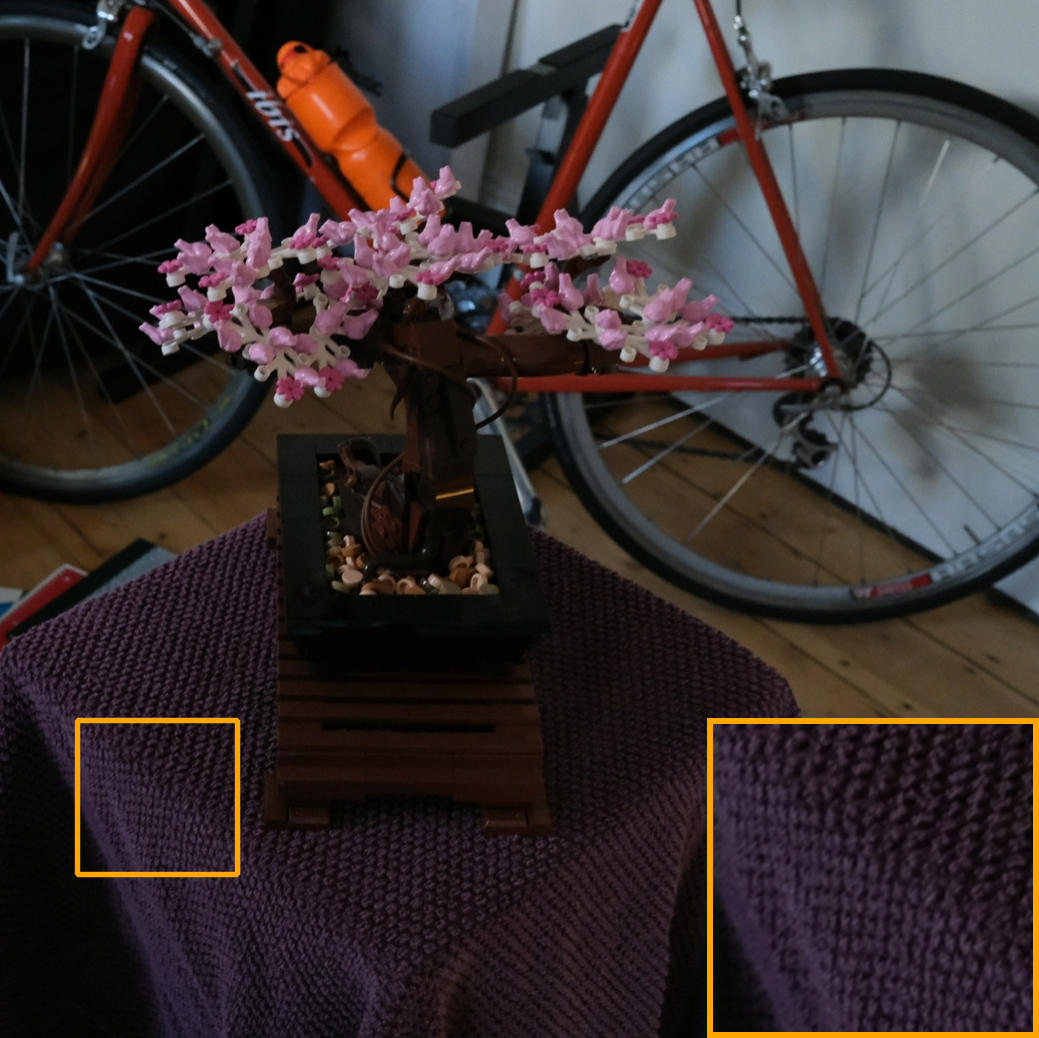} &
\includegraphics[width=0.19\textwidth]{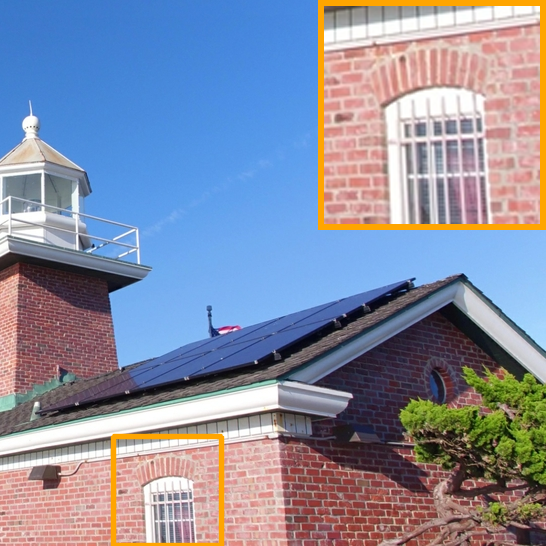} &
\includegraphics[width=0.19\textwidth]{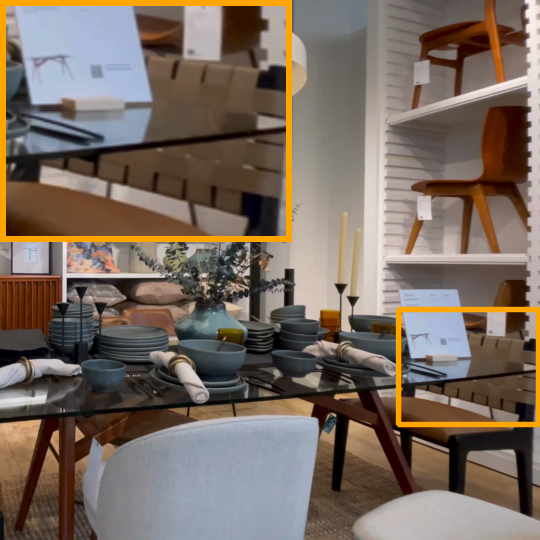} &
\includegraphics[width=0.19\textwidth]{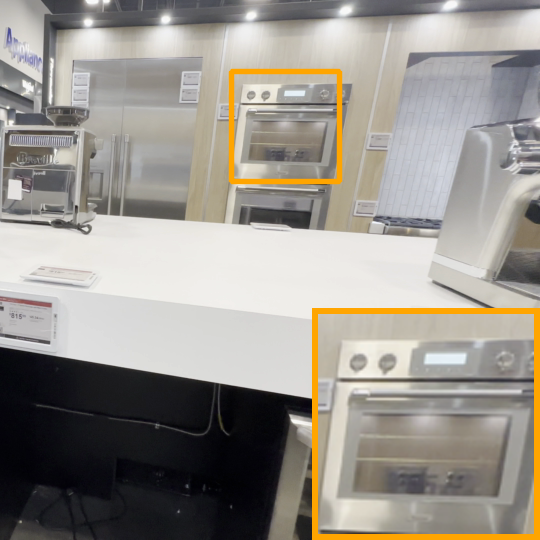} \\

\end{tabular}
}

\caption{Additional qualitative comparison across multiple scenes. MipNeRF360 - scenes 'kitchen' and 'bonsai', TnT - scene 'lighthouse', DL3DV - scenes '5c3a' and '669c'.}
\label{fig:suppl_qual_grid}
\end{figure}

Also, some more qualitative results from all three datasets are shown in Fig.~\ref{fig:suppl_qual_grid}. In addition to the earlier mentioned distinction of better reflectance and far-away details, another specific improvement in color quality can be noticed in the scenes \textbf{bonsai} (cloth and petals) and \textbf{lighthouse} (window) which shows that the color contrast for Our method matches the ground truth better.

We further report per scene level averages together with the standard deviation across multiple runs in Tab.~\ref{tab:mipnerf_per_scene} for MipNeRF 360 dataset. Please note that GSNB went OOM on scenes \emph{bicycle} and \emph{garden}.

\begin{table*}[hp]
\centering
\caption{Per-scene test-set results on MipNeRF360.}
\label{tab:mipnerf_per_scene}
\resizebox{\textwidth}{!}{
\begin{tabular}{llcccccc}
\toprule
Scene & Method & Points (M)$\downarrow$ & Time (min)$\downarrow$ & MS-SSIM$\uparrow$ & LPIPS$\downarrow$ & L1$\downarrow$ & PSNR$\uparrow$ \\
\midrule
\multirow{3}{*}{Bicycle}
& 3DGS & 6.137 $\pm$ 0.076 & 34.08 $\pm$ 0.13 & 0.765 $\pm$ 0.001 & 0.175 $\pm$ 0.001 & 0.039 $\pm$ 0.001 & 25.17 $\pm$ 0.04 \\
& VDGS & 6.441 $\pm$ 0.069 & 58.88 $\pm$ 0.29 & 0.755 $\pm$ 0.002 & 0.195 $\pm$ 0.002 & 0.039 $\pm$ 0.001 & 25.14 $\pm$ 0.14 \\
& NRF-GS & 3.187 $\pm$ 0.024 & 44.12 $\pm$ 0.28 & 0.759 $\pm$ 0.003 & 0.203 $\pm$ 0.006 & 0.041 $\pm$ 0.001 & 25.06 $\pm$ 0.16 \\
\midrule
\multirow{4}{*}{Bonsai}
& 3DGS & 1.257 $\pm$ 0.005 & 20.12 $\pm$ 0.10 & 0.943 $\pm$ 0.001 & 0.133 $\pm$ 0.001 & 0.015 $\pm$ 0.001 & 32.23 $\pm$ 0.10 \\
& VDGS & 1.617 $\pm$ 0.012 & 29.70 $\pm$ 0.33 & 0.945 $\pm$ 0.002 & 0.134 $\pm$ 0.005 & 0.014 $\pm$ 0.000 & 32.72 $\pm$ 0.15 \\
& GSNB & 1.477 $\pm$ 0.042 & 39.15 $\pm$ 0.18 & 0.945 $\pm$ 0.002 & 0.127 $\pm$ 0.003 & 0.014 $\pm$ 0.001 & 33.19 $\pm$ 0.05 \\
& NRF-GS & 0.573 $\pm$ 0.003 & 22.73 $\pm$ 0.12 & 0.944 $\pm$ 0.002 & 0.137 $\pm$ 0.001 & 0.014 $\pm$ 0.001 & 33.24 $\pm$ 0.09 \\
\midrule
\multirow{4}{*}{Counter}
& 3DGS & 1.207 $\pm$ 0.003 & 22.78 $\pm$ 0.08 & 0.909 $\pm$ 0.001 & 0.156 $\pm$ 0.002 & 0.018 $\pm$ 0.001 & 29.02 $\pm$ 0.03 \\
& VDGS & 1.548 $\pm$ 0.007 & 33.50 $\pm$ 0.46 & 0.912 $\pm$ 0.002 & 0.153 $\pm$ 0.001 & 0.018 $\pm$ 0.001 & 29.59 $\pm$ 0.20 \\
& GSNB & 1.282 $\pm$ 0.002 & 36.83 $\pm$ 0.18 & 0.913 $\pm$ 0.001 & 0.147 $\pm$ 0.003 & 0.017 $\pm$ 0.001 & 29.81 $\pm$ 0.03 \\
& NRF-GS & 0.538 $\pm$ 0.001 & 24.98 $\pm$ 0.10 & 0.913 $\pm$ 0.003 & 0.158 $\pm$ 0.003 & 0.017 $\pm$ 0.000 & 29.83 $\pm$ 0.01 \\
\midrule
\multirow{4}{*}{Flowers}
& 3DGS & 3.650 $\pm$ 0.046 & 23.90 $\pm$ 0.15 & 0.600 $\pm$ 0.008 & 0.345 $\pm$ 0.005 & 0.057 $\pm$ 0.001 & 21.40 $\pm$ 0.09 \\
& VDGS & 3.977 $\pm$ 0.060 & 38.83 $\pm$ 1.15 & 0.598 $\pm$ 0.002 & 0.348 $\pm$ 0.006 & 0.057 $\pm$ 0.001 & 21.39 $\pm$ 0.08 \\
& GSNB & 3.400 $\pm$ 0.027 & 69.05 $\pm$ 0.15 & 0.576 $\pm$ 0.004 & 0.357 $\pm$ 0.006 & 0.063 $\pm$ 0.001 & 20.96 $\pm$ 0.05 \\
& NRF-GS & 2.041 $\pm$ 0.019 & 31.75 $\pm$ 0.15 & 0.599 $\pm$ 0.002 & 0.349 $\pm$ 0.010 & 0.057 $\pm$ 0.000 & 21.41 $\pm$ 0.03 \\
\midrule
\multirow{3}{*}{Garden}
& 3DGS & 5.869 $\pm$ 0.013 & 34.60 $\pm$ 1.82 & 0.865 $\pm$ 0.002 & 0.078 $\pm$ 0.002 & 0.028 $\pm$ 0.001 & 27.25 $\pm$ 0.06 \\
& VDGS & 5.701 $\pm$ 0.016 & 55.73 $\pm$ 0.64 & 0.864 $\pm$ 0.003 & 0.081 $\pm$ 0.004 & 0.027 $\pm$ 0.001 & 27.52 $\pm$ 0.08 \\
& NRF-GS & 2.190 $\pm$ 0.001 & 36.50 $\pm$ 0.00 & 0.864 $\pm$ 0.002 & 0.082 $\pm$ 0.002 & 0.026 $\pm$ 0.000 & 27.67 $\pm$ 0.02 \\
\midrule
\multirow{4}{*}{Kitchen}
& 3DGS & 1.818 $\pm$ 0.026 & 27.75 $\pm$ 0.39 & 0.924 $\pm$ 0.002 & 0.097 $\pm$ 0.002 & 0.017 $\pm$ 0.001 & 31.46 $\pm$ 0.02 \\
& VDGS & 2.031 $\pm$ 0.014 & 38.45 $\pm$ 0.22 & 0.927 $\pm$ 0.002 & 0.095 $\pm$ 0.001 & 0.017 $\pm$ 0.001 & 31.62 $\pm$ 0.05 \\
& GSNB & 2.182 $\pm$ 0.061 & 54.88 $\pm$ 0.94 & 0.926 $\pm$ 0.003 & 0.095 $\pm$ 0.002 & 0.017 $\pm$ 0.001 & 31.95 $\pm$ 0.04 \\
& NRF-GS & 0.652 $\pm$ 0.002 & 25.47 $\pm$ 0.06 & 0.927 $\pm$ 0.000 & 0.096 $\pm$ 0.005 & 0.016 $\pm$ 0.000 & 32.10 $\pm$ 0.02 \\
\midrule
\multirow{4}{*}{Room}
& 3DGS & 1.535 $\pm$ 0.018 & 23.32 $\pm$ 0.13 & 0.918 $\pm$ 0.002 & 0.168 $\pm$ 0.001 & 0.017 $\pm$ 0.001 & 31.35 $\pm$ 0.06 \\
& VDGS & 1.845 $\pm$ 0.021 & 34.28 $\pm$ 0.13 & 0.922 $\pm$ 0.001 & 0.165 $\pm$ 0.002 & 0.016 $\pm$ 0.000 & 31.77 $\pm$ 0.03 \\
& GSNB & 1.724 $\pm$ 0.010 & 44.33 $\pm$ 0.45 & 0.919 $\pm$ 0.002 & 0.161 $\pm$ 0.001 & 0.016 $\pm$ 0.001 & 31.55 $\pm$ 0.08 \\
& NRF-GS & 0.802 $\pm$ 0.011 & 25.92 $\pm$ 0.03 & 0.920 $\pm$ 0.001 & 0.168 $\pm$ 0.002 & 0.015 $\pm$ 0.000 & 32.14 $\pm$ 0.05 \\
\midrule
\multirow{4}{*}{Stump}
& 3DGS & 4.934 $\pm$ 0.061 & 27.40 $\pm$ 0.05 & 0.763 $\pm$ 0.002 & 0.165 $\pm$ 0.001 & 0.033 $\pm$ 0.001 & 26.33 $\pm$ 0.12 \\
& VDGS & 4.837 $\pm$ 0.025 & 43.82 $\pm$ 0.10 & 0.761 $\pm$ 0.001 & 0.164 $\pm$ 0.007 & 0.034 $\pm$ 0.001 & 26.50 $\pm$ 0.20 \\
& GSNB & 4.422 $\pm$ 0.048 & 88.97 $\pm$ 0.10 & 0.707 $\pm$ 0.004 & 0.191 $\pm$ 0.001 & 0.041 $\pm$ 0.001 & 24.94 $\pm$ 0.14 \\
& NRF-GS & 2.559 $\pm$ 0.028 & 36.58 $\pm$ 0.29 & 0.762 $\pm$ 0.001 & 0.178 $\pm$ 0.003 & 0.034 $\pm$ 0.001 & 26.39 $\pm$ 0.04 \\
\midrule
\multirow{4}{*}{Treehill}
& 3DGS & 3.831 $\pm$ 0.023 & 24.27 $\pm$ 0.03 & 0.632 $\pm$ 0.001 & 0.332 $\pm$ 0.003 & 0.047 $\pm$ 0.001 & 22.42 $\pm$ 0.02 \\
& VDGS & 3.938 $\pm$ 0.016 & 38.28 $\pm$ 0.19 & 0.632 $\pm$ 0.002 & 0.337 $\pm$ 0.006 & 0.047 $\pm$ 0.001 & 22.68 $\pm$ 0.03 \\
& GSNB & 4.181 $\pm$ 0.037 & 80.22 $\pm$ 0.19 & 0.613 $\pm$ 0.012 & 0.332 $\pm$ 0.009 & 0.053 $\pm$ 0.002 & 21.60 $\pm$ 0.06 \\
& NRF-GS & 2.361 $\pm$ 0.019 & 34.40 $\pm$ 0.17 & 0.636 $\pm$ 0.001 & 0.350 $\pm$ 0.000 & 0.045 $\pm$ 0.000 & 22.84 $\pm$ 0.01 \\
\bottomrule
\end{tabular}}
\end{table*}

\newpage
Tab.~\ref{tab:dl3dv_per_scene} shows the per scene level average results on DL3DV dataset with standard deviations across all metrics.

\begin{table*}[h]
\centering
\caption{Per-scene test-set results on DL3DV.}
\label{tab:dl3dv_per_scene}
\resizebox{\textwidth}{!}{
\begin{tabular}{llcccccc}
\toprule
Scene & Method & Points (M)$\downarrow$ & Time (min)$\downarrow$ & MS-SSIM$\uparrow$ & LPIPS$\downarrow$ & L1$\downarrow$ & PSNR$\uparrow$ \\
\midrule
\multirow{4}{*}{06da}
& 3DGS & 1.330 $\pm$ 0.007 & 12.85 $\pm$ 0.09 & 0.905 $\pm$ 0.001 & 0.107 $\pm$ 0.001 & 0.024 $\pm$ 0.001 & 28.21 $\pm$ 0.01 \\
& VDGS & 1.687 $\pm$ 0.012 & 22.30 $\pm$ 0.05 & 0.906 $\pm$ 0.002 & 0.110 $\pm$ 0.002 & 0.023 $\pm$ 0.001 & 28.41 $\pm$ 0.13 \\
& GSNB & 1.404 $\pm$ 0.007 & 31.57 $\pm$ 0.50 & 0.908 $\pm$ 0.002 & 0.102 $\pm$ 0.003 & 0.021 $\pm$ 0.001 & 29.34 $\pm$ 0.01 \\
& NRF-GS & 0.703 $\pm$ 0.009 & 15.13 $\pm$ 0.13 & 0.928 $\pm$ 0.000 & 0.098 $\pm$ 0.001 & 0.017 $\pm$ 0.001 & 30.91 $\pm$ 0.03 \\
\midrule
\multirow{4}{*}{093e}
& 3DGS & 0.897 $\pm$ 0.011 & 11.05 $\pm$ 0.13 & 0.952 $\pm$ 0.001 & 0.073 $\pm$ 0.001 & 0.016 $\pm$ 0.000 & 31.21 $\pm$ 0.02 \\
& VDGS & 0.957 $\pm$ 0.004 & 16.28 $\pm$ 0.10 & 0.955 $\pm$ 0.002 & 0.071 $\pm$ 0.001 & 0.016 $\pm$ 0.000 & 31.46 $\pm$ 0.15 \\
& GSNB & 0.812 $\pm$ 0.007 & 22.00 $\pm$ 0.10 & 0.956 $\pm$ 0.001 & 0.067 $\pm$ 0.002 & 0.016 $\pm$ 0.001 & 32.22 $\pm$ 0.08 \\
& NRF-GS & 0.377 $\pm$ 0.001 & 11.53 $\pm$ 0.20 & 0.953 $\pm$ 0.002 & 0.079 $\pm$ 0.003 & 0.016 $\pm$ 0.001 & 31.83 $\pm$ 0.02 \\
\midrule
\multirow{4}{*}{0a1b}
& 3DGS & 0.975 $\pm$ 0.003 & 10.82 $\pm$ 0.06 & 0.966 $\pm$ 0.001 & 0.039 $\pm$ 0.001 & 0.012 $\pm$ 0.001 & 33.43 $\pm$ 0.02 \\
& VDGS & 1.021 $\pm$ 0.015 & 15.88 $\pm$ 0.13 & 0.968 $\pm$ 0.001 & 0.038 $\pm$ 0.000 & 0.011 $\pm$ 0.001 & 33.76 $\pm$ 0.22 \\
& GSNB & 0.889 $\pm$ 0.019 & 22.18 $\pm$ 0.13 & 0.967 $\pm$ 0.001 & 0.038 $\pm$ 0.001 & 0.011 $\pm$ 0.000 & 34.51 $\pm$ 0.03 \\
& NRF-GS & 0.422 $\pm$ 0.007 & 11.90 $\pm$ 0.09 & 0.968 $\pm$ 0.001 & 0.040 $\pm$ 0.001 & 0.011 $\pm$ 0.000 & 33.95 $\pm$ 0.02 \\
\midrule
\multirow{4}{*}{0a48}
& 3DGS & 1.047 $\pm$ 0.005 & 12.57 $\pm$ 0.03 & 0.945 $\pm$ 0.001 & 0.053 $\pm$ 0.002 & 0.017 $\pm$ 0.001 & 31.02 $\pm$ 0.21 \\
& VDGS & 1.277 $\pm$ 0.007 & 21.23 $\pm$ 0.06 & 0.949 $\pm$ 0.001 & 0.051 $\pm$ 0.001 & 0.016 $\pm$ 0.000 & 31.47 $\pm$ 0.55 \\
& GSNB & 0.835 $\pm$ 0.002 & 21.85 $\pm$ 0.05 & 0.946 $\pm$ 0.001 & 0.056 $\pm$ 0.001 & 0.017 $\pm$ 0.000 & 32.74 $\pm$ 0.11 \\
& NRF-GS & 0.518 $\pm$ 0.005 & 13.92 $\pm$ 0.10 & 0.950 $\pm$ 0.002 & 0.054 $\pm$ 0.001 & 0.015 $\pm$ 0.001 & 32.56 $\pm$ 0.17 \\
\midrule
\multirow{4}{*}{119f}
& 3DGS & 1.070 $\pm$ 0.028 & 11.02 $\pm$ 0.08 & 0.864 $\pm$ 0.001 & 0.167 $\pm$ 0.001 & 0.028 $\pm$ 0.001 & 27.17 $\pm$ 0.03 \\
& VDGS & 1.209 $\pm$ 0.007 & 17.90 $\pm$ 0.05 & 0.867 $\pm$ 0.002 & 0.169 $\pm$ 0.001 & 0.028 $\pm$ 0.001 & 27.36 $\pm$ 0.12 \\
& GSNB & 1.131 $\pm$ 0.061 & 25.68 $\pm$ 0.43 & 0.863 $\pm$ 0.001 & 0.170 $\pm$ 0.001 & 0.030 $\pm$ 0.002 & 27.33 $\pm$ 0.03 \\
& NRF-GS & 0.551 $\pm$ 0.001 & 13.18 $\pm$ 0.06 & 0.874 $\pm$ 0.001 & 0.170 $\pm$ 0.001 & 0.025 $\pm$ 0.001 & 27.96 $\pm$ 0.04 \\
\midrule
\multirow{4}{*}{387e}
& 3DGS & 1.866 $\pm$ 0.004 & 18.45 $\pm$ 0.05 & 0.953 $\pm$ 0.000 & 0.054 $\pm$ 0.001 & 0.017 $\pm$ 0.001 & 32.20 $\pm$ 0.07 \\
& VDGS & 2.288 $\pm$ 0.019 & 32.65 $\pm$ 0.31 & 0.954 $\pm$ 0.000 & 0.056 $\pm$ 0.001 & 0.015 $\pm$ 0.001 & 33.23 $\pm$ 0.10 \\
& GSNB & 1.441 $\pm$ 0.042 & 33.82 $\pm$ 0.16 & 0.957 $\pm$ 0.002 & 0.054 $\pm$ 0.000 & 0.014 $\pm$ 0.001 & 33.46 $\pm$ 0.04 \\
& NRF-GS & 0.885 $\pm$ 0.007 & 19.42 $\pm$ 0.08 & 0.955 $\pm$ 0.001 & 0.057 $\pm$ 0.002 & 0.013 $\pm$ 0.000 & 33.92 $\pm$ 0.02 \\
\midrule
\multirow{4}{*}{5c3a}
& 3DGS & 0.938 $\pm$ 0.004 & 11.95 $\pm$ 0.05 & 0.949 $\pm$ 0.001 & 0.038 $\pm$ 0.002 & 0.016 $\pm$ 0.001 & 31.15 $\pm$ 0.04 \\
& VDGS & 1.099 $\pm$ 0.009 & 18.35 $\pm$ 0.05 & 0.953 $\pm$ 0.001 & 0.035 $\pm$ 0.001 & 0.014 $\pm$ 0.001 & 31.97 $\pm$ 0.06 \\
& GSNB & 0.706 $\pm$ 0.009 & 18.85 $\pm$ 0.13 & 0.956 $\pm$ 0.001 & 0.032 $\pm$ 0.001 & 0.013 $\pm$ 0.001 & 32.65 $\pm$ 0.05 \\
& NRF-GS & 0.361 $\pm$ 0.002 & 11.85 $\pm$ 0.13 & 0.956 $\pm$ 0.001 & 0.035 $\pm$ 0.001 & 0.013 $\pm$ 0.000 & 32.51 $\pm$ 0.04 \\
\midrule
\multirow{4}{*}{669c}
& 3DGS & 0.926 $\pm$ 0.004 & 11.35 $\pm$ 0.05 & 0.925 $\pm$ 0.000 & 0.086 $\pm$ 0.001 & 0.021 $\pm$ 0.001 & 27.88 $\pm$ 0.02 \\
& VDGS & 1.141 $\pm$ 0.003 & 19.12 $\pm$ 0.08 & 0.929 $\pm$ 0.001 & 0.085 $\pm$ 0.001 & 0.020 $\pm$ 0.001 & 28.43 $\pm$ 0.16 \\
& GSNB & 0.853 $\pm$ 0.002 & 22.15 $\pm$ 0.05 & 0.935 $\pm$ 0.001 & 0.079 $\pm$ 0.001 & 0.018 $\pm$ 0.001 & 29.75 $\pm$ 0.05 \\
& NRF-GS & 0.463 $\pm$ 0.001 & 13.42 $\pm$ 0.03 & 0.935 $\pm$ 0.001 & 0.082 $\pm$ 0.002 & 0.017 $\pm$ 0.000 & 29.95 $\pm$ 0.05 \\
\midrule
\multirow{4}{*}{e2ce}
& 3DGS & 1.372 $\pm$ 0.008 & 14.62 $\pm$ 0.08 & 0.790 $\pm$ 0.001 & 0.178 $\pm$ 0.003 & 0.042 $\pm$ 0.001 & 23.11 $\pm$ 0.03 \\
& VDGS & 1.953 $\pm$ 0.003 & 26.62 $\pm$ 0.12 & 0.788 $\pm$ 0.001 & 0.185 $\pm$ 0.001 & 0.041 $\pm$ 0.000 & 23.26 $\pm$ 0.01 \\
& GSNB & 1.226 $\pm$ 0.016 & 29.40 $\pm$ 0.09 & 0.798 $\pm$ 0.003 & 0.177 $\pm$ 0.006 & 0.042 $\pm$ 0.001 & 23.34 $\pm$ 0.04 \\
& NRF-GS & 1.074 $\pm$ 0.006 & 20.33 $\pm$ 0.06 & 0.808 $\pm$ 0.002 & 0.180 $\pm$ 0.001 & 0.038 $\pm$ 0.000 & 23.86 $\pm$ 0.01 \\
\bottomrule
\end{tabular}}
\end{table*}

\newpage
Tab.~\ref{tab:tnt_per_scene} and Tab.~\ref{tab:tnt_per_scene_cont2} are the 19 scenes from 'Tanks and Temples' dataset taken from NerfBaselines package. We report the scene level average and standard deviation for all metrics.

\begin{table*}[hp]
\centering
\caption{Per-scene test-set results on Tanks and Temples.}
\label{tab:tnt_per_scene}
\resizebox{\textwidth}{!}{
\begin{tabular}{llcccccc}
\toprule
Scene & Method & Points (M)$\downarrow$ & Time (min)$\downarrow$ & MS-SSIM$\uparrow$ & LPIPS$\downarrow$ & L1$\downarrow$ & PSNR$\uparrow$ \\
\midrule
\multirow{4}{*}{Auditorium}
& 3DGS & 0.735 $\pm$ 0.005 & 10.92 $\pm$ 0.08 & 0.871 $\pm$ 0.002 & 0.205 $\pm$ 0.005 & 0.044 $\pm$ 0.001 & 24.28 $\pm$ 0.03 \\
& VDGS & 0.856 $\pm$ 0.002 & 17.42 $\pm$ 0.08 & 0.858 $\pm$ 0.003 & 0.209 $\pm$ 0.001 & 0.047 $\pm$ 0.001 & 24.10 $\pm$ 0.06 \\
& GSNB & 0.765 $\pm$ 0.004 & 21.17 $\pm$ 0.06 & 0.867 $\pm$ 0.002 & 0.199 $\pm$ 0.001 & 0.046 $\pm$ 0.001 & 24.25 $\pm$ 0.04 \\
& NRF-GS & 0.375 $\pm$ 0.003 & 12.17 $\pm$ 0.06 & 0.886 $\pm$ 0.002 & 0.193 $\pm$ 0.001 & 0.038 $\pm$ 0.001 & 25.69 $\pm$ 0.11 \\
\midrule
\multirow{4}{*}{Ballroom}
& 3DGS & 3.077 $\pm$ 0.068 & 22.48 $\pm$ 0.08 & 0.824 $\pm$ 0.001 & 0.102 $\pm$ 0.003 & 0.038 $\pm$ 0.000 & 24.15 $\pm$ 0.06 \\
& VDGS & 3.802 $\pm$ 0.007 & 43.83 $\pm$ 0.13 & 0.830 $\pm$ 0.002 & 0.105 $\pm$ 0.001 & 0.035 $\pm$ 0.001 & 24.83 $\pm$ 0.58 \\
& GSNB & 2.394 $\pm$ 0.062 & 48.23 $\pm$ 0.18 & 0.821 $\pm$ 0.001 & 0.105 $\pm$ 0.001 & 0.037 $\pm$ 0.001 & 24.14 $\pm$ 0.08 \\
& NRF-GS & 1.780 $\pm$ 0.021 & 29.15 $\pm$ 0.05 & 0.844 $\pm$ 0.001 & 0.102 $\pm$ 0.001 & 0.033 $\pm$ 0.001 & 24.60 $\pm$ 0.58 \\
\midrule
\multirow{4}{*}{Barn}
& 3DGS & 1.071 $\pm$ 0.040 & 11.72 $\pm$ 0.08 & 0.853 $\pm$ 0.001 & 0.163 $\pm$ 0.001 & 0.026 $\pm$ 0.001 & 27.45 $\pm$ 0.03 \\
& VDGS & 1.414 $\pm$ 0.006 & 19.57 $\pm$ 0.03 & 0.856 $\pm$ 0.001 & 0.166 $\pm$ 0.001 & 0.024 $\pm$ 0.001 & 27.80 $\pm$ 0.02 \\
& GSNB & 0.976 $\pm$ 0.006 & 22.95 $\pm$ 0.05 & 0.853 $\pm$ 0.001 & 0.163 $\pm$ 0.002 & 0.023 $\pm$ 0.001 & 27.77 $\pm$ 0.04 \\
& NRF-GS & 0.694 $\pm$ 0.001 & 15.37 $\pm$ 0.06 & 0.855 $\pm$ 0.001 & 0.164 $\pm$ 0.001 & 0.022 $\pm$ 0.001 & 27.98 $\pm$ 0.02 \\
\midrule
\multirow{4}{*}{Caterpillar}
& 3DGS & 1.308 $\pm$ 0.026 & 12.65 $\pm$ 0.15 & 0.791 $\pm$ 0.001 & 0.206 $\pm$ 0.005 & 0.053 $\pm$ 0.001 & 23.46 $\pm$ 0.02 \\
& VDGS & 1.621 $\pm$ 0.013 & 21.58 $\pm$ 0.03 & 0.793 $\pm$ 0.003 & 0.200 $\pm$ 0.002 & 0.052 $\pm$ 0.001 & 23.50 $\pm$ 0.10 \\
& GSNB & 1.179 $\pm$ 0.004 & 27.17 $\pm$ 0.08 & 0.791 $\pm$ 0.001 & 0.196 $\pm$ 0.001 & 0.050 $\pm$ 0.001 & 23.83 $\pm$ 0.12 \\
& NRF-GS & 0.860 $\pm$ 0.003 & 17.15 $\pm$ 0.05 & 0.806 $\pm$ 0.005 & 0.198 $\pm$ 0.001 & 0.046 $\pm$ 0.001 & 24.07 $\pm$ 0.05 \\
\midrule
\multirow{4}{*}{Church}
& 3DGS & 2.323 $\pm$ 0.033 & 18.62 $\pm$ 0.08 & 0.810 $\pm$ 0.001 & 0.181 $\pm$ 0.001 & 0.052 $\pm$ 0.001 & 22.85 $\pm$ 0.03 \\
& VDGS & 2.536 $\pm$ 0.015 & 32.73 $\pm$ 0.10 & 0.814 $\pm$ 0.002 & 0.181 $\pm$ 0.001 & 0.052 $\pm$ 0.001 & 23.16 $\pm$ 0.04 \\
& GSNB & 2.270 $\pm$ 0.016 & 45.60 $\pm$ 0.05 & 0.810 $\pm$ 0.003 & 0.174 $\pm$ 0.001 & 0.052 $\pm$ 0.001 & 22.92 $\pm$ 0.03 \\
& NRF-GS & 1.510 $\pm$ 0.033 & 25.10 $\pm$ 1.26 & 0.828 $\pm$ 0.002 & 0.177 $\pm$ 0.001 & 0.047 $\pm$ 0.001 & 23.29 $\pm$ 0.02 \\
\midrule
\multirow{4}{*}{Courtroom}
& 3DGS & 2.937 $\pm$ 0.011 & 19.60 $\pm$ 0.05 & 0.786 $\pm$ 0.001 & 0.171 $\pm$ 0.001 & 0.052 $\pm$ 0.001 & 22.78 $\pm$ 0.02 \\
& VDGS & 2.892 $\pm$ 0.028 & 32.02 $\pm$ 0.08 & 0.798 $\pm$ 0.002 & 0.157 $\pm$ 0.001 & 0.053 $\pm$ 0.001 & 23.09 $\pm$ 0.09 \\
& GSNB & 2.886 $\pm$ 0.071 & 54.22 $\pm$ 0.08 & 0.770 $\pm$ 0.003 & 0.172 $\pm$ 0.001 & 0.055 $\pm$ 0.000 & 22.55 $\pm$ 0.13 \\
& NRF-GS & 1.407 $\pm$ 0.016 & 24.07 $\pm$ 0.10 & 0.819 $\pm$ 0.002 & 0.151 $\pm$ 0.000 & 0.044 $\pm$ 0.001 & 24.17 $\pm$ 0.04 \\
\midrule
\multirow{4}{*}{Family}
& 3DGS & 2.279 $\pm$ 0.087 & 15.70 $\pm$ 0.10 & 0.866 $\pm$ 0.001 & 0.098 $\pm$ 0.002 & 0.036 $\pm$ 0.001 & 24.41 $\pm$ 0.03 \\
& VDGS & 2.510 $\pm$ 0.028 & 26.48 $\pm$ 0.03 & 0.868 $\pm$ 0.002 & 0.097 $\pm$ 0.003 & 0.036 $\pm$ 0.001 & 24.67 $\pm$ 0.03 \\
& GSNB & 2.058 $\pm$ 0.056 & 41.42 $\pm$ 0.10 & 0.880 $\pm$ 0.065 & 0.106 $\pm$ 0.004 & 0.040 $\pm$ 0.001 & 23.89 $\pm$ 0.05 \\
& NRF-GS & 1.164 $\pm$ 0.021 & 19.90 $\pm$ 0.05 & 0.869 $\pm$ 0.001 & 0.106 $\pm$ 0.001 & 0.033 $\pm$ 0.001 & 24.77 $\pm$ 0.03 \\
\midrule
\multirow{4}{*}{Francis}
& 3DGS & 0.807 $\pm$ 0.005 & 10.43 $\pm$ 0.08 & 0.895 $\pm$ 0.001 & 0.178 $\pm$ 0.001 & 0.032 $\pm$ 0.001 & 27.09 $\pm$ 0.03 \\
& VDGS & 0.816 $\pm$ 0.004 & 15.77 $\pm$ 0.15 & 0.898 $\pm$ 0.003 & 0.176 $\pm$ 0.001 & 0.033 $\pm$ 0.001 & 27.23 $\pm$ 0.02 \\
& GSNB & 0.796 $\pm$ 0.007 & 20.08 $\pm$ 0.13 & 0.904 $\pm$ 0.004 & 0.172 $\pm$ 0.002 & 0.032 $\pm$ 0.001 & 27.61 $\pm$ 0.04 \\
& NRF-GS & 0.443 $\pm$ 0.009 & 12.33 $\pm$ 0.08 & 0.911 $\pm$ 0.001 & 0.177 $\pm$ 0.001 & 0.029 $\pm$ 0.001 & 28.06 $\pm$ 0.06 \\
\midrule
\multirow{4}{*}{Horse}
& 3DGS & 1.360 $\pm$ 0.022 & 12.33 $\pm$ 0.08 & 0.876 $\pm$ 0.001 & 0.110 $\pm$ 0.001 & 0.037 $\pm$ 0.002 & 23.89 $\pm$ 0.05 \\
& VDGS & 1.488 $\pm$ 0.015 & 19.43 $\pm$ 0.08 & 0.884 $\pm$ 0.001 & 0.112 $\pm$ 0.002 & 0.035 $\pm$ 0.001 & 24.23 $\pm$ 0.03 \\
& GSNB & 1.349 $\pm$ 0.008 & 29.88 $\pm$ 0.10 & 0.863 $\pm$ 0.002 & 0.109 $\pm$ 0.002 & 0.039 $\pm$ 0.001 & 23.74 $\pm$ 0.04 \\
& NRF-GS & 0.781 $\pm$ 0.011 & 15.98 $\pm$ 0.08 & 0.887 $\pm$ 0.003 & 0.109 $\pm$ 0.001 & 0.027 $\pm$ 0.001 & 24.98 $\pm$ 0.02 \\
\midrule
\multirow{4}{*}{Ignatius}
& 3DGS & 3.149 $\pm$ 0.005 & 19.52 $\pm$ 0.03 & 0.778 $\pm$ 0.002 & 0.156 $\pm$ 0.003 & 0.063 $\pm$ 0.001 & 21.71 $\pm$ 0.03 \\
& VDGS & 3.310 $\pm$ 0.007 & 31.92 $\pm$ 0.10 & 0.779 $\pm$ 0.001 & 0.160 $\pm$ 0.002 & 0.062 $\pm$ 0.001 & 21.68 $\pm$ 0.04 \\
& GSNB & 2.720 $\pm$ 0.013 & 52.88 $\pm$ 0.08 & 0.767 $\pm$ 0.004 & 0.162 $\pm$ 0.001 & 0.062 $\pm$ 0.001 & 21.70 $\pm$ 0.02 \\
& NRF-GS & 1.631 $\pm$ 0.017 & 24.88 $\pm$ 0.10 & 0.787 $\pm$ 0.001 & 0.167 $\pm$ 0.001 & 0.051 $\pm$ 0.001 & 22.76 $\pm$ 0.02 \\
\bottomrule
\end{tabular}}
\end{table*}

\newpage
\begin{table*}[hp]
\centering
\caption{Per-scene test-set results on Tanks and Temples, continued.}
\label{tab:tnt_per_scene_cont2}
\resizebox{\textwidth}{!}{
\begin{tabular}{llcccccc}
\toprule
Scene & Method & Points (M)$\downarrow$ & Time (min)$\downarrow$ & MS-SSIM$\uparrow$ & LPIPS$\downarrow$ & L1$\downarrow$ & PSNR$\uparrow$ \\
\midrule
\multirow{4}{*}{Lighthouse}
& 3DGS & 0.870 $\pm$ 0.025 & 12.40 $\pm$ 0.22 & 0.843 $\pm$ 0.002 & 0.161 $\pm$ 0.001 & 0.052 $\pm$ 0.001 & 22.31 $\pm$ 0.02 \\
& VDGS & 1.201 $\pm$ 0.031 & 22.37 $\pm$ 0.10 & 0.841 $\pm$ 0.002 & 0.162 $\pm$ 0.001 & 0.051 $\pm$ 0.001 & 22.54 $\pm$ 0.01 \\
& GSNB & 0.819 $\pm$ 0.006 & 21.45 $\pm$ 0.13 & 0.846 $\pm$ 0.004 & 0.165 $\pm$ 0.002 & 0.052 $\pm$ 0.001 & 22.57 $\pm$ 0.01 \\
& NRF-GS & 0.522 $\pm$ 0.008 & 15.32 $\pm$ 0.18 & 0.854 $\pm$ 0.001 & 0.158 $\pm$ 0.001 & 0.045 $\pm$ 0.001 & 23.52 $\pm$ 0.01 \\
\midrule
\multirow{4}{*}{M60}
& 3DGS & 1.652 $\pm$ 0.006 & 14.73 $\pm$ 0.06 & 0.903 $\pm$ 0.001 & 0.120 $\pm$ 0.001 & 0.023 $\pm$ 0.001 & 27.82 $\pm$ 0.05 \\
& VDGS & 2.349 $\pm$ 0.003 & 27.78 $\pm$ 0.08 & 0.904 $\pm$ 0.001 & 0.119 $\pm$ 0.001 & 0.022 $\pm$ 0.001 & 28.05 $\pm$ 0.05 \\
& GSNB & 1.518 $\pm$ 0.005 & 32.62 $\pm$ 0.03 & 0.904 $\pm$ 0.001 & 0.111 $\pm$ 0.004 & 0.022 $\pm$ 0.001 & 28.31 $\pm$ 0.03 \\
& NRF-GS & 1.090 $\pm$ 0.037 & 20.42 $\pm$ 0.03 & 0.910 $\pm$ 0.001 & 0.120 $\pm$ 0.001 & 0.020 $\pm$ 0.001 & 28.45 $\pm$ 0.04 \\
\midrule
\multirow{4}{*}{Museum}
& 3DGS & 4.601 $\pm$ 0.017 & 25.77 $\pm$ 0.06 & 0.738 $\pm$ 0.001 & 0.159 $\pm$ 0.001 & 0.066 $\pm$ 0.001 & 20.83 $\pm$ 0.01 \\
& VDGS & 4.113 $\pm$ 0.015 & 41.10 $\pm$ 0.05 & 0.740 $\pm$ 0.001 & 0.160 $\pm$ 0.001 & 0.068 $\pm$ 0.001 & 20.88 $\pm$ 0.04 \\
& GSNB & 4.101 $\pm$ 0.031 & 75.18 $\pm$ 0.16 & 0.746 $\pm$ 0.002 & 0.185 $\pm$ 0.001 & 0.077 $\pm$ 0.000 & 19.75 $\pm$ 0.04 \\
& NRF-GS & 2.020 $\pm$ 0.029 & 31.42 $\pm$ 0.08 & 0.784 $\pm$ 0.001 & 0.150 $\pm$ 0.001 & 0.063 $\pm$ 0.001 & 21.33 $\pm$ 0.02 \\
\midrule
\multirow{4}{*}{Palace}
& 3DGS & 0.725 $\pm$ 0.012 & 11.18 $\pm$ 0.08 & 0.736 $\pm$ 0.001 & 0.351 $\pm$ 0.001 & 0.080 $\pm$ 0.001 & 19.92 $\pm$ 0.09 \\
& VDGS & 0.905 $\pm$ 0.008 & 19.75 $\pm$ 0.05 & 0.740 $\pm$ 0.002 & 0.331 $\pm$ 0.001 & 0.072 $\pm$ 0.001 & 20.62 $\pm$ 0.02 \\
& GSNB & 0.781 $\pm$ 0.008 & 19.97 $\pm$ 0.03 & 0.771 $\pm$ 0.001 & 0.333 $\pm$ 0.001 & 0.077 $\pm$ 0.001 & 20.23 $\pm$ 0.03 \\
& NRF-GS & 0.478 $\pm$ 0.003 & 18.90 $\pm$ 0.05 & 0.765 $\pm$ 0.001 & 0.330 $\pm$ 0.001 & 0.069 $\pm$ 0.001 & 20.84 $\pm$ 0.01 \\
\midrule
\multirow{4}{*}{Panther}
& 3DGS & 1.780 $\pm$ 0.005 & 14.87 $\pm$ 0.12 & 0.908 $\pm$ 0.002 & 0.115 $\pm$ 0.005 & 0.024 $\pm$ 0.001 & 28.39 $\pm$ 0.01 \\
& VDGS & 2.048 $\pm$ 0.048 & 24.85 $\pm$ 0.05 & 0.909 $\pm$ 0.001 & 0.111 $\pm$ 0.001 & 0.023 $\pm$ 0.001 & 28.58 $\pm$ 0.02 \\
& GSNB & 1.616 $\pm$ 0.013 & 34.05 $\pm$ 0.05 & 0.907 $\pm$ 0.001 & 0.111 $\pm$ 0.002 & 0.023 $\pm$ 0.001 & 28.59 $\pm$ 0.01 \\
& NRF-GS & 0.944 $\pm$ 0.033 & 18.93 $\pm$ 0.03 & 0.912 $\pm$ 0.001 & 0.111 $\pm$ 0.001 & 0.021 $\pm$ 0.001 & 28.68 $\pm$ 0.02 \\
\midrule
\multirow{4}{*}{Playground}
& 3DGS & 2.269 $\pm$ 0.008 & 17.42 $\pm$ 0.08 & 0.861 $\pm$ 0.001 & 0.154 $\pm$ 0.002 & 0.036 $\pm$ 0.001 & 25.77 $\pm$ 0.04 \\
& VDGS & 2.665 $\pm$ 0.035 & 31.08 $\pm$ 0.10 & 0.864 $\pm$ 0.001 & 0.160 $\pm$ 0.001 & 0.035 $\pm$ 0.001 & 26.19 $\pm$ 0.04 \\
& GSNB & 2.049 $\pm$ 0.046 & 42.13 $\pm$ 0.08 & 0.863 $\pm$ 0.001 & 0.148 $\pm$ 0.002 & 0.031 $\pm$ 0.001 & 26.69 $\pm$ 0.05 \\
& NRF-GS & 1.240 $\pm$ 0.003 & 22.20 $\pm$ 0.05 & 0.866 $\pm$ 0.001 & 0.159 $\pm$ 0.001 & 0.029 $\pm$ 0.001 & 26.94 $\pm$ 0.03 \\
\midrule
\multirow{4}{*}{Temple}
& 3DGS & 0.850 $\pm$ 0.004 & 11.23 $\pm$ 0.12 & 0.799 $\pm$ 0.002 & 0.226 $\pm$ 0.002 & 0.065 $\pm$ 0.001 & 20.45 $\pm$ 0.05 \\
& VDGS & 1.057 $\pm$ 0.038 & 18.65 $\pm$ 0.05 & 0.809 $\pm$ 0.002 & 0.224 $\pm$ 0.001 & 0.063 $\pm$ 0.001 & 20.71 $\pm$ 0.01 \\
& GSNB & 0.910 $\pm$ 0.003 & 22.25 $\pm$ 0.00 & 0.794 $\pm$ 0.003 & 0.223 $\pm$ 0.001 & 0.064 $\pm$ 0.001 & 20.49 $\pm$ 0.05 \\
& NRF-GS & 0.624 $\pm$ 0.006 & 15.02 $\pm$ 0.03 & 0.822 $\pm$ 0.001 & 0.211 $\pm$ 0.001 & 0.052 $\pm$ 0.001 & 22.00 $\pm$ 0.03 \\
\midrule
\multirow{4}{*}{Train}
& 3DGS & 1.107 $\pm$ 0.004 & 12.35 $\pm$ 0.05 & 0.788 $\pm$ 0.003 & 0.173 $\pm$ 0.001 & 0.060 $\pm$ 0.001 & 21.63 $\pm$ 0.01 \\
& VDGS & 1.631 $\pm$ 0.004 & 24.23 $\pm$ 0.03 & 0.799 $\pm$ 0.001 & 0.170 $\pm$ 0.001 & 0.055 $\pm$ 0.002 & 22.10 $\pm$ 0.01 \\
& GSNB & 1.115 $\pm$ 0.002 & 26.50 $\pm$ 0.05 & 0.785 $\pm$ 0.002 & 0.173 $\pm$ 0.001 & 0.062 $\pm$ 0.002 & 21.93 $\pm$ 0.56 \\
& NRF-GS & 0.831 $\pm$ 0.002 & 17.55 $\pm$ 0.05 & 0.805 $\pm$ 0.003 & 0.169 $\pm$ 0.001 & 0.054 $\pm$ 0.001 & 22.47 $\pm$ 0.01 \\
\midrule
\multirow{4}{*}{Truck}
& 3DGS & 2.594 $\pm$ 0.006 & 17.72 $\pm$ 0.19 & 0.853 $\pm$ 0.001 & 0.111 $\pm$ 0.002 & 0.035 $\pm$ 0.001 & 24.10 $\pm$ 0.08 \\
& VDGS & 2.949 $\pm$ 0.010 & 30.47 $\pm$ 0.10 & 0.855 $\pm$ 0.002 & 0.114 $\pm$ 0.001 & 0.033 $\pm$ 0.001 & 24.44 $\pm$ 0.02 \\
& GSNB & 2.091 $\pm$ 0.029 & 41.48 $\pm$ 0.13 & 0.835 $\pm$ 0.003 & 0.120 $\pm$ 0.002 & 0.041 $\pm$ 0.001 & 23.55 $\pm$ 0.03 \\
& NRF-GS & 1.361 $\pm$ 0.012 & 22.58 $\pm$ 0.08 & 0.855 $\pm$ 0.001 & 0.120 $\pm$ 0.000 & 0.033 $\pm$ 0.001 & 24.45 $\pm$ 0.02 \\
\bottomrule
\end{tabular}}
\end{table*}


\newpage

\end{document}